%% file: acl.tex
\pdfoutput=1

\documentclass[11pt]{article}

\usepackage{acl}

\usepackage{times}
\usepackage{latexsym}

\usepackage[T1]{fontenc}

\usepackage[utf8]{inputenc}

\usepackage{microtype}

\usepackage{amsmath}
\usepackage{amssymb}
\usepackage{amsfonts}
\usepackage{graphicx}
\usepackage{subcaption}
\usepackage{array}
\usepackage{makecell}
\usepackage{comment}

\newcommand{\rv}[1]{\mathbf{#1}}
\newcommand{\myvec}[1]{\mathbf{#1}}
\newcommand{\mymodele}{\textbf{E}}
\newcommand{\mymodelc}{\textbf{C}}

\newcommand{\ignore}[1]{}

\newenvironment{itemizesquish}[2]{\begin{list}{\labelitemi}{\setlength{\itemsep}{#1}\setlength{\labelwidth}{#2}\setlength{\leftmargin}{\labelwidth}\addtolength{\leftmargin}{\labelsep}}}{\end{list}}

\title{Understanding Domain Learning in Language Models \\ Through Subpopulation Analysis}

\author{Zheng Zhao \quad Yftah Ziser \quad Shay B. Cohen \\
  Institute for Language, Cognition and Computation \\
  School of Informatics, University of Edinburgh \\
  10 Crichton Street, Edinburgh, EH8 9AB \\
  \texttt{\{zheng.zhao,yftah.ziser\}@ed.ac.uk} ,
  \texttt{scohen@inf.ed.ac.uk}}
  
\newcommand{\shaycomment}[1]{\textcolor{blue}{#1 -- Shay}}
\newcommand{\yftahcomment}[1]{\textcolor{red}{#1 -- yftah}}

\begin{document}
\maketitle
\begin{abstract}
We investigate how different domains are encoded in modern neural network architectures. We analyze the relationship between natural language domains, model size, and the amount of training data used. The primary analysis tool we develop is based on subpopulation analysis with Singular Vector Canonical Correlation Analysis (SVCCA), which we apply to Transformer-based language models (LMs). We compare the latent representations of such a language model at its different layers from a pair of models: a model trained on multiple domains (an experimental model) and a model trained on a single domain (a control model). Through our method, we find that increasing the model capacity impacts how domain information is stored in upper and lower layers differently. In addition, we show that larger experimental models simultaneously embed domain-specific information as if they were conjoined control models. These findings are confirmed qualitatively, demonstrating the validity of our method.

\end{abstract}

\input{introduction}

\input{methodology}

\input{experiments}

\input{results}

\input{related_works}

\input{conclusion}

\section*{Acknowledgments}
This work was supported by the UKRI Centre for Doctoral Training (CDT) in Natural Language Processing through UKRI grant EP/S022481/1 and CDT funding from Huawei. We would like to thank Bonnie Webber, Ivan Titov and the anonymous reviewers for their helpful feedback. We appreciate the use of computing resources through the CSD3 cluster at the University of Cambridge.

\input{limitation}

\bibliography{anthology,custom}
\bibliographystyle{acl_natbib}

\newpage
\appendix

\input{appendix}

\end{document}

%% file: introduction.tex
\section{Introduction}
\label{sec:introduction}

Pre-trained language models (PLMs) have become an essential modeling component for state-of-the-art natural language processing (NLP) models. They process text into latent representations in such a way that allows an NLP practitioner to seamlessly use these representations for prediction problems of various degrees of difficulty
\cite{wang-etal-2018-glue,wang-etal-2019-superglue}. The opaqueness in obtaining these representations has been an important research topic in the NLP community. PLMs, and more generally, neural models, are currently studied to understand their process and behavior in obtaining their latent representations. These PLMs are often trained on large datasets, with inputs originating from different sources. In this paper, we further develop our understanding of how neural networks obtain their latent representation and study the effect of learning from various domains on the characteristics of the corresponding latent representations.

Texts come from various domains that differ in their writing styles, authors and topics \cite{plank-2016-what}. In this work, we follow a simple definition of a domain as \emph{a corpus of documents sharing a common
topic}. We rely on a simple tool of subpopulation analysis to compare and contrast latent representations obtained with and without a specific domain.
\begin{figure}

     \centering
     \begin{subfigure}[b]{0.235\textwidth}
         \centering
         \includegraphics[width=\textwidth]{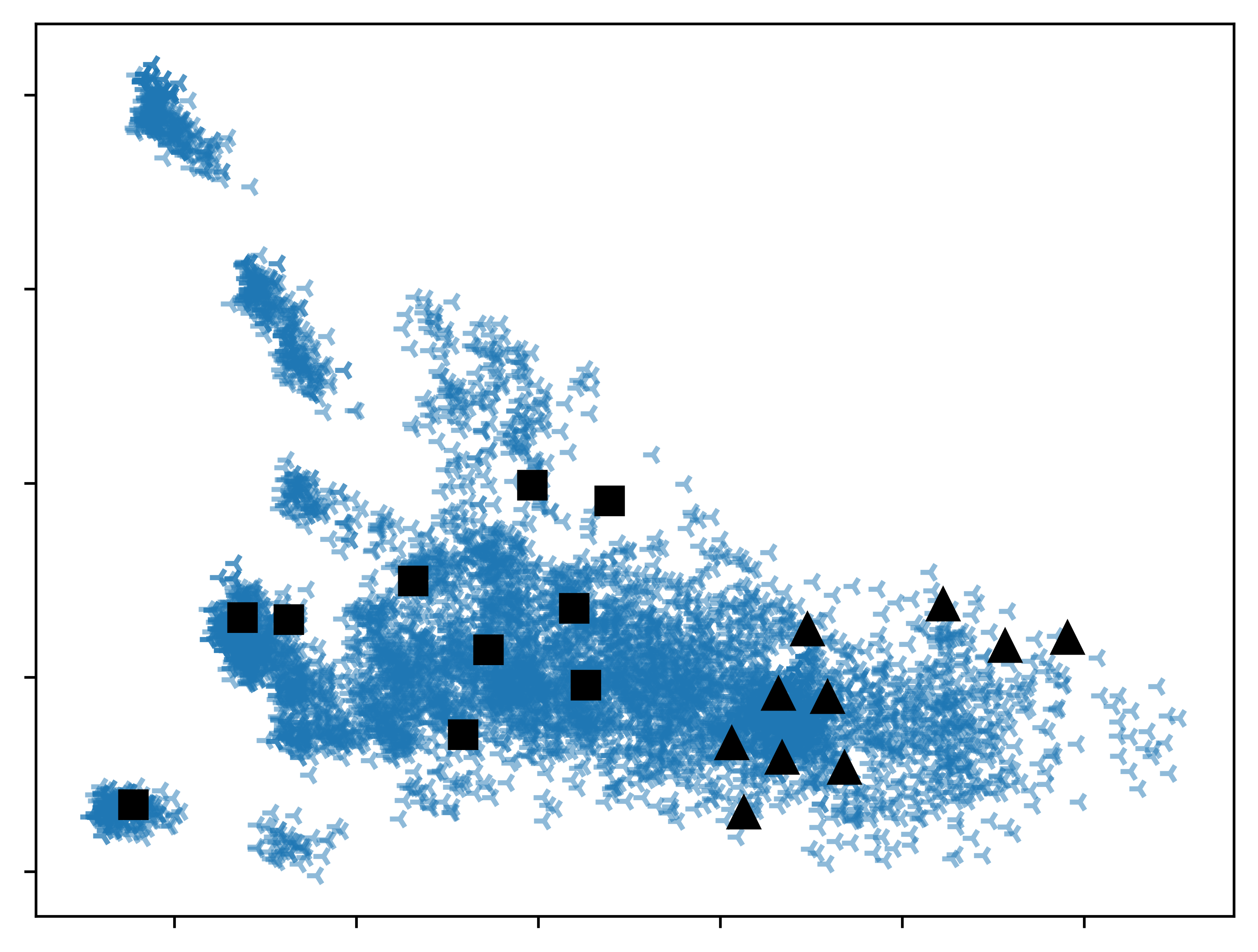}
         \caption{Experimental model}
     \end{subfigure}
     \hfill
     \begin{subfigure}[b]{0.235\textwidth}
         \centering
         \includegraphics[width=\textwidth]{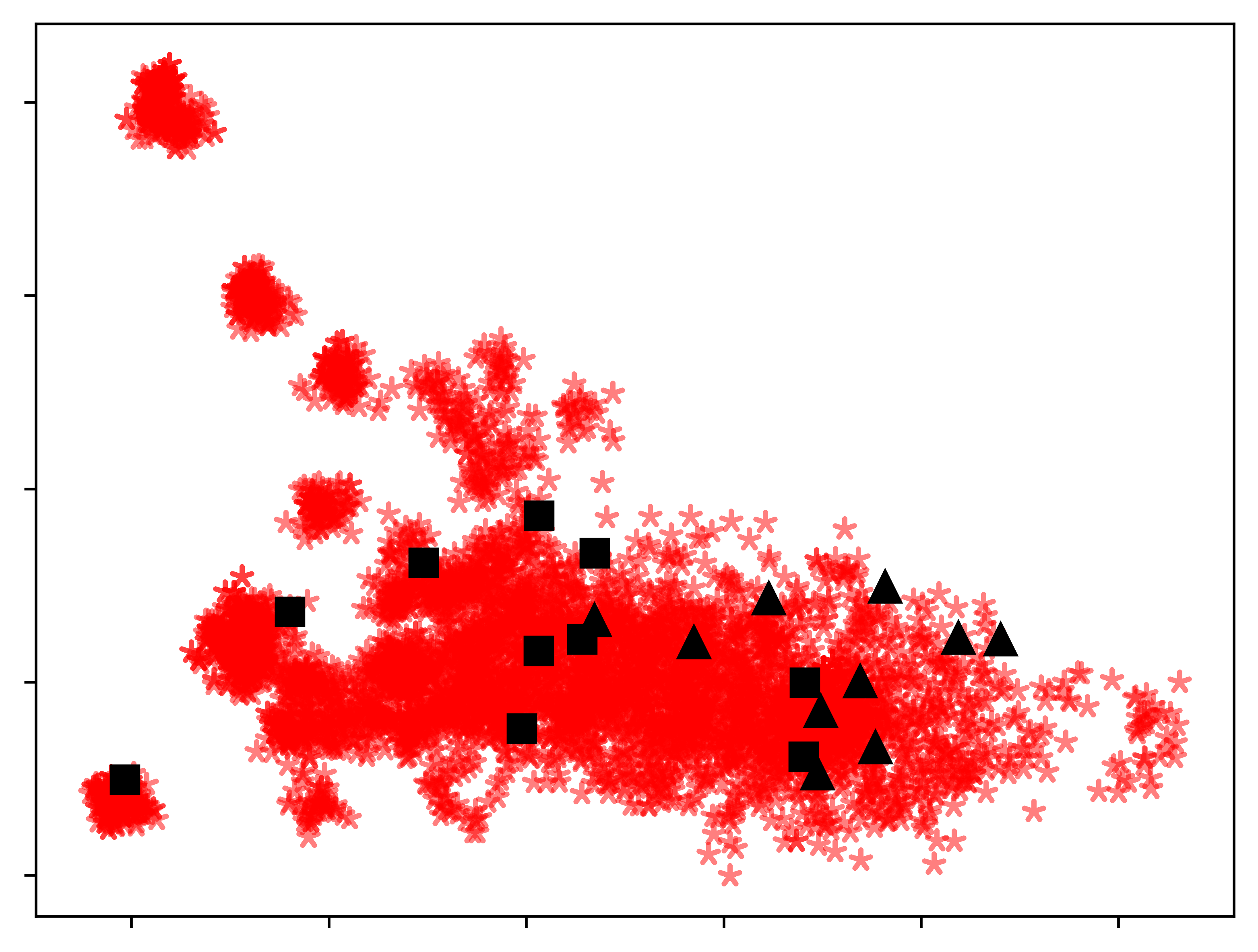}         \caption{Control model}
     \end{subfigure}
\caption{\label{fig:A}An example of a visualization used with our subpopulation analysis tool. The experimental model, which includes all domain data, separates in its latent representations words related to the \texttt{Books} domain ($\blacktriangle$) from general words ($\blacksquare$). The control model, on the other hand, mixes them together.}
\end{figure}
Our analysis relies on constructing two types of models: \emph{experimental} models, from multi-domain
data, and \emph{control} models, from single-domain data. Figure~\ref{fig:A}
describes an example in which this analysis is applied to study the way embeddings for domain-specific
words cluster together in the experimental and control model representations.

We believe training in an implicit multi-domain setup is widespread and often overlooked. For example, SQuAD \cite{rajpurkar-etal-2016-squad}, a widely used question-answering dataset composed of Wikipedia articles from multiple domains, is often referred to as a single-domain dataset in domain adaptation works for simplicity \cite{hazen2019towards, shakeri-etal-2020-end, yue-etal-2021-contrastive}. This scenario is also common in text summarization, where many datasets consist of a bundle of domains for news articles \cite{grusky-etal-2018-newsroom}, academic papers \cite{cohan-etal-2018-discourse,fonseca2022}, and do-it-yourself (DIY) guides \cite{cohen-etal-2021-wikisum}. While models that learn from multiple domains are frequently used, their nature and behavior have hardly been explored. 

Our work sheds light on the way state-of-the-art multi-domain models encode domain-specific information. We focus on two main aspects highly relevant for many training procedures: model capacity and data size. We discover that model capacity, indicated by the number of its parameters, strongly impacts the amount of domain-specific information multi-domain models store. This property might explain the performance gains of larger models \citep{devlin-etal-2019-bert,raffel-etal-2020-exploring,clark2020electra,srivastava2022beyond}. 
While this paper focuses on studying the effect of domains on latent representations, the subpopulation analysis tool could be used for studying other NLP setups, such as multitask and multimodal learning.\footnote{Our code is available at: \url{https://github.com/zsquaredz/subpopulation_analysis}}

%% file: methodology.tex
\section{Methodology}
\label{sec:method}




For an integer $n$, we denote by $[n]$ the set $\{ 1, \ldots, n\}$. Our analysis tool assumes a distribution $p(\rv{X})$ from which a set of examples $\mathcal{X} = \{ \myvec{x}^{(i)} \mid i \in [n] \}$
is drawn. It also assumes a family of binary indicators $\pi_1, \ldots, \pi_d$ such that $\pi_i(\rv{x})$ indicates whether the example $\rv{x}$ satisfies a certain \emph{subpopulation} attribute $i$. For example, in this paper
we focus on domain analysis, so $\pi_5$ could indicate if an example belongs to a \texttt{Books} domain.

We denote by $\mathcal{X} \big|_{\pi_i}$ the set $\{ \myvec{x}^{(j)} \mid \pi_i(\myvec{x}^{(j)}) = 1 \}$, the subset of $\mathcal{X}$ that satisfies attribute $i$.
Unlike standard diagnostic classifier methods \cite{belinkov-etal-2017-neural,belinkov-etal-2017-evaluating,giulianelli-etal-2018-hood}, rather than building a model to \emph{predict} the attribute, we
perform subpopulation analysis by training a set of models: $\mymodele{}$, trained from $\mathcal{X}$ (the \emph{experimental} model), and
$\mymodelc{}_i$, trained from $\mathcal{X} \big|_{\pi_i}$ (the \emph{control} model). We borrow the terminology of ``experimental'' and ``control''
from experimental design such as in clinical trials \cite{hinkelmann2007design}. The experimental model corresponds to the experimental (or ``treatment'' in the case
of medical trials) group in such trials and the control model corresponds to the control group. Unlike a standard experimental design, rather than comparing
a function (such as squared difference) between the outcomes of the two groups to calculate a statistic with an underlying distribution, we instead
calculate the similarity values between the representations of the two models. Our analysis is also related
to Representational Similarity Analysis \cite{dimsdale2018representational}, aimed at studying similarities (across different experimental settings)
between activation levels in brain neurons.

Through their latent representations, the set of models $\mymodelc{}_i$ represent the information that is captured about $p(\rv{X})$
from the relevant subpopulation of data. By comparing the different models to each other, we can learn what information
is captured in the latent representations when a subset of the data is used and whether this information is different from the one
captured when the whole set of data is used. With a proper control for model size and subpopulation sizes, we can determine
the relationship between the different attributes $\pi_i$ and the corresponding representations in different model components. The remaining question now is \emph{how do we compare these representations?} Here, we follow previous work \cite{saphra-lopez-2019-understanding, bau-etal-2019-identifying, kudugunta-etal-2019-investigating},
and apply Singular Vector Canonical Correlation Analysis (SVCCA; \citealt{raghu-etal-2017-svcca}) 
to the latent representations of the experimental and control models.

We assume that each example $\myvec{x}^{(i)}$ is associated with a latent representation $\myvec{h}^{(i)}_j$ given by $\mymodelc{}_j$.
For example, this could be the representation in the embedding layer for the input example,
or the representation in the final pre-output layer. We define $\mathcal{H}_j$ to be a set of latent
representations $\mathcal{H}_j = \{ \myvec{h}^{(k)}_j \mid k \in [n] \}$ for model $\mymodelc{}_j$.
We define $\mathcal{H}_j \big|_{\pi_i} = \{ \myvec{h}^{(k)}_j \mid \pi_i(\myvec{x}^{(k)}) = 1 \}$ -- the latent representations of
$\mymodelc{}_j$ for which attribute $i$ fires. Similarly, we define $\mathcal{H}_0$ for the model $\mymodele{}$.
We calculate the SVCCA value between subsets of $\mathcal{H}_0$ and subsets of $\mathcal{H}_j$ for $j \ge 1$. The procedure of SVCCA in this case follows:

\begin{itemizesquish}{-0.3em}{0.5em}

\item Performing Singular Value Decomposition (SVD) on the matrix forms of $\mathcal{H}_0$ and $\mathcal{H}_j$ (matching the representations in each
through the index of the example $\myvec{x}^{(i)}$ from which they originate).
We use the lowest number of principal directions that preserve 99\% of the variance in the data to project the latent representations.

\item Performing Canonical Correlation Analysis (CCA; \citealt{hardoon-etal-2004-cca}) between the projections
of the latent representations from the SVD step, and calculating the average correlation value, denoted by $\rho_{0j}$.

\end{itemizesquish}

The SVD step, which may seem redundant, is actually crucial, as it had been shown that low variance directions
in neural network representations are primarily noise \citep{raghu-etal-2017-svcca,frankle-and-carbin-2019-lottery}.
The intensity of $\rho_{0j}$ indicates the level of overlap between the latent representations of each model
\cite{saphra-lopez-2019-understanding}.

\begin{figure}[t]
    \centering
    \includegraphics[width=0.8\columnwidth]{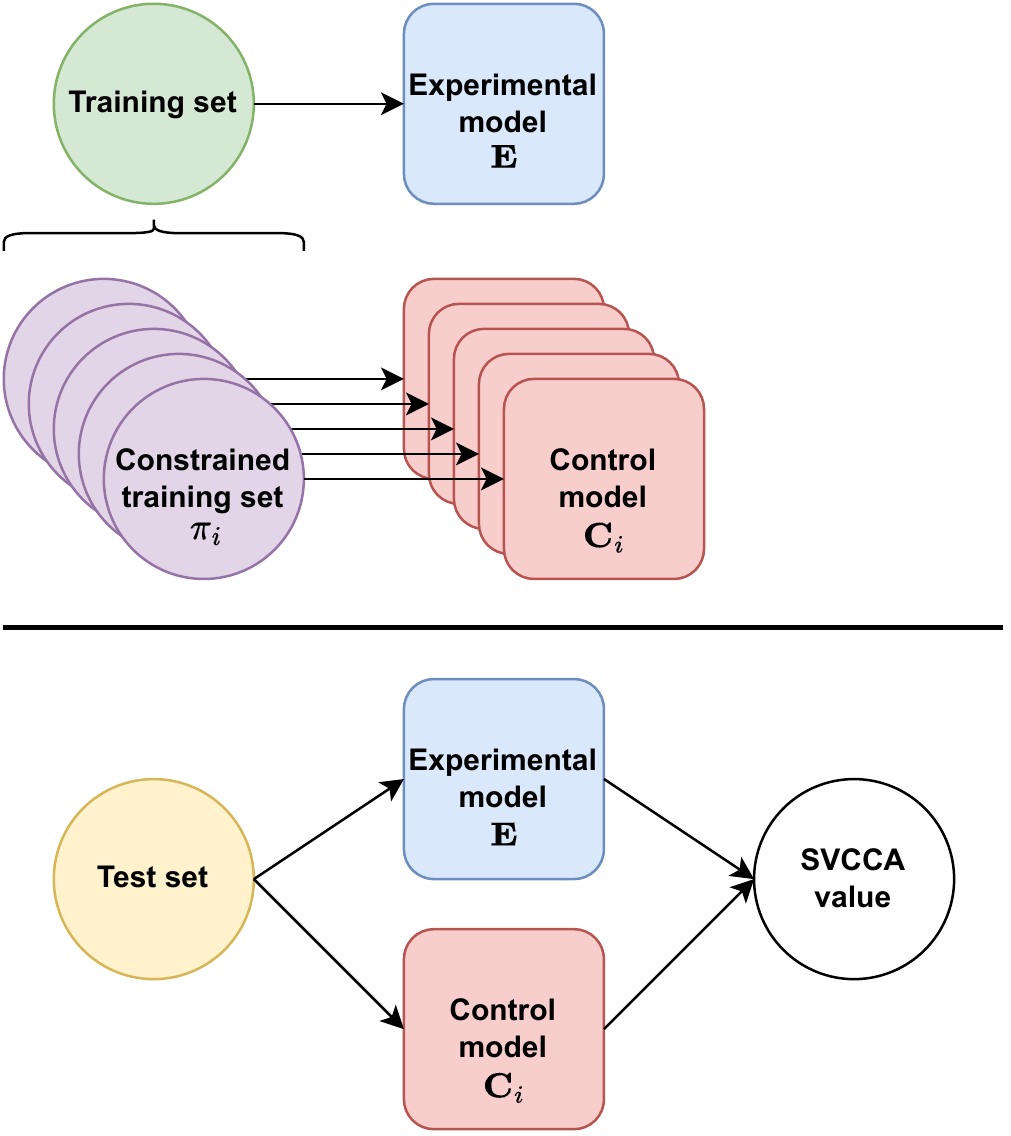}
    \caption{A diagram explaining the analysis we perform. 
    At the top, during training, we create two sets of models from constrained datasets (based on different $\pi_i$) and a dataset that is not constrained.
    The result of this training is two set of models, the experimental model ($\mymodele{}$) and control models ($\mymodelc{}_i$). 
    To perform the similarity analysis, we compute latent representation from a common test set for both models, and then run SVCCA (bottom).}
    \label{fig:scheme}
\end{figure}

In the rest of this paper, we use the tool of subpopulation analysis with $\mymodele{}$/$\mymodelc{}_i$ as above for the case of domain learning in neural networks.
We note that each time we use this tool, the following decisions need to be made: (a) what training set we use for each $\mymodele{}$ and $\mymodelc{}_i$; (b) the subset of $\mathcal{H}_j$ for $j \ge 0$ for which we perform the similarity analysis; (c) the component in the model from which we take the latent representations.
For (c), the component can be, for example, a layer. Indeed, for most of our experiments, we use the first and last layer to create the latent representation sets, as they stand in stark contrast to each other in their behavior (see \S~\ref{section:results}). We provide an illustration of our proposed pipeline in Figure~\ref{fig:scheme}. We are particularly interested in studying the effect of two aspects of learning: dataset size and model capacity.

\paragraph{The case of domains}
\ignore{
The use of massively pre-trained language models such as BERT \citep{devlin-etal-2019-bert}, BART \citep{lewis-etal-2020-bart}, or T5 \citep{raffel-etal-2020-exploring} has led to significant
progress in many NLP tasks. Previous work attempted to ``probe'' these models for the morphological, syntactic, and semantic information they capture \citep{tenney-etal-2019-bert,Goldberg2019AssessingBS,clark-etal-2019-bert}.
Further work \citet{aharoni-goldberg-2020-unsupervised} makes the claim that the \textit{domain}, where the data comes from or the data distribution, is an important aspect that remained overlooked. While the way to define the notion of domain is not yet agreed upon, a definition that has a reasonably significant consensus is for domain to be a corpus from a specific data source that differs from other domains in topic, genre, style,
level of formality  \citep{koehn-knowles-2017-six}.}

In this paper, we define a domain as a corpus of documents with a common topic. Since a single massive web-crawled corpus used to pre-train language models usually contains many domains, we examine to what extent domain-specific information is encoded in the pre-trained model learned on this corpus. 
Such domain membership is indicated by our attribute functions $\pi_i$. For example, we may use $\pi_5(\myvec{x})$ to indicate whether $\myvec{x}$ is an input example from the domain \texttt{Books}. Given this notion of a domain, we can readily use subpopulation analysis through experimental and control models to analyze the effect on neural representations of
learning from multiple domains or a single domain.

%% file: experiments.tex
\section{Experimental Setup}
\label{sec:experimental_setup}

\paragraph{Data}
We use the Amazon Reviews dataset \citep{ni-etal-2019-justifying}, a dataset that facilitates research in tasks like sentiment analysis \cite{zhang2020multiclassification}, aspect-based sentiment analysis, and recommendation systems \cite{wang2020next}. The reviews in this dataset are explicitly divided into different product categories that serve as domains, which makes it a natural testbed for many multi-domain studies. A noteworthy example of a research field that heavily relies on this dataset is domain adaptation \cite{blitzer-etal-2007-biographies, ziser-reichart-2018-pivot, du-etal-2020-adversarial, lekhtman-etal-2021-dilbert,long-etal-2022-domain}, which is the task of learning robust models across different domains, closely related to our research.\footnote{We use the latest version of the dataset, consisting reviews from 1996 up to 2019.} 
We sort the domains by their review counts and pick the top five, which results in: \texttt{Books}, \texttt{Clothing Shoes and Jewelry}, \texttt{Electronics}, \texttt{Home and Kitchen}, and \texttt{Movies and TV} domains.  To further validate our data quality, we use the 5-core subset of the data, ensuring that all reviewed items have at least five reviews authored by reviewers who wrote at least five reviews.

A representative dataset sample is presented in Table \ref{tab:example_reviews}. We consider the different domains within the Amazon review dataset as \emph{lexical domains}, i.e., domains that share a similar textual structure and functionality but differ with respect to their vocabulary. For example, we see the review snippet from the \texttt{Books} domain contains an aspect (``ending'') for which a negative sentiment is conveyed (``didn't have a proper''). Similarly, we find an aspect (``handle'') with a corresponding conveyed sentiment (``too hot'') for the \texttt{Home} domain. We can see this shared pattern across all domains, with different aspects and sentiment terms. We would not expect this to be the case for other datasets, which might have different differentiators for domains. For example, Amazon reviews and Wikipedia pages on \texttt{Books} domain may have a similar vocabulary, however, a review is more likely to convey sentiment toward a particular book, and a Wikipedia article is more likely to focus on describing the book. Thus, the Amazon Reviews dataset is an ideal testbed for our analysis.

\begin{table}[t]
\footnotesize
\centering
\begin{tabular}{m{7.25cm}}
\Xhline{1pt}
\textbf{Books}: \ldots the book didn't have a proper ending but rather a rushed attempt to conclude the story and put everyone away neatly \ldots       \\
\textbf{Clothing}: \ldots clearly of awful quality, the design and paint was totally wrong, the mask was short and stumpy as well as slightly deformed and bent to the left \ldots    \\
\textbf{Home}: \ldots there are no handles, and the plastic gets too hot to hold, so you have to awkwardly pour by the top \ldots        \\
\Xhline{1pt}
\end{tabular}
\caption{A representative sample of review snippets.}
\label{tab:example_reviews}
\end{table}

In addition to the Amazon Reviews dataset, we experimented on the WikiSum dataset \citep{cohen-etal-2021-wikisum} to further validate our findings. The WikiSum dataset is a coherent paragraph summarization dataset based on the WikiHow website.\footnote{\url{https://www.wikihow.com}} WikiHow consists of do-it-yourself (DIY) guides for the general public, thus is written using simple English and ranges over many domains.  Similar to Amazon Reviews, we also pick the top five domains for our experiments: \texttt{Education}, \texttt{Food}, \texttt{Health}, \texttt{Home}, and \texttt{Pets}. Since the dataset is designed for summarization, we concatenate the document and summary together for our MLM task. We present the results for this dataset at the end of \S~\ref{section:results}. 

\paragraph{Task}
We study the language modeling task to understand the nature of multi-domain learning better. More precisely, we experiment with the masked language modeling (MLM) task, which randomly masks some of the tokens from the input, then predicts the masked word based on the context as the training objective.
We focus on the MLM task as it is a prevalent pre-training task for many standard models such as BERT \cite{devlin-etal-2019-bert} and RoBERTa \cite{liu2019roberta} that serve as building blocks for many downstream tasks. 
Using examples from a set of pre-defined domains, we train a BERT model from scratch to fully control our experiment and isolate the effect of different domains. This is crucial since a pre-trained BERT model is already trained on multiple domains, hence hard to drive correct conclusions through our analysis from such a model. Moreover, recent studies \cite{magar-schwartz-2022-data,brown-etal-2020-language} showed the risk of exposure of large language models to test data in the pre-training phase, also known as \emph{data contamination}.

\paragraph{Model}
We use the BERT\textsubscript{BASE} \citep{devlin-etal-2019-bert} architecture for all of our experiments. We train two types of models: the experimental model $\mymodele{}$, trained on all five domains with the MLM objective, and the control model $\mymodelc{}_i$ for $i \in [5]$ trained on the $i$th domain.
We are particularly interested in the effect of two aspects on the model representation: model capacity and data size. 
We use the capacity of 100\% for BERT\textsubscript{BASE} size. BERT\textsubscript{BASE}  has 768-dimensional vectors for each layer, adding up to a total of 110M parameters.
We also experiment with a reduced model capacity of 75\%, 50\%, 25\%, and 10\% by reducing the dimension of the hidden layers.
We follow \citet{devlin-etal-2019-bert} design choices, e.g., 12 layers with 12 attention heads per layer. We set the base training data size (100\%) for $\mymodele{}$ to be 50K, composed of 10K reviews per domain. Each $\mymodelc{}_i$ is trained on single domain data containing 10K reviews. $\mymodele{}$ and $\mymodelc{}_i$ \emph{share all the examples of domain $i$}. To study the effect of data size on model representation, we take subsets from the data split and create smaller datasets: a 10\% split and a 50\% split. We also create a 200\% split to simulate the case with abundant data.
We provide additional details about our training procedure in Appendix \ref{app:details_exp}.

%% file: results.tex
\section{Experiments and Results}
\label{section:results}
\ignore{
\begin{table*}
\small
\begin{tabular}{|l|p{0.43\textwidth}|p{0.43\textwidth}|}
\hline
\textbf{RQ1} & How does the similarity between $\mymodele{}$ and $\mymodelc{}_i$ evolve over training? & $\mymodele{}$ stores more domain-specific information in lower layers
than in deeper layers throughout the training procedure.\\ \hline
\textbf{RQ2} & How do data size and model capacity affect domain-specific encoding in $\ell_0$ and $\ell_{12}$? &   As the capacity of $\mymodele{}$ increases, more domain-specific
information is stored in $\ell_0$, and less in $\ell_{12}$. Increasing the data size results in less domain-specific information stored for both layers. \\ \hline
\textbf{RQ3} & To what extent does $\mymodele{}$ encode domain-specific information for domain-specific words? & As the capacity of $\mymodele{}$ increases, it stores more domain information for domain-specific words, for both $\ell_0$ and $\ell_{12}$. \\ \hline
\textbf{RQ4} &Do the observed trends manifest in the models' behavior? & We found evidence for the observed trends using the MLM task predictions for $\ell_{12}$ and the k-nearest neighbors analysis for $\ell_0$.\\
\hline
\end{tabular}

\caption{Our main findings are organized into the different research questions. See text for the transition between the questions and more elaborated explanations.\label{table:findings}}

\end{table*}

}

\begin{figure*}[ht]
     \centering
     \begin{subfigure}[b]{0.3\textwidth}
         \centering
         \includegraphics[width=\textwidth]{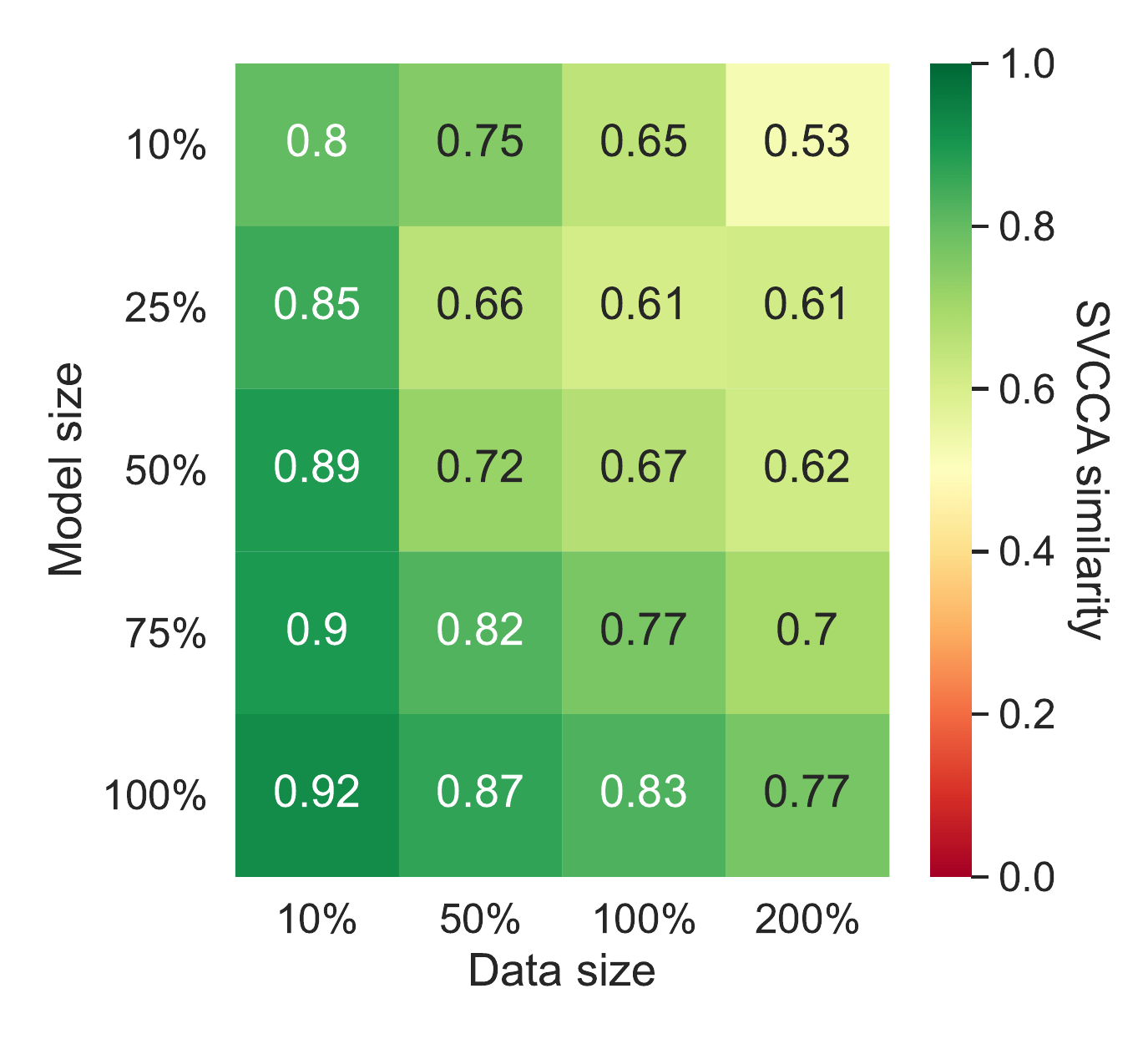}
         \caption{$\mymodelc{}_{Books}$: $\ell_0$}
         \label{fig:svcca-l0-books}
     \end{subfigure}
     \hfill
     \begin{subfigure}[b]{0.3\textwidth}
         \centering
         \includegraphics[width=\textwidth]{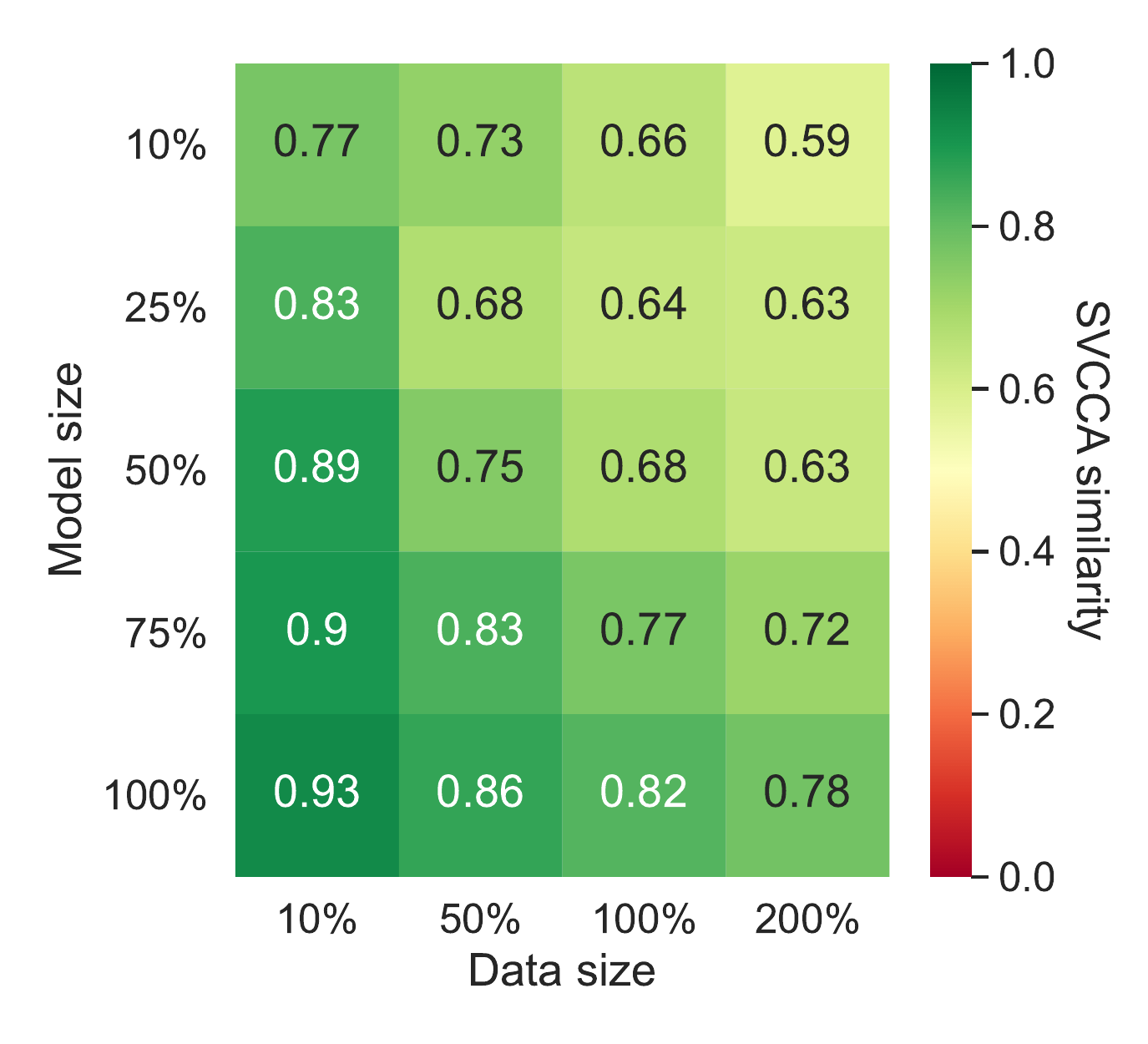}
         \caption{$\mymodelc{}_{Clothing}$: $\ell_0$}
         \label{fig:svcca-l0-clothing}
     \end{subfigure}
     \hfill
     \begin{subfigure}[b]{0.3\textwidth}
         \centering
         \includegraphics[width=\textwidth]{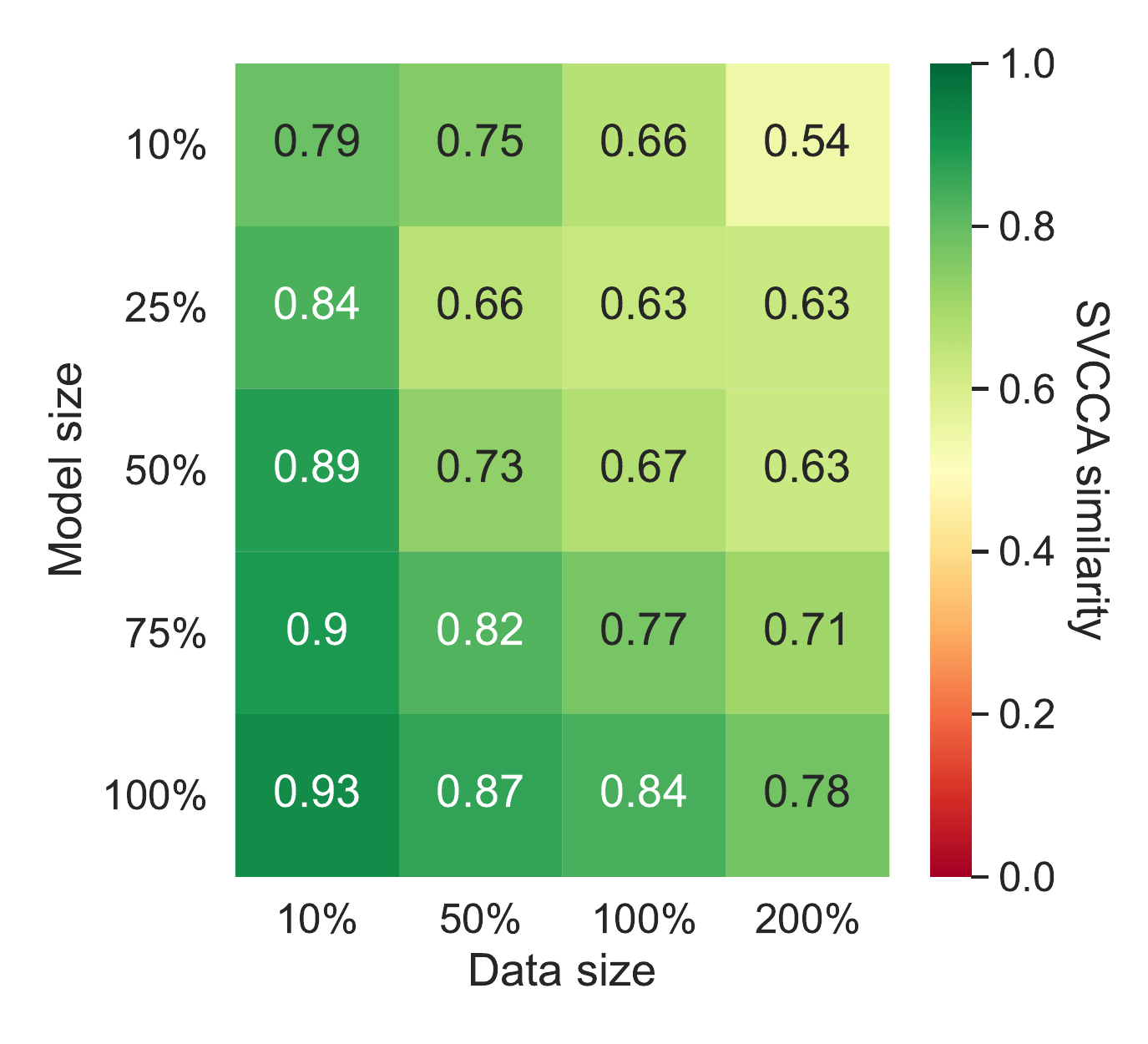}
         \caption{$\mymodelc{}_{Electronics}$: $\ell_0$}
         \label{fig:svcca-l0-electronics}
     \end{subfigure}
     \hfill
     \begin{subfigure}[b]{0.3\textwidth}
         \centering
         \includegraphics[width=\textwidth]{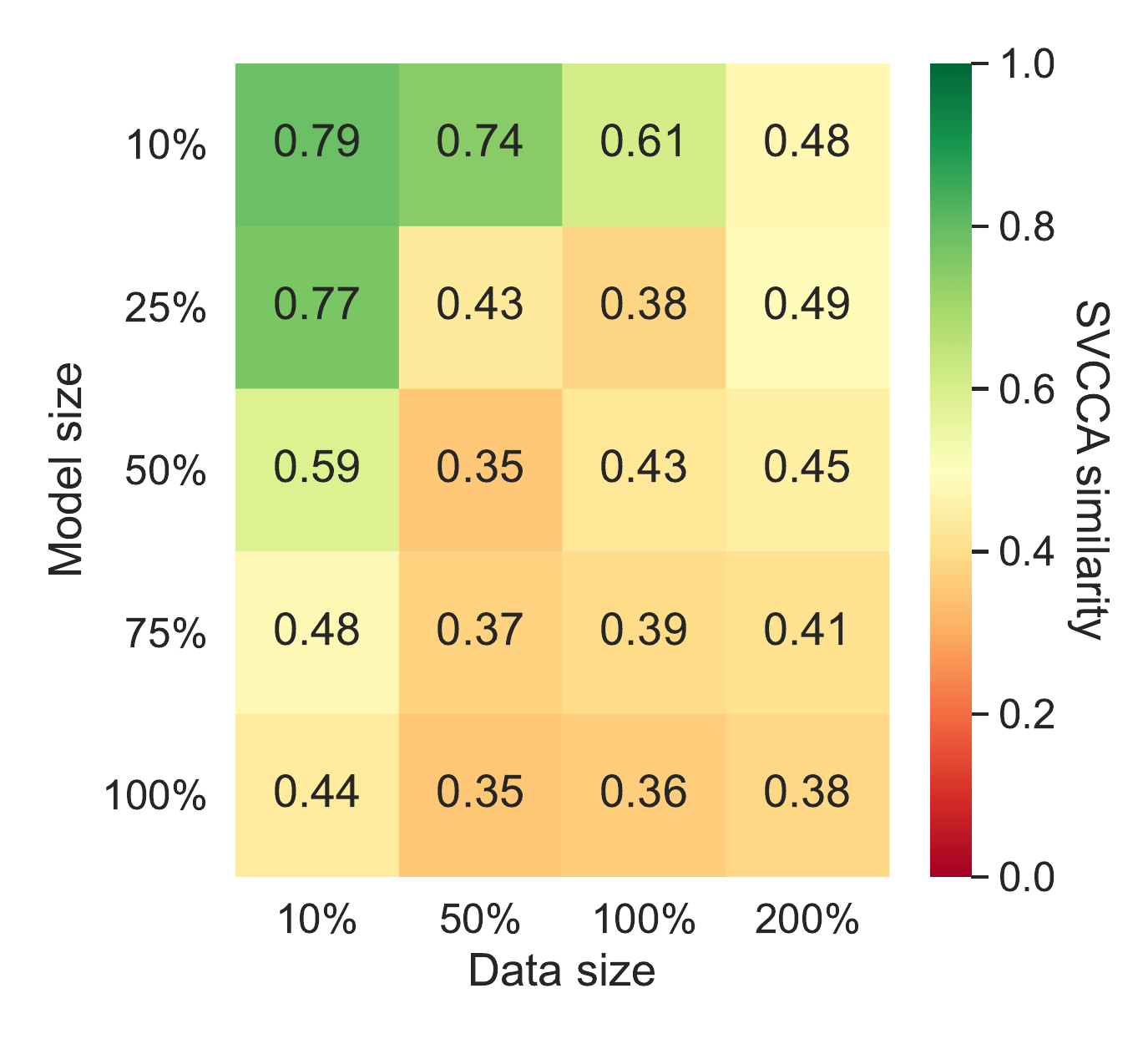}
         \caption{$\mymodelc{}_{Books}$: $\ell_{12}$}
         \label{fig:svcca-l12-books}
     \end{subfigure}
     \hfill
     \begin{subfigure}[b]{0.3\textwidth}
         \centering
         \includegraphics[width=\textwidth]{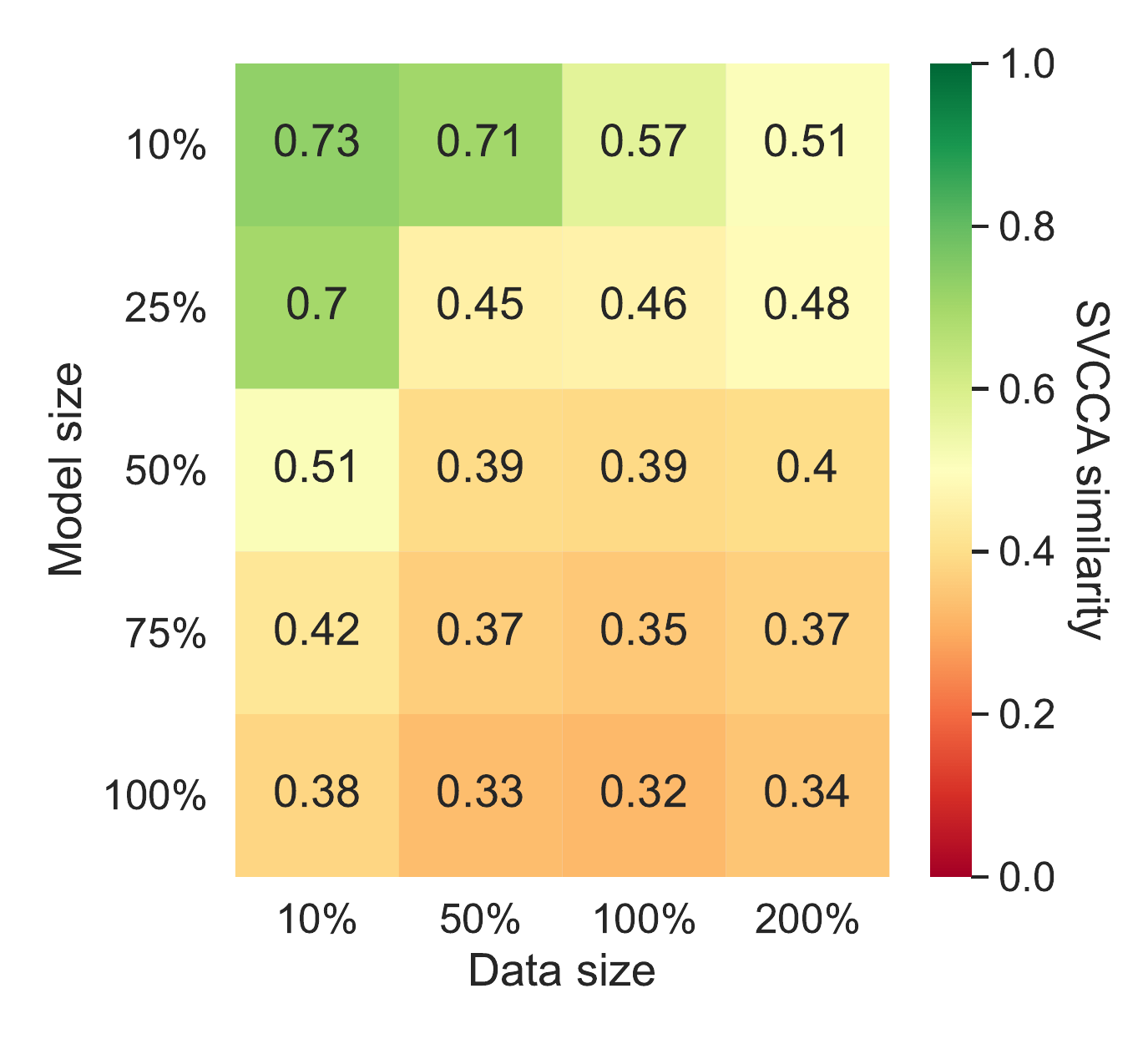}
         \caption{$\mymodelc{}_{Clothing}$: $\ell_{12}$}
         \label{fig:svcca-l12-clothing}
     \end{subfigure}
     \hfill
     \begin{subfigure}[b]{0.3\textwidth}
         \centering
         \includegraphics[width=\textwidth]{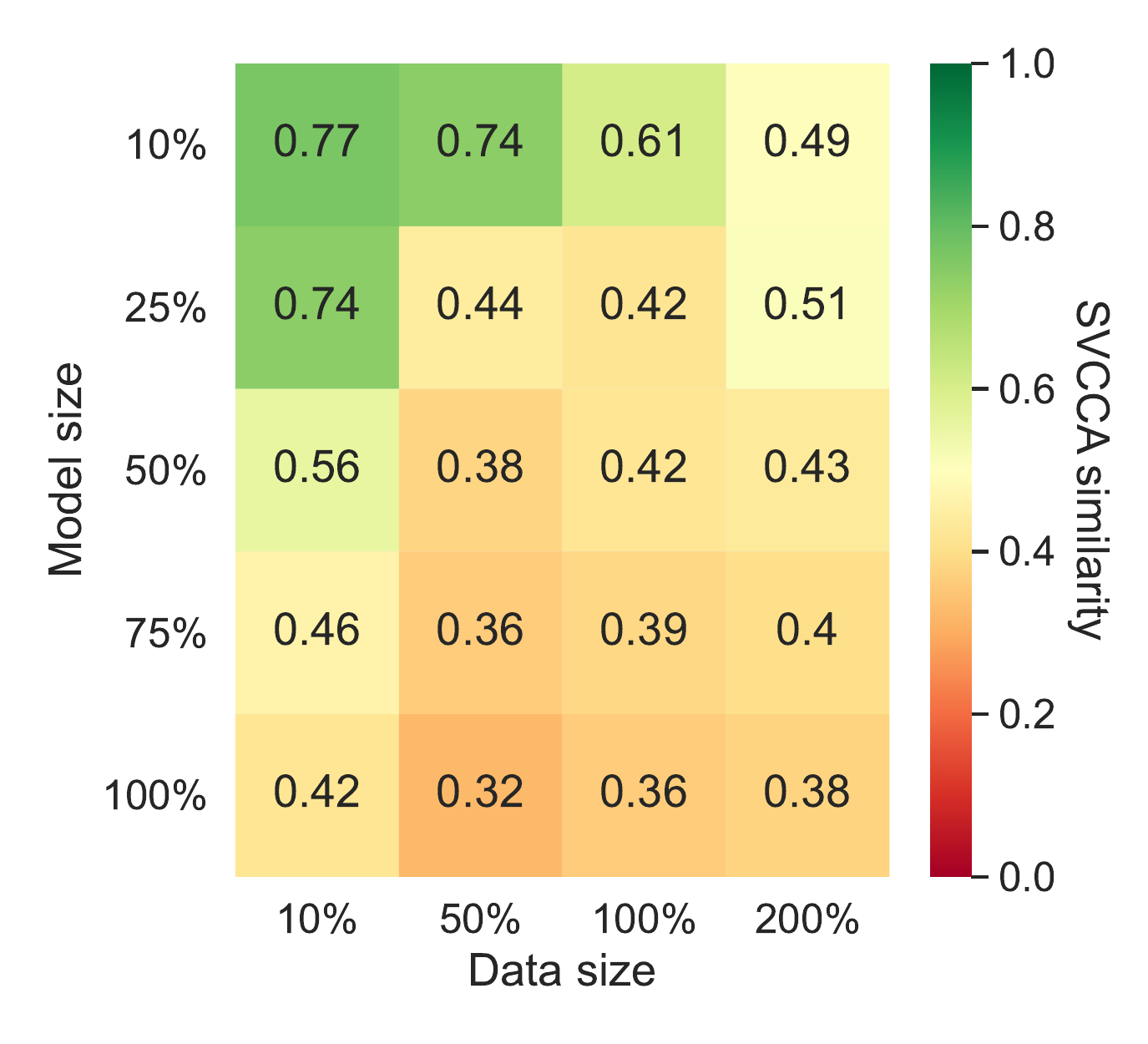}
         \caption{$\mymodelc{}_{Electronics}$: $\ell_{12}$}
         \label{fig:svcca-l12-electronics}
     \end{subfigure}
\caption{The SVCCA scores between $\mymodele{}$ and different $\mymodelc{}_i$s for different data sizes and model capacities. We only display for three domains here, and we provide the rest in Appendix~\ref{app:rq2}. The top row presents the results for the embedding layer $\ell_0$, and the bottom row presents them for the last layer $\ell_{12}$.}
\label{fig:svcca-three-domains}
\end{figure*}

\begin{figure}[ht]
    \centering
    \includegraphics[width=\linewidth]{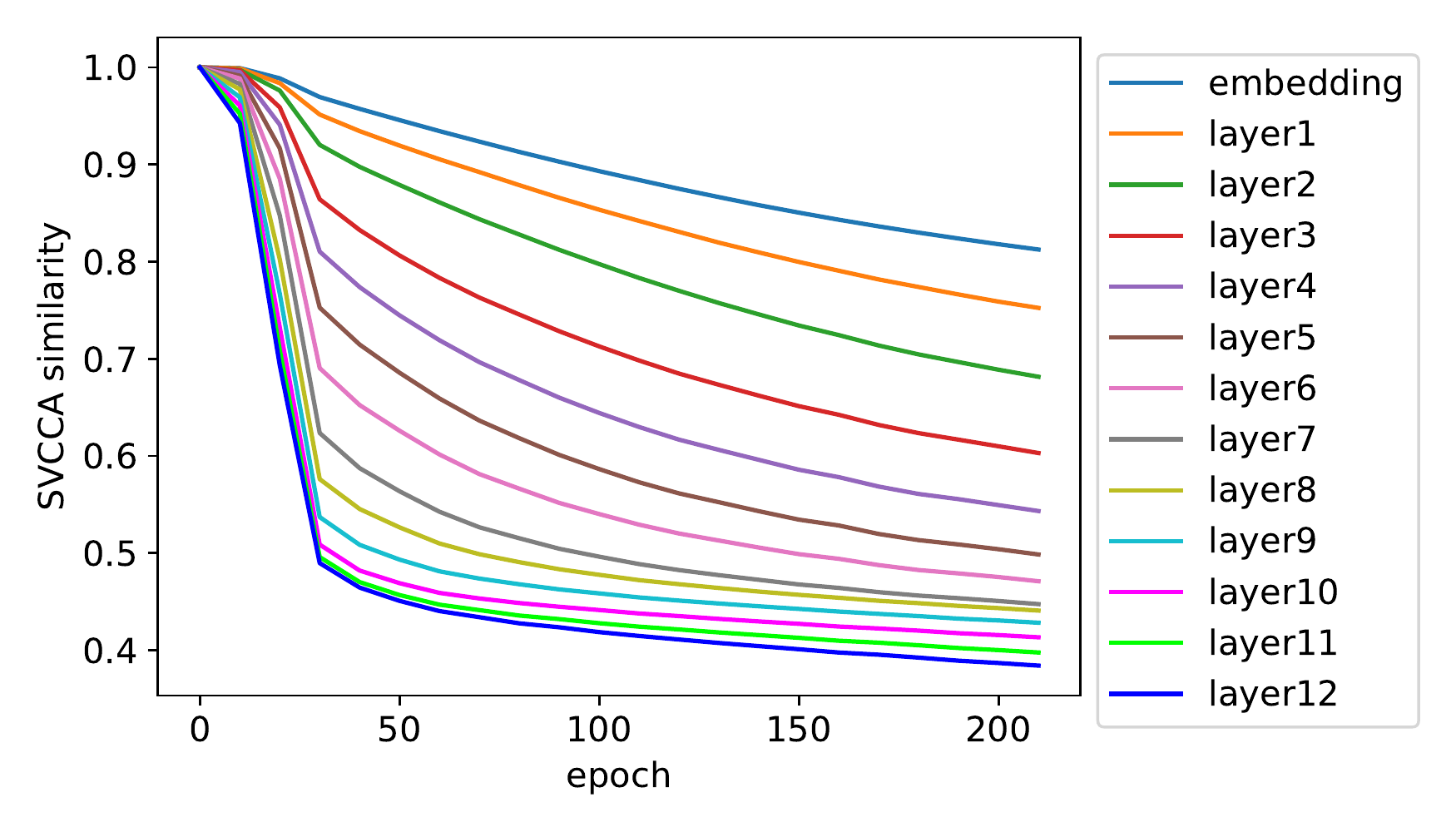}
    \caption{Training dynamics for all layers between $\mymodele{}$ and $\mymodelc{}_{Books}$. Here both model and data size are 100\%.}
    \label{fig:training_dynamics_all_layers_books}
\end{figure}

Our research questions (RQs) examine how domain-specific information is encoded in $\mymodele{}$ by calculating its SVCCA score with $\mymodelc{}_i$ for a specific $i$.
For a given domain, we use a held-out test set for getting the experimental and control model representations as an input for the SVCCA method. 
Intuitively, a high SVCCA score between $\mymodele{}$ and $\mymodelc{}_i$ indicates $\mymodele{}$ stores domain-specific information for domain $i$, as $\mymodelc{}_i$ was train solely on data from domain $i$. A low SVCCA score between $\mymodele{}$ and $\mymodelc{}_i$ could mean one of two things: a) $\mymodele{}$ can generalize to data from $d_i$ without explicitly storing domain-specific information about it, or b)  $\mymodele{}$ can not store information about $\mymodelc{}_i$, as a result of, for example, lack of model capacity. The way to distinguish between the two is subjective and depends on whether one finds $\mymodele{}$ performance when applied to data from $d_i$ to be satisfactory. This paper analyzes how information is stored at the model layers. As we inspect highly complex models consisting of multiple layers, it is challenging to determine to what extent a certain layer contributes to a model's overall performance. For those reasons, when comparing equivalent layers of different models, we focus on the amount of domain-specific information encoded in $\mymodele{}$ for a given layer.
With these preliminaries in mind, we are now ready to ask the following research questions:


\paragraph{RQ1: How does the similarity between the corresponding layers in $\mymodele{}$ and $\mymodelc{}$ evolve over training?}
We perform an iterative comparison between the $\mymodele{}$ and $\mymodelc{}_i$ for each $i \in [5]$.
After each epoch, we calculate the SVCCA score between corresponding layers of the models, i.e., layer $j$ of $\mymodele{}$ is
compared to layer $j$ of $\mymodelc{}_i$. As $\mymodele{}$ is trained on more data points than $\mymodelc{}_i$, and both use the same batch size, for any given epoch, $\mymodele{}$ had more weights' updates than $\mymodelc{}_i$. More precisely, after the $k$th epoch, $\mymodelc{}_i$ and $\mymodele{}$  had completed $k$ passes on data points from $d_i$, but $\mymodele{}$ used additional data points from the rest of the domains. We choose this alignment to examine the effect of the additional training data drawn from other domains.

Figure~\ref{fig:training_dynamics_all_layers_books} presents the training dynamics analysis for the \texttt{Books} domain
(we denote the \texttt{Books} control model as $\mymodelc{}_{Books}$). We include training dynamics analyses of
other control models and domains in Appendix~\ref{app:rq1}, as they demonstrate similar trends.
Since both $\mymodelc{}_{Books}$ and $\mymodele{}$ are initialized with the same weights, the initial SVCCA score is $1$ for
all layers before training. We observe that as training
progresses, the SVCCA values of higher layers (closer to the output) consistently become lower compared
to the first layer. The order of SVCCA values is almost perfectly preserved with respect to the order of the layers in the network.
The separation is higher for lower layers, with higher layers receiving similar SVCCA values.
This is evidence that \emph{$\mymodele{}$ stores more domain-specific information in lower layers
than in deeper layers throughout the training procedure.} \newcite{singh-etal-2019-bert}, who researched the nature of multilingual models, observed a similar pattern of dissimilarity in deeper layers for multilingual model representations of parallel sentences in different languages.

The alignment between the similarity of the layer pairs ($\mymodele{}$ and $\mymodelc{}$) and their depth also exists for models with random weights.
It can be partially attributed to the mathematical artifact of decreasing
correlation values for layers that are deeper because of the use of nonlinear activation units. To see to what extent this artifact plays a role in this alignment, we created ten models with random weights (no training, so there is no longer an experimental/control distinction) and calculated SVCCA between all 45 pairs for the first and last layers. We discovered that the mean difference between SVCCA scores of the first layer comparison and the last layer comparison is 0.139 (with a standard deviation of 0.001 over 45 pairs). In Figure~\ref{fig:training_dynamics_all_layers_books}, the difference is much larger when comparing the control model to the experimental model (0.428), indicating that the difference in layer SVCCA score cannot be only attributed to the mathematical artifact of increasing depth with more nonlinear activation.
We still note that one should exercise caution when using linear methods, such as SVCCA, to analyze nonlinear processes.


The observed training dynamics motivates us to focus on the embedding layer ($\ell_0$) and final layer ($\ell_{12}$) for the
rest of our analysis, as they serve as a lower bound ($\ell_0$) and an upper bound ($\ell_{12}$) with respect to the SVCCA scores of 
$\mymodelc{}_i$ and $\mymodele{}$ throughout the training process.
In addition, those layers have interesting attributes that we would like to explore. $\ell_0$, a non-contextualized word embeddings layer,
is known for encoding mainly lexical information \cite{de-vries-etal-2020-whats,vulic-etal-2020-probing}. The highly contextualized $\ell_{12}$ is fed
directly to the masked word classifier, thus playing a significant role in the MLM task.
Our interest in the fully-trained models  leads us to the following question:

\begin{figure*}[htb]
     \centering
     \begin{subfigure}[b]{0.3\textwidth}
         \centering
         \includegraphics[width=\textwidth]{figures/svcca/general_vs_control_domain_Books_layer0.pdf}
         \caption{$\ell_0$: all words}
         \label{fig:books-l0-all}
     \end{subfigure}
     \hfill
     \begin{subfigure}[b]{0.3\textwidth}
         \centering
         \includegraphics[width=\textwidth]{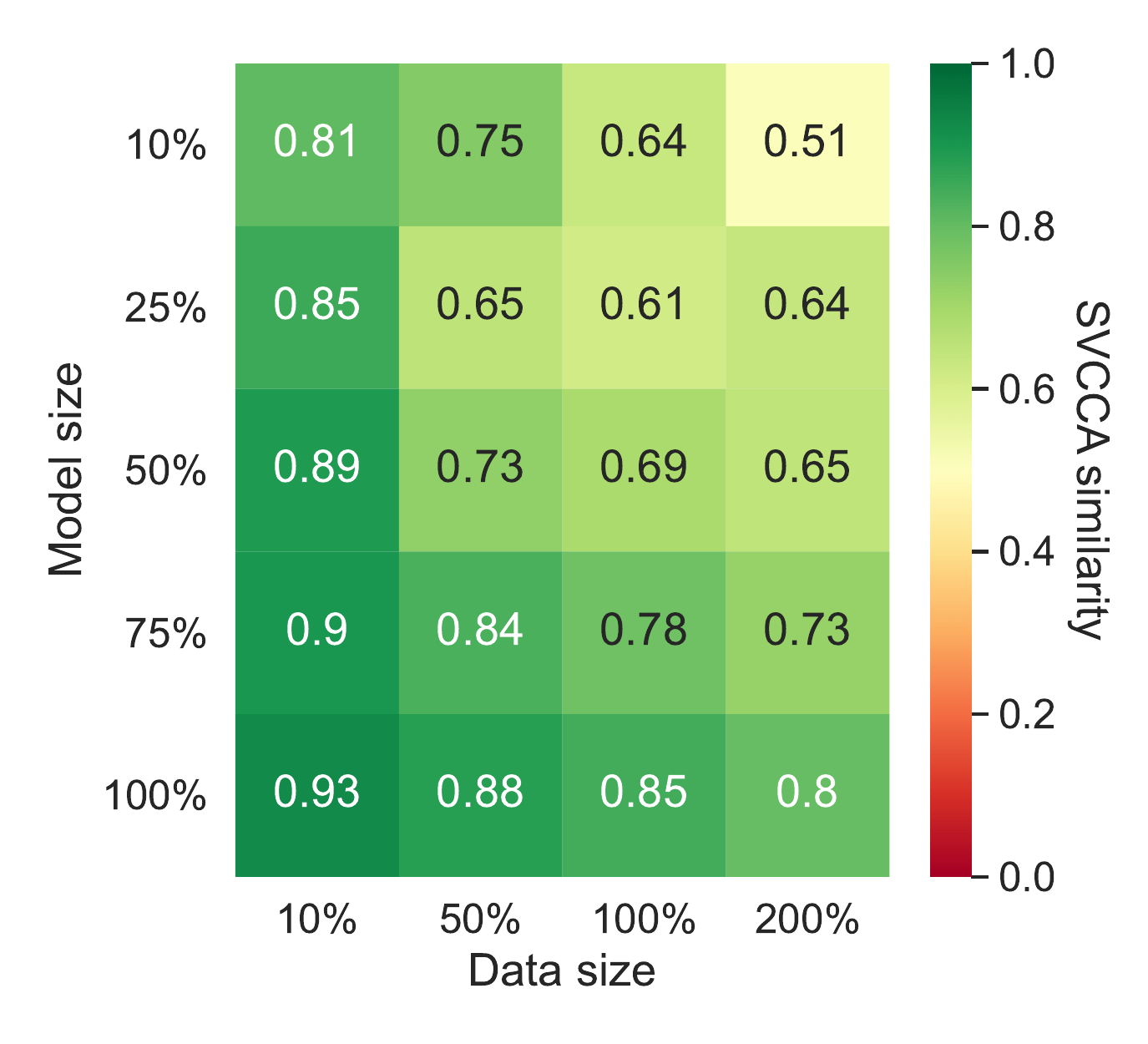}
         \caption{$\ell_0$: general words}
         \label{fig:books-l0-general}
     \end{subfigure}
     \hfill
     \begin{subfigure}[b]{0.3\textwidth}
         \centering
         \includegraphics[width=\textwidth]{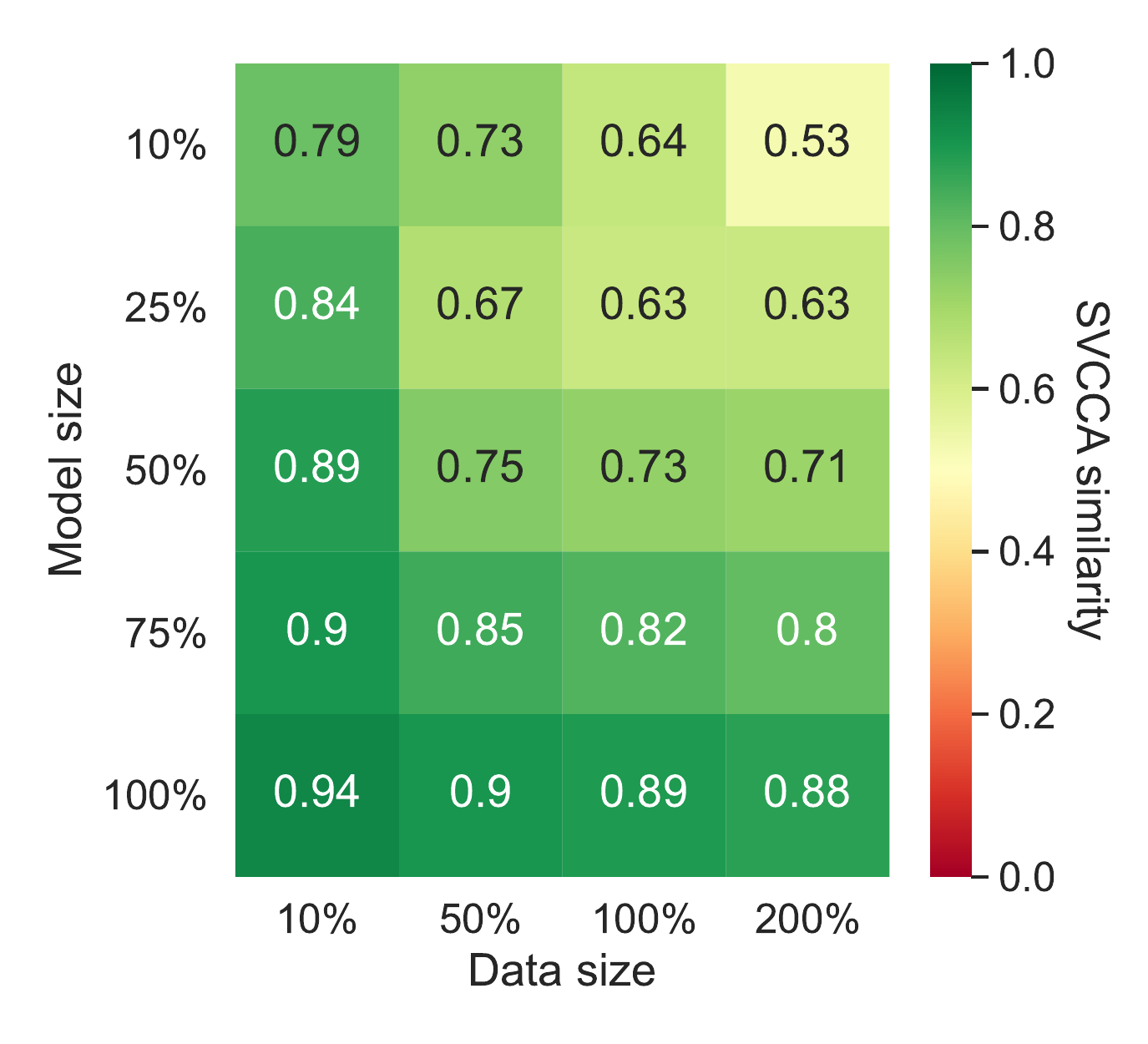}
         \caption{$\ell_0$: domain-specific words}
         \label{fig:books-l0-specific}
     \end{subfigure}
     \hfill
     \begin{subfigure}[b]{0.3\textwidth}
         \centering
         \includegraphics[width=\textwidth]{figures/svcca/general_vs_control_domain_Books_layer12.pdf}
         \caption{$\ell_{12}$: all words}
         \label{fig:books-l12-all}
     \end{subfigure}
     \hfill
     \begin{subfigure}[b]{0.3\textwidth}
         \centering
         \includegraphics[width=\textwidth]{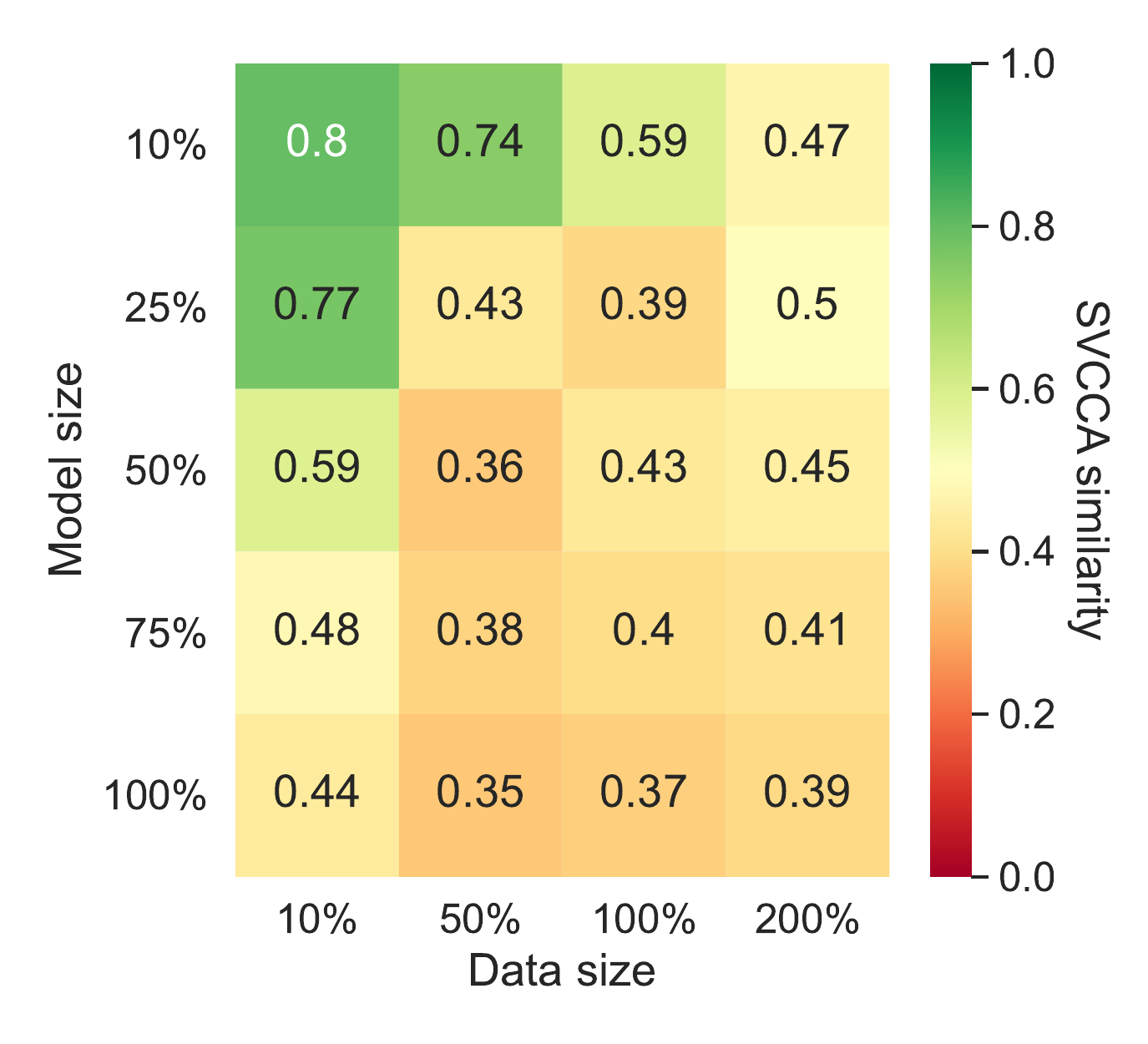}
         \caption{$\ell_{12}$: general words}
         \label{fig:books-l12-general}
     \end{subfigure}
     \hfill
     \begin{subfigure}[b]{0.3\textwidth}
         \centering
         \includegraphics[width=\textwidth]{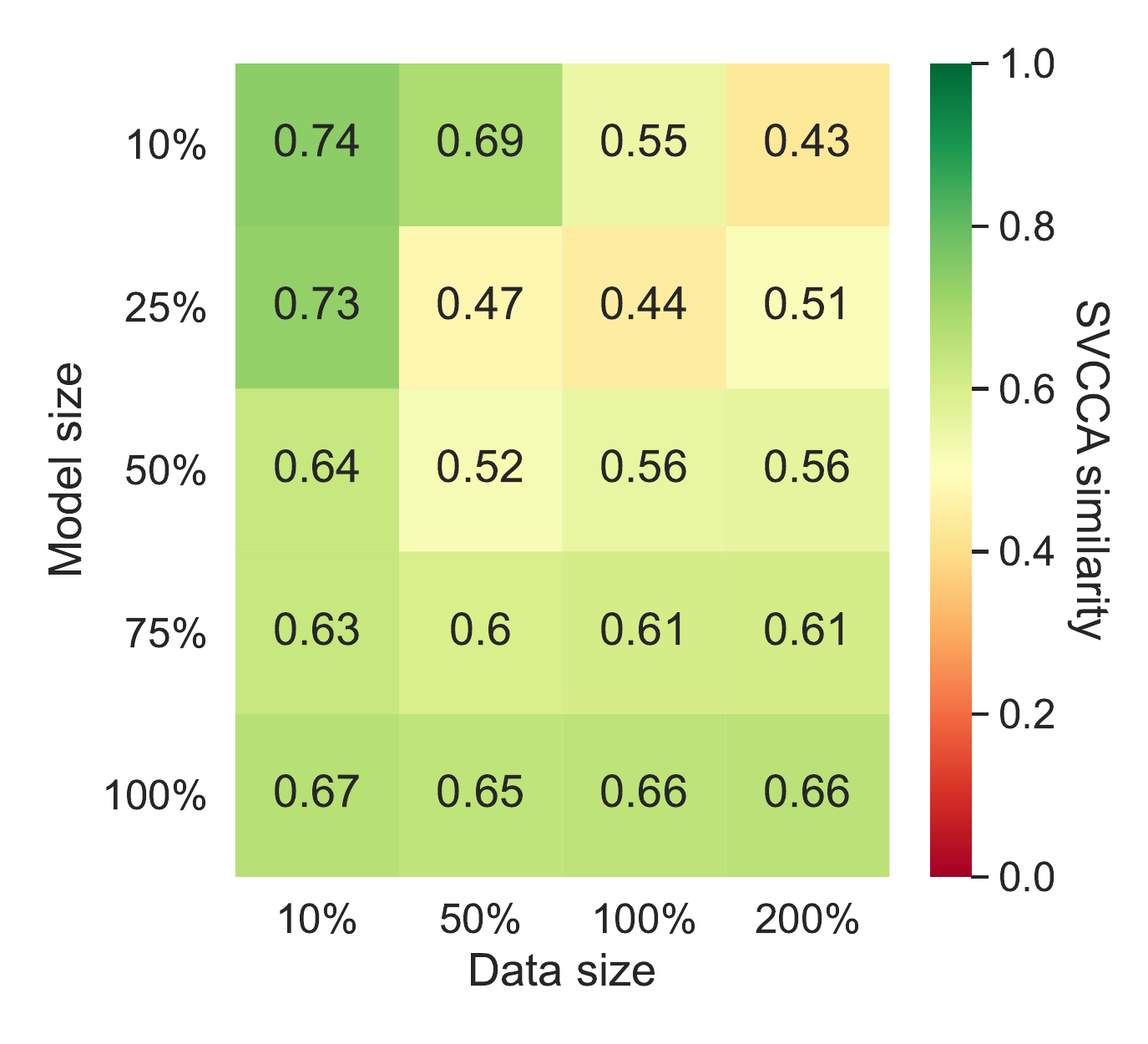}
         \caption{$\ell_{12}$: domain-specific words}
         \label{fig:books-l12-specific}
     \end{subfigure}
\caption{The SVCCA score between $\mymodele{}$ and $\mymodelc{}_{Books}$ for different subsets of tokens. The top row presents the results for the embedding layer $\ell_{0}$, and the bottom row presents them for the last layer $\ell_{12}$. }
\label{fig:books-all}
\end{figure*}

\paragraph{RQ2: How do data size and model capacity affect domain encoding in $\ell_0$ and $\ell_{12}$?}
To answer this question, we measure the SVCCA score between variants of $\mymodele{}$ and their corresponding $\mymodelc{}_i$ for different domains.
The variants differ with respect to two parameters, data size and model capacity. 

Figure \ref{fig:svcca-three-domains} presents our results. We observe training the model on larger datasets decreases the SVCCA scores across all model capacities and domains for both $\ell_0$ and $\ell_{12}$. 
For each data point we add to the control model, we add $d$ data points to the general model, where $d-1$ out of them belong to other domains.
This means while we keep a constant ratio between the number of datapoints for the domains, the absolute gap between a given domain and the rest of the domains is
growing for larger data sizes.
This might explain why adding more data points increase $\mymodele{}$ and $\mymodelc{}$ divergence.

A possible explanation for these trends might be how we define domains.
The Amazon reviews dataset is divided by product categories which can be seen
as lexical domains (see \S~\ref{sec:experimental_setup}). More precisely, all the domains share a similar
structure and writing style of Amazon product reviews. The differences lie in the vocabulary of each domain.
We hypothesize that the \emph{$\mymodele{}$  uses the increased capacity to keep more domain-specific
information in $\ell_0$, where the lexical information is kept and diverges from $\mymodelc{}$ in $\ell_{12}$, where
the highly contextualized representations are stored}. As we hypothesize that our domains differ mostly with
respect to their vocabularies, we refine the mentioned above experiment by raising the following research question:

\paragraph{RQ3: To what extent does $\mymodele{}$ encode domain-specific information for domain-specific words?}
\begin{table*}[h]
\small
    \begin{subtable}[h]{0.45\textwidth}
        \centering
        \begin{tabular}{|cc|cc|}
            \hline
            \multicolumn{2}{|c|}{m=50\%} & \multicolumn{2}{c|}{m=100\%} \\ \hline
            \multicolumn{1}{|c|}{$\mymodele{}$} & $\mymodelc{}_i$ & \multicolumn{1}{c|}{$\mymodele{}$} & $\mymodelc{}_i$     \\ \hline
            \multicolumn{1}{|c|}{blackberry}        & proxy       & \multicolumn{1}{c|}{linux} & mac        \\ 
            \multicolumn{1}{|c|}{linux}        & linux       & \multicolumn{1}{c|}{mac}        & linux        \\ 
            \multicolumn{1}{|c|}{biologist}        & peer      & \multicolumn{1}{c|}{blackberry} & computers        \\ 
            \multicolumn{1}{|c|}{viking}        & windows       & \multicolumn{1}{c|}{vista}    & windows        \\ 
            \multicolumn{1}{|c|}{samsung}        & servers       & \multicolumn{1}{c|}{xp}        & xp        \\ \hline
            \end{tabular}
       \caption{5-nearest neighbors for the domain-specific word \textbf{Macintosh} with $i$=\texttt{Electronics}. }
       \label{tab:exp1}
    \end{subtable}
    \hfill
    \begin{subtable}[h]{0.45\textwidth}
        \centering
        \begin{tabular}{|cc|cc|}
            \hline
            \multicolumn{2}{|c|}{m=50\%}    & \multicolumn{2}{c|}{m=100\%}           \\ \hline
            \multicolumn{1}{|c|}{$\mymodele{}$} & $\mymodelc{}_i$ & \multicolumn{1}{c|}{$\mymodele{}$} & $\mymodelc{}_i$ \\ \hline
            \multicolumn{1}{|c|}{functioning}   & riding & \multicolumn{1}{c|}{functioning}   & functioning             \\ 
            \multicolumn{1}{|c|}{work}      & running         & \multicolumn{1}{c|}{work}    & repair             \\ 
            \multicolumn{1}{|c|}{worked}       & work         & \multicolumn{1}{c|}{worked}  & work             \\ 
            \multicolumn{1}{|c|}{playing}       & walking     & \multicolumn{1}{c|}{looking}    & riding             \\ 
            \multicolumn{1}{|c|}{responding}    & cleaning     & \multicolumn{1}{c|}{works}      & looking             \\ \hline
        \end{tabular}
       \caption{5-nearest neighbors for the general word \textbf{working} with $i$=\texttt{Home and Kitchen}.}
       \label{tab:exp2}
    \end{subtable}
    \hfill
    \begin{subtable}[h]{0.45\textwidth}
        \centering
        \begin{tabular}{|cc|cc|}
            \hline
            \multicolumn{2}{|c|}{m=50\%}            & \multicolumn{2}{c|}{m=100\%}           \\ \hline
            \multicolumn{1}{|c|}{$\mymodele{}$} & $\mymodelc{}_i$ & \multicolumn{1}{c|}{$\mymodele{}$} & $\mymodelc{}_i$ \\ \hline
            \multicolumn{1}{|c|}{networks}   & connections    & \multicolumn{1}{c|}{routers}  & router             \\ 
            \multicolumn{1}{|c|}{phones}   & networks    & \multicolumn{1}{c|}{products}  & networks             \\ 
            \multicolumn{1}{|c|}{devices}   & ports    & \multicolumn{1}{c|}{systems}  & connections             \\ 
            \multicolumn{1}{|c|}{problems}   & computers    & \multicolumn{1}{c|}{mice}  & computers             \\ 
            \multicolumn{1}{|c|}{models}   & cables    & \multicolumn{1}{c|}{connections}  & products             \\ \hline
        \end{tabular}
       \caption{\textit{Other wired and wireless [MASK] I had never had this problem.} The masked word is a domain-specific word \textbf{routers} with $i$=\texttt{Electronics}.}
       \label{tab:exp3}
    \end{subtable}
    \hfill
    \begin{subtable}[h]{0.45\textwidth}
        \centering
        \begin{tabular}{|cc|cc|}
            \hline
            \multicolumn{2}{|c|}{m=50\%}            & \multicolumn{2}{c|}{m=100\%}           \\ \hline
            \multicolumn{1}{|c|}{$\mymodele{}$} & $\mymodelc{}_i$ & \multicolumn{1}{c|}{$\mymodele{}$} & $\mymodelc{}_i$ \\ \hline
            \multicolumn{1}{|c|}{away}  & apart & \multicolumn{1}{c|}{apart}   & aside             \\ 
            \multicolumn{1}{|c|}{apart} & off  & \multicolumn{1}{c|}{flat}   & apart             \\ 
            \multicolumn{1}{|c|}{aside} & away  & \multicolumn{1}{c|}{short}    & down             \\ 
            \multicolumn{1}{|c|}{downhill} & downhill   & \multicolumn{1}{c|}{out}   & back             \\ 
            \multicolumn{1}{|c|}{asleep}  & asleep   & \multicolumn{1}{c|}{off}   & along            \\ \hline
        \end{tabular}
       \caption{\textit{Sadly, those hopes began to fall [MASK] shortly after I finished the Prologue.} The masked word is a general word \textbf{apart} with $i$=\texttt{Books}.}
       \label{tab:exp4}
    \end{subtable}
    \caption{(a) and (b) are the 5-nearest neighbors using the embedding layer weights. (c) and (d) are model predictions using last layer representations. m denotes model capacity. All models here use a data size of 100\%. }
    \label{tab:predictions_using_embedding}
\end{table*}

To shed light on the domains' lexical nature, we inspect the patterns of domain-specific and general words.
Domain-specific words need to appear with at least 20 reviews in the domain in hand and no more than 10 reviews in total for the rest of the other domains. General words must appear in at least 20 reviews in each domain. Those definitions are often used in domain adaptation works to describe domain discrepancy and find adaptable features \cite{blitzer-etal-2007-biographies, ziser-reichart-2017-neural}.  We provide some examples of domain-specific and
general words in Appendix~\ref{app:example_specific_general_words}.
It is noteworthy that the union of the domain-specific and general words is not the complete vocabulary. To calculate the SVCCA scores for a subset of words, we first apply SVD  to all inputs. Then we use the corresponding representations of the subset tokens to calculate the CCA similarity. 

Figure~\ref{fig:books-all} presents our results for the \texttt{Books} domain.\footnote{The rest of the domains exhibit similar patterns. We provide all results in Appendix~\ref{app:rq3}} We present the \texttt{Books} domain analysis for all the words taken from RQ2 for reference (on the left-hand side of the figure).
We observe high SVCCA scores for domain-specific words for $\ell_{12}$. For large data sizes (100\% and 200\%), the trends of domain-specific words are opposite
to the ones of RQ2, i.e., $\mymodele{}$ uses the additional capacity to encode more domain-specific information. This indicates that as
model capacity increases, $\mymodele{}$ can capture similar information
to $\mymodelc{}_{Books}$ for domain-specific words. This justifies the construction of large language models, mixing
multiple subpopulations, as it demonstrates that \emph{if the $\mymodele{}$ model has large enough capacity, it separately
creates representations for the different subpopulations that are similar to $\mymodelc{}_i$ model, which is a
specialized model for a given domain}. Domain-specific words and their representations are crucial for the success of many NLP tasks, for example, Named Entity Recognition \citep{rocktaschel-etal-2013-wbi,shang-etal-2018-learning,gu-etal-2021-domain}. We can see that the SVCCA scores for all the words and general words are almost identical.
These findings make us suspect that word frequency and domain specificity are strongly connected. Indeed, we find out that the average frequency for \texttt{Books} domain-specific words is 75 with a median of 43. For general words, the average is 7696, and the median is 1440, making general words the main factor in the SVCCA scores for all words.

Finally, we would like to ensure the patterns we observe throughout this paper affect the behavior of the model:

\paragraph{RQ4: Do the observed trends manifest in the models' behavior?}

\begin{figure*}
     \centering
     \begin{subfigure}[b]{0.3\textwidth}
         \centering
         \includegraphics[width=\textwidth]{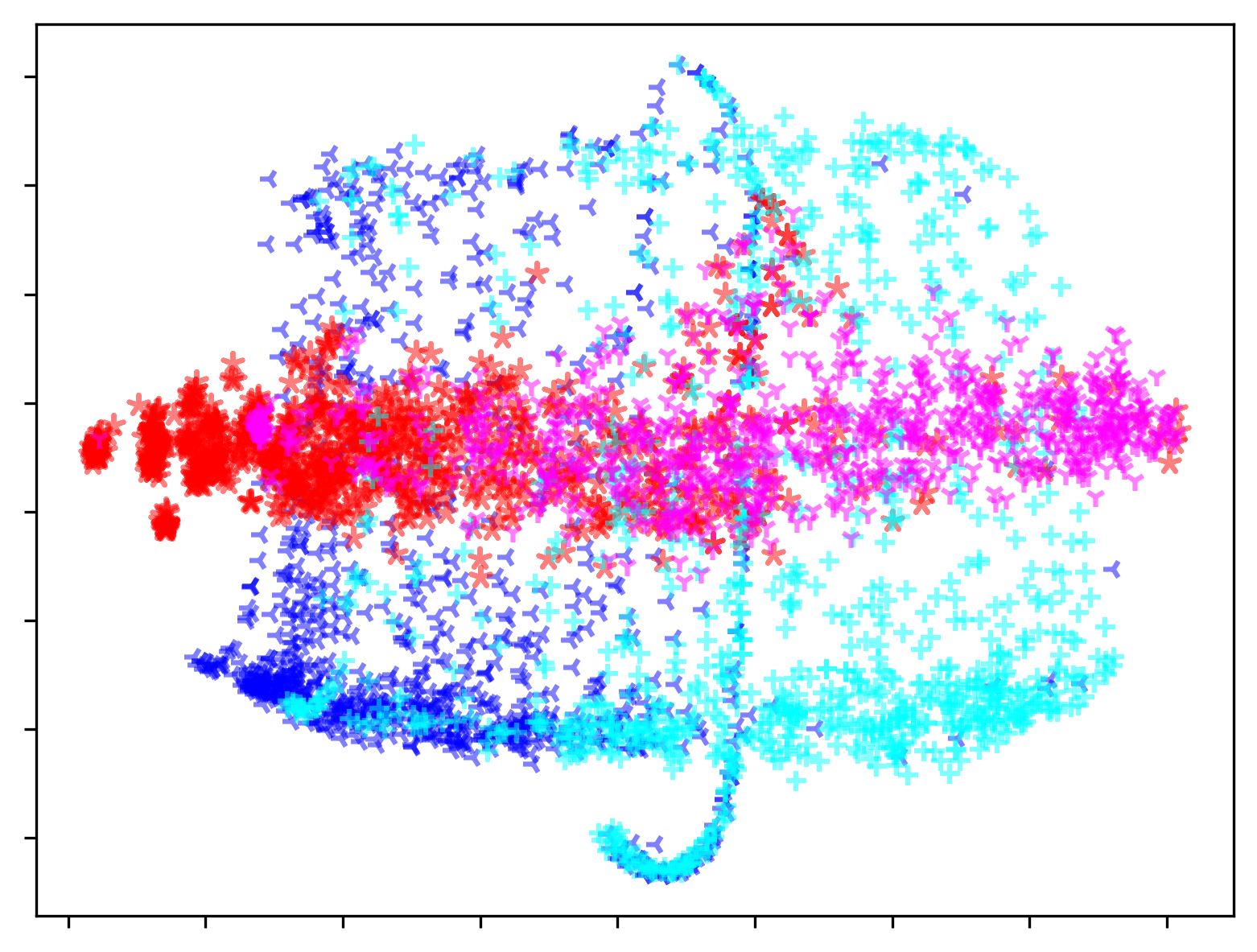}
         \caption{$\ell_0$: m=10\%}
         \label{fig:books-l0-pca2}
     \end{subfigure}
     \hfill
     \begin{subfigure}[b]{0.3\textwidth}
         \centering
         \includegraphics[width=\textwidth]{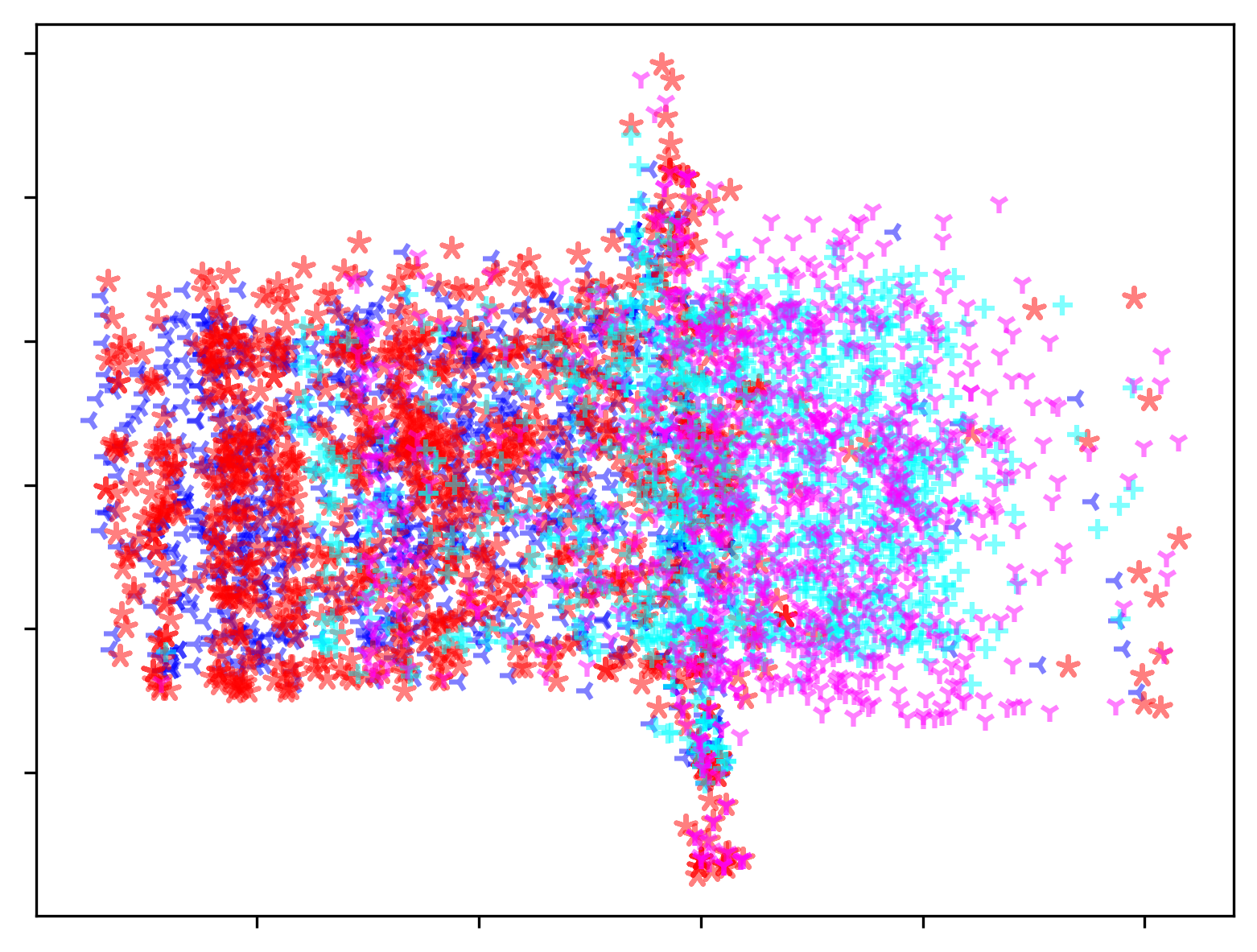}
         \caption{$\ell_0$: m=50\%}
         \label{fig:books-l0-pca3}
     \end{subfigure}
     \hfill
     \begin{subfigure}[b]{0.3\textwidth}
         \centering
         \includegraphics[width=\textwidth]{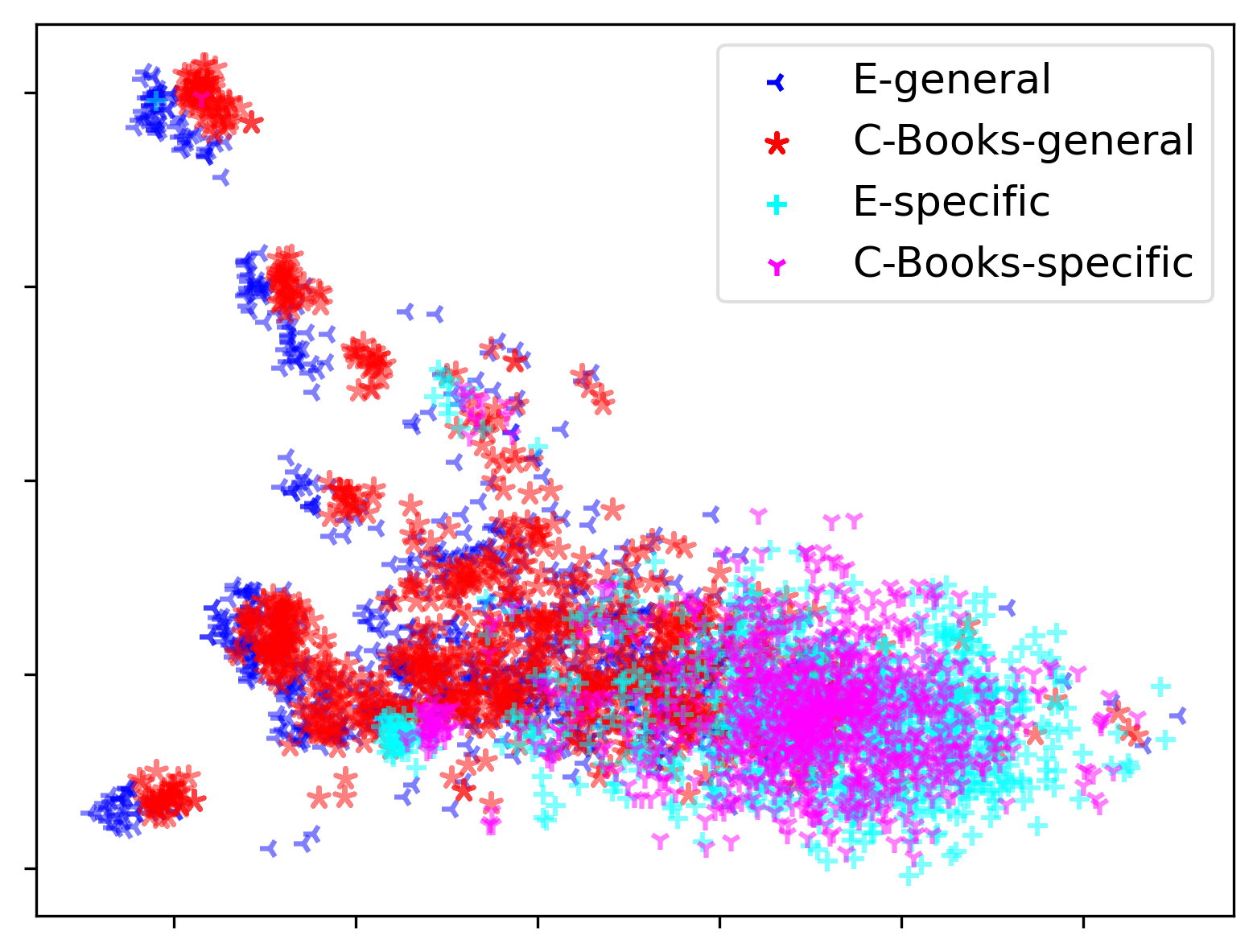}
         \caption{$\ell_0$: m=100\%}
         \label{fig:books-l0-pca4}
     \end{subfigure}
     \hfill
     \begin{subfigure}[b]{0.3\textwidth}
         \centering
         \includegraphics[width=\textwidth]{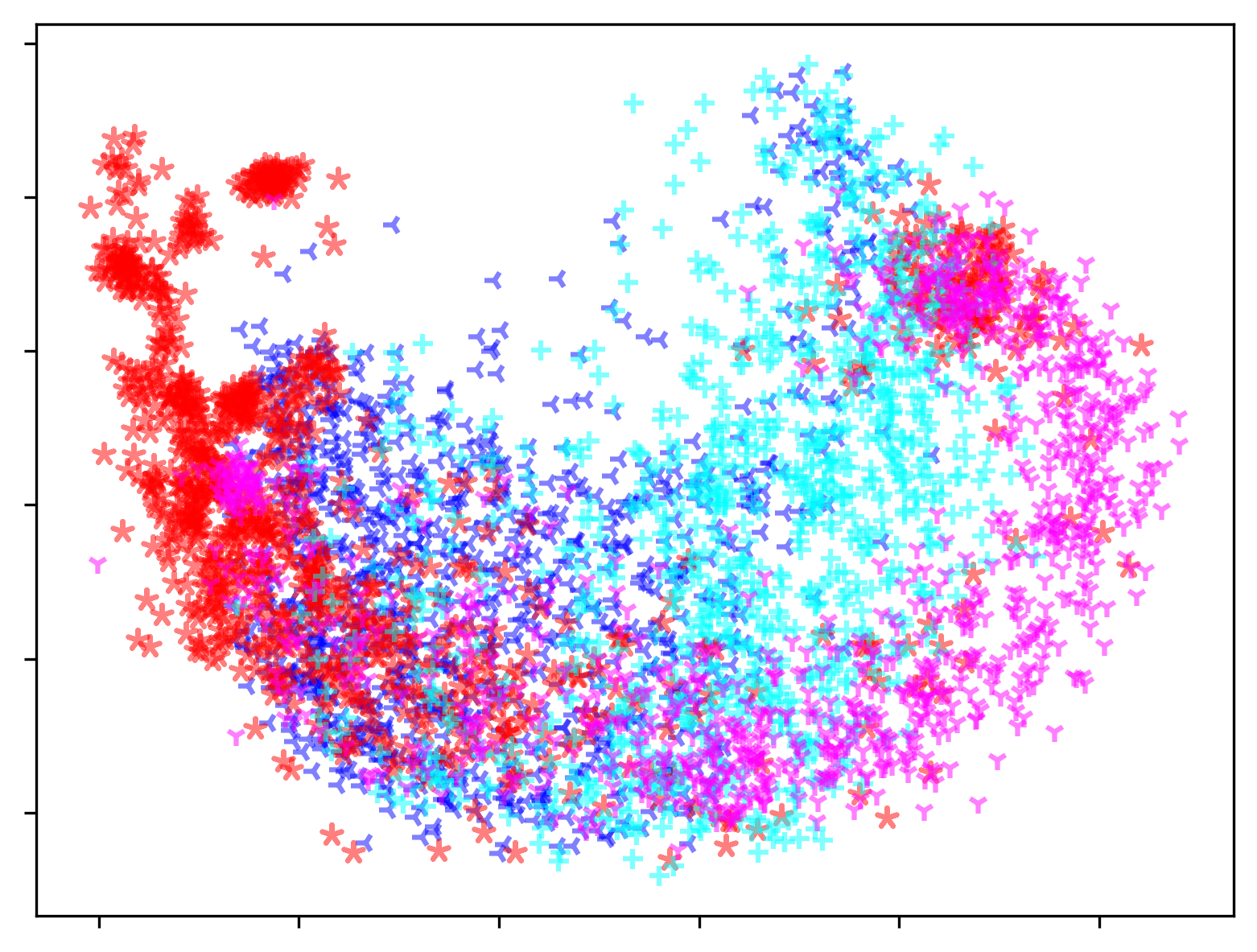}
         \caption{$\ell_{12}$: m=10\%}
         \label{fig:books-l12-pca2}
     \end{subfigure}
     \hfill
     \begin{subfigure}[b]{0.3\textwidth}
         \centering
         \includegraphics[width=\textwidth]{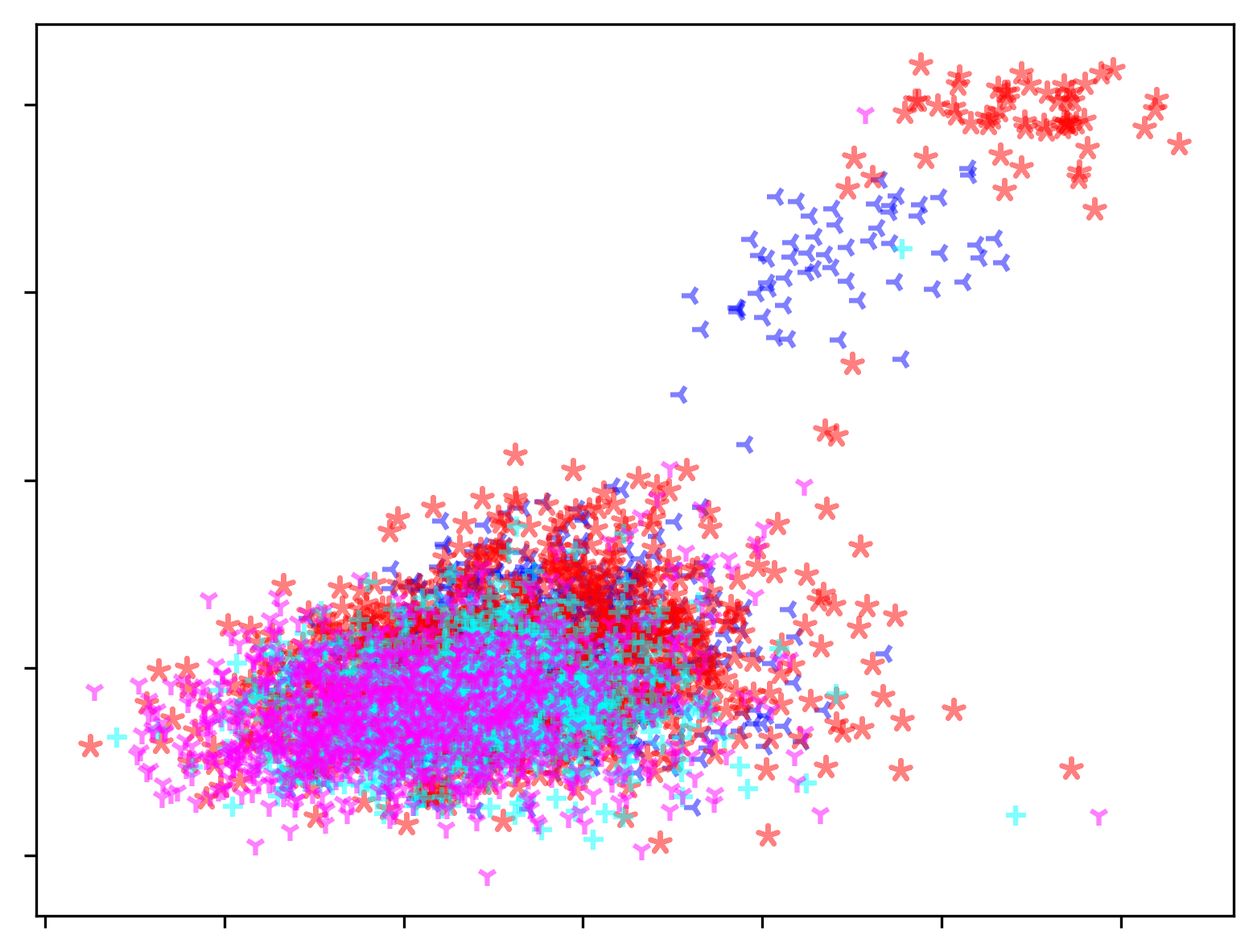}
         \caption{$\ell_{12}$: m=50\%}
         \label{fig:books-l12-pca3}
     \end{subfigure}
     \hfill
     \begin{subfigure}[b]{0.3\textwidth}
         \centering
         \includegraphics[width=\textwidth]{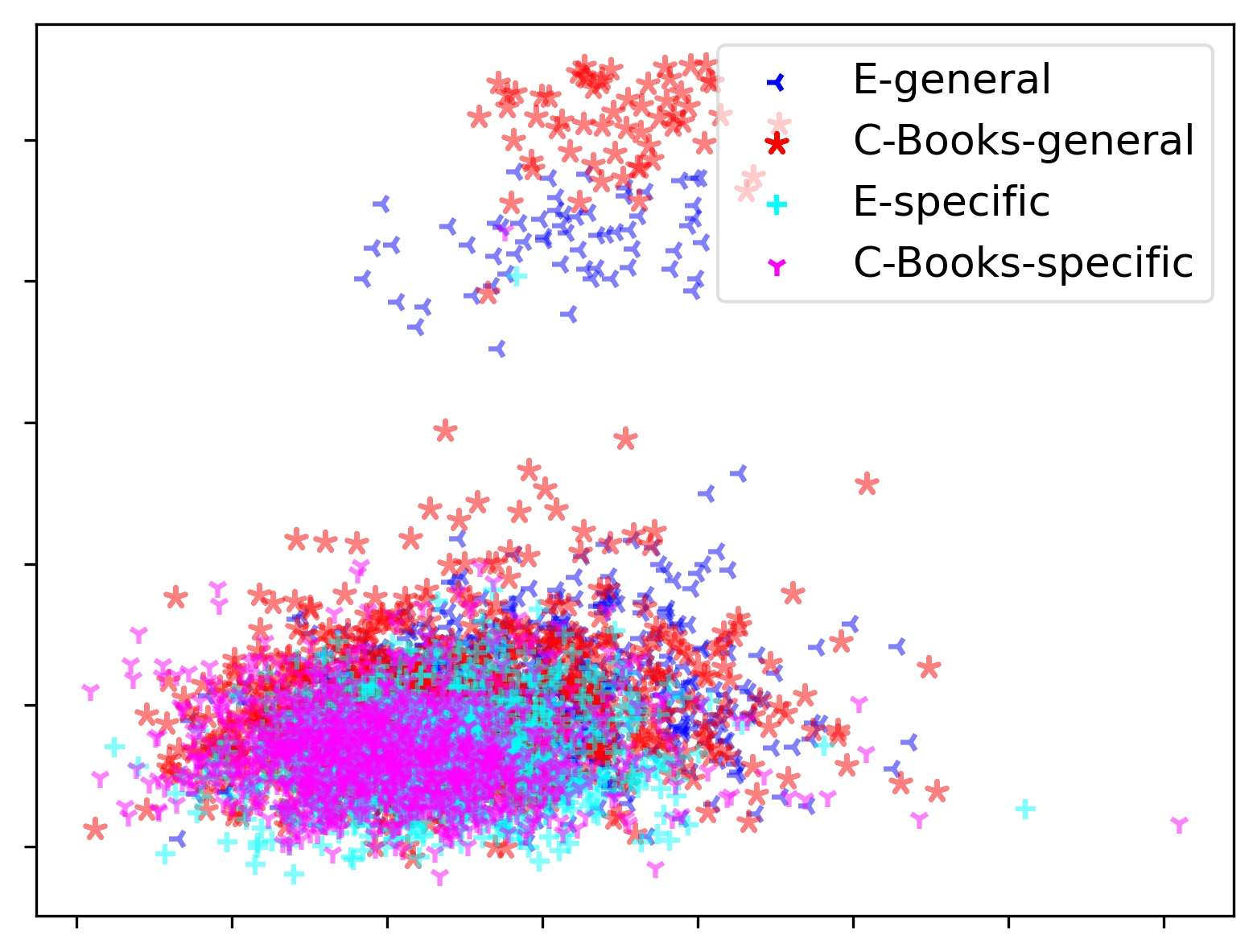}
         \caption{$\ell_{12}$: m=100\%}
         \label{fig:books-l12-pca4}
     \end{subfigure}
\caption{Visualization for $\ell_0$ and $\ell_{12}$ representations for $\mymodele{}$ and $\mymodelc{}_{Books}$. We use colors (blue/cyan for $\mymodele{}$ and red/magenta for $\mymodelc{}_{Books}$) to separate representations for generals and domain-specific words. m denotes model capacity. All models here use a data size of 100\%.}
\label{fig:books-l0-l12-pca-all}
\end{figure*}

We conducted two qualitative analyses to understand better if the models' behavior expresses our findings. 
For the first analysis, we compare MLM predictions of $\mymodele{}$ and $\mymodelc{}$ to check whether higher SVCCA values are associated with similar word predictions.
For $\ell_0$, we calculate the k-nearest neighbors of the word embeddings for a given word as a proxy to make predictions. For $\ell_{12}$, we follow the standard procedure by feeding the last layer representation to the final MLM classifier in BERT.
Table \ref{tab:predictions_using_embedding} presents our analyses. We can see that for $\ell_0$, as we increase
the model capacity, we get more similar predictions for both domain-specific and general words. This finding agrees with the trend in Figure~\ref{fig:svcca-three-domains} that higher model capacity is associated with higher SVCCA similarity for $\ell_0$. For $\ell_{12}$, we can see that as model capacity increases, predictions for the general word becomes inconsistent, whereas, for domain-specific words, it is the opposite. This finding also agrees with our findings in RQ2 and RQ3, in which we observe the $\ell_{12}$ SVCCA values are decreasing for general words as we increase the model capacity and decrease for domain-specific words. We provide additional examples in Appendix~\ref{app:rq4}.

For the second analysis, we employ principal component analysis (PCA) to reduce the dimension of general and domain-specific representations for $\ell_0$ and $\ell_{12}$ for both $\mymodele{}$ and $\mymodelc{}_{Books}$. We provide visualizations in Figure~\ref{fig:books-l0-l12-pca-all}. We can see that as model capacity increases, $\ell_0$ representations of both general and domain-specific words from $\mymodele{}$ and $\mymodelc{}_{Books}$ are aligned to a similar subspace. Additionally, $\ell_{12}$ representations of general words and domain-specific words for both models exhibit opposite behavior: domain-specific words are more aligned with increasing model capacity while general words start to detach. All of these agree with our findings in corresponding SVCCA scores trends in Figure~\ref{fig:books-all}.
Even though we did not explicitly examine the relations between general and specific words in our work, we can observe that general and domain-specific word representations form different clusters in both models. Those clusters are more separated in $\ell_0$ than in $\ell_{12}$, suggesting that models use their increased capacity to keep more domain-specific information in $\ell_0$.

\paragraph{WikiSum results}
Due to the lack of computational resources required, we only validate our main findings, namely, RQ2 and RQ3, using WikiSum. We present the results in Appendix~\ref{app:wikisum}. We choose \texttt{Health} domain as it is the largest domain of this dataset. We observe that the trend in SVCCA scores across different scenarios on WikiSum is generally the same as those on Amazon Reviews, demonstrating that our findings are consistent.

%% file: related_works.tex
\section{Related Work}

\paragraph{Analyzing neural representations}
\citet{raghu-etal-2017-svcca} proposed SVCCA for comparing representations for the same data points from different layers and networks invariant to an affine transform.
They also discovered that lower layers in a multi-layer neural network converge more quickly to their final representations in contrast to higher layers. Building off of SVCCA, \citet{morcos-etal-2018-insights} developed projection weighted CCA (PWCCA) using an aggregation technique. Using the SVCCA tool, \citet{saphra-lopez-2019-understanding} studied the learning dynamics of neural language models by probing the evolution of syntactic, semantic, and topic representations across time and models.  \citet{kudugunta-etal-2019-investigating} used SVCCA to understand massively multilingual neural machine translation representations over 100 languages. Their major findings are that encoder representations of different languages form clusters based on their linguistic similarities. 

\paragraph{Diagnostic Classifiers}
Another prominent tool for analyzing learned representations is  diagnostic classifiers  \citep[DCs;][]{belinkov-etal-2017-neural, belinkov-etal-2017-evaluating, giulianelli-etal-2018-hood}. DCs measure the amount of information encoded in representations about a particular task by using them as input to a classifier, which is trained on the task in a supervised manner. DC users assume that the higher their performance for this task, the more task-specific information is encoded in the representations. While widely adopted, DCs have several pitfalls. For example, \citet{zhang-bowman-2018-language} showed that learning a classifier on top of random embeddings is often competitive and, in some cases, even better than doing so with representations taken from a pre-trained model when trained on enough data. \citet{saphra-lopez-2019-understanding} demonstrated that, unlike SVCCA, DCs showed a stable correlation between language models and target labels throughout training epochs, in contrast to the language models' immense improvement over time.

%% file: conclusion.tex
\section{Conclusions and Future Work}
We present a novel methodology based on subpopulation analysis which helps understand how sub-domains are represented in a multi-domain model.
Our findings show that neural models encode domain information differently in lower and upper layers and that larger models (in our case, $\mymodele{}$) tend to ``preserve a copy'' of small, more specialized models ($\mymodelc{}$).
Generally, we observe rapid model improvements in NLP tasks when model capacity and dataset size, the two dimensions we study, increase.
We encourage the research community to study the cause for these improvements from a
multi-domain angle (i.e., the ability to encode specific
information about many domains at once using the increased capacity).
In future work, we would like to apply our methodology to examine the behavior of multilingual, multitask, and multimodal models.

%% file: limitation.tex
\section*{Limitations}
Throughout this work, we use the BERT\textsubscript{BASE} model. While it is widely adopted in the NLP community, there are other more advanced models (such as BERT\textsubscript{LARGE}, RoBERTa and GPT3) that we do not experiment with due to a lack of resources. Given that the differences between models of the BERT family are mostly irrelevant to the way we conduct our experiments, we believe our results would generalize, at the very least, to this family of models.

In addition, we do not experiment with a large amount of training data for two reasons: a) We want to control for the domains from which we draw examples, and those have a size limitation, and b) Training many models on a large dataset is computationally expensive. Our multi-domain setup is comprised of five domains. We believe a higher number of domains should be considered for real-world scenarios.

To control our experiments, we train all models from scratch. For real-world scenarios, it would be harder to divide the training data into homogeneous and natural domains. While our proposed methodology can be easily adapted to different similarity measurement methods, we focus on SVCCA, which restricts us to linear correlations. In future work, we plan to investigate the nature of domains using non-linear techniques.

We identify domains through a common topic, and as a result, the shared lexical choices within the domain. This is the most common case for classifying domains, but we acknowledge that there are additional valuable ways to define domains. For example, domains could be separated based on writing style while still having a significant shared vocabulary (Amazon book reviews and Wikipedia articles about books).

%% file: appendix.tex
\section{Additional Details for Experiments}
\label{app:details_exp}
Here we provide some additional details for our experiments.

\paragraph{Training}

We set the validation data size for $\mymodele{}$ to be 10K, which is composed of 2K reviews from each domain. For validation set of each $\mymodelc{}_i$, we use the same 2K reviews used for $\mymodele{}$ from each domain. For consistency, we use the same validation set for all data sizes. We use a test set with 2.5K reviews for each domain. The same test set is fed to both $\mymodele{}$ and $\mymodelc{}_i$ across all model capacities and data sizes to obtain representations for subpopulation analysis.  When it is clear from the context which $\mymodelc{}_i$ for $i \in [5]$ we are referring to (and under which training regime), we will
use the simplification $\mymodelc{}$.

All models use the validation set cross-entropy loss to perform early stopping, and we train a model for a maximum of 500 epochs. 
We provide the validation loss (cross-entropy) for the $\mymodele{}$ model in Table~\ref{tab:validation_results}. From the results, we can see that for fixed data size,  model performance saturates when reaching model capacity of 100\%. Thus, unlike data size, we do not perform further experiments with model capacity larger than 100\%.

\begin{table}[h]
\centering
\begin{tabular}{|c|cccc|}
\hline
     & 10\%d & 50\%d & 100\%d & 200\%d \\ \hline
10\%m & 6.052 & 5.541 & 4.788 & 3.886 \\ 
25\%m & 5.764 & 3.257 & 2.745 & 2.354 \\ 
50\%m & 4.366 & 2.758 & 2.451 & 2.144 \\ 
75\%m & 4.017 & 2.781 & 2.435 & 2.149 \\ 
100\%m & 4.012 & 2.786 & 2.436 & 2.16  \\
\hline 
\end{tabular}
\caption{Validation cross-entropy loss on the experimental model for different model capacities and data sizes where m refers to model capacity and d refers to data size used to train the model.}
\label{tab:validation_results}
\end{table}

\begin{figure}[ht]
    \centering
    \includegraphics[width=\linewidth]{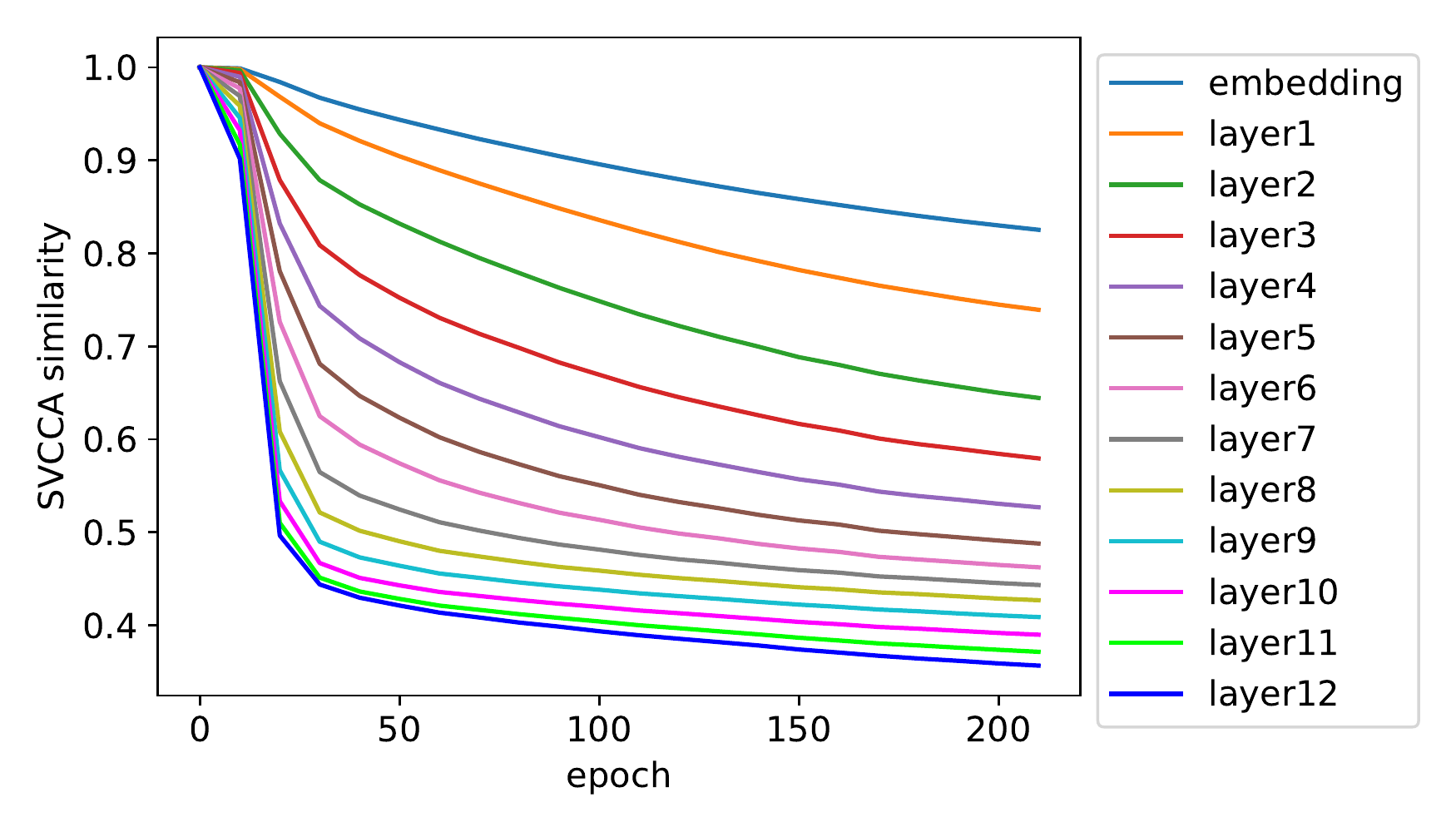}
    \caption{Training dynamics for all layers between $\mymodele{}$ and $\mymodelc{}_{Clothing}$. Here both model and data size are 100\%.}
    \label{fig:training_dynamics_all_layers_clothing}
\end{figure}

\begin{figure}[ht]
    \centering
    \includegraphics[width=\linewidth]{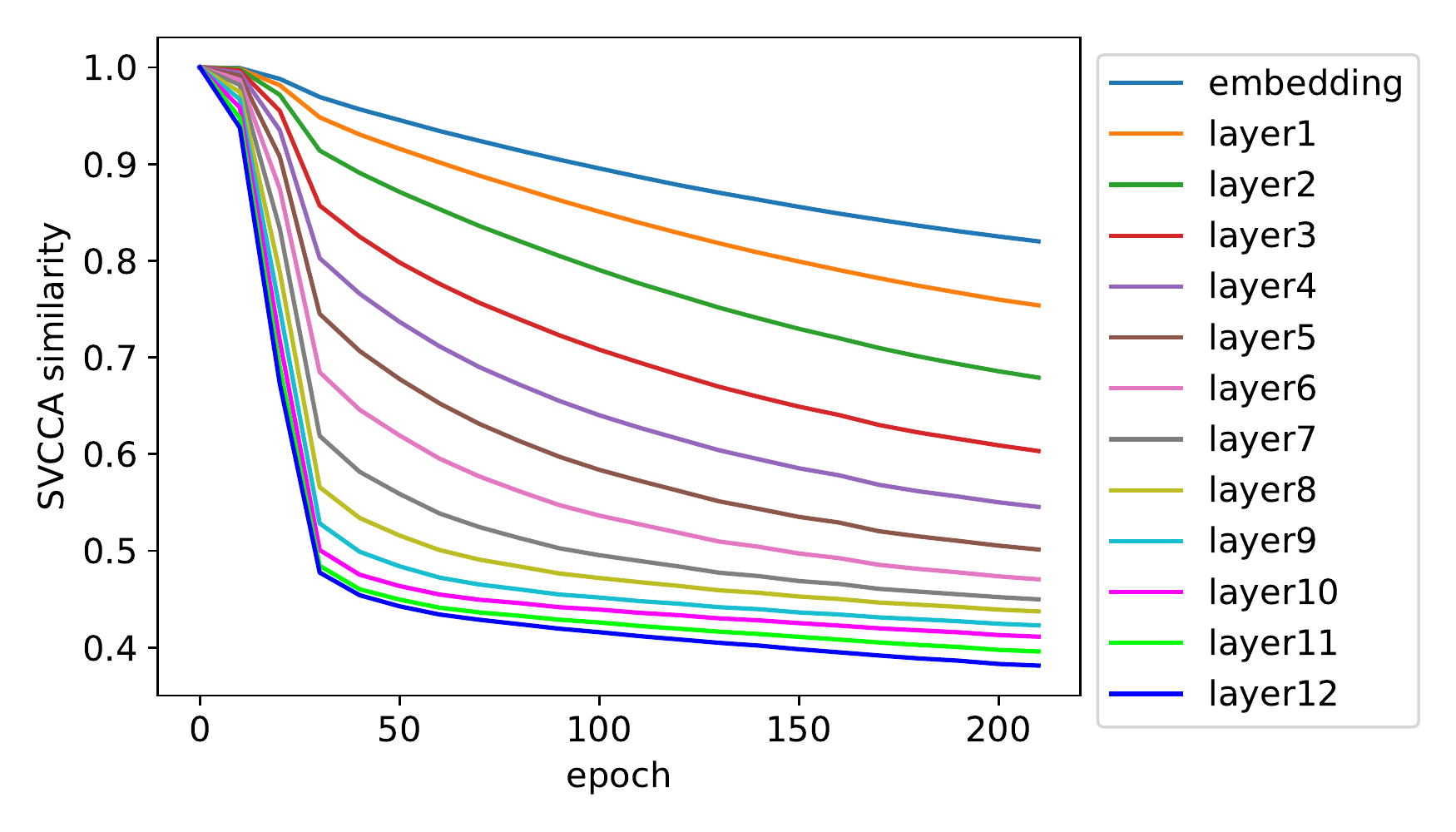}
    \caption{Training dynamics for all layers between $\mymodele{}$ and $\mymodelc{}_{Electronics}$. Here both model and data size are 100\%.}
    \label{fig:training_dynamics_all_layers_electronics}
\end{figure}

\begin{figure}[ht]
    \centering
    \includegraphics[width=\linewidth]{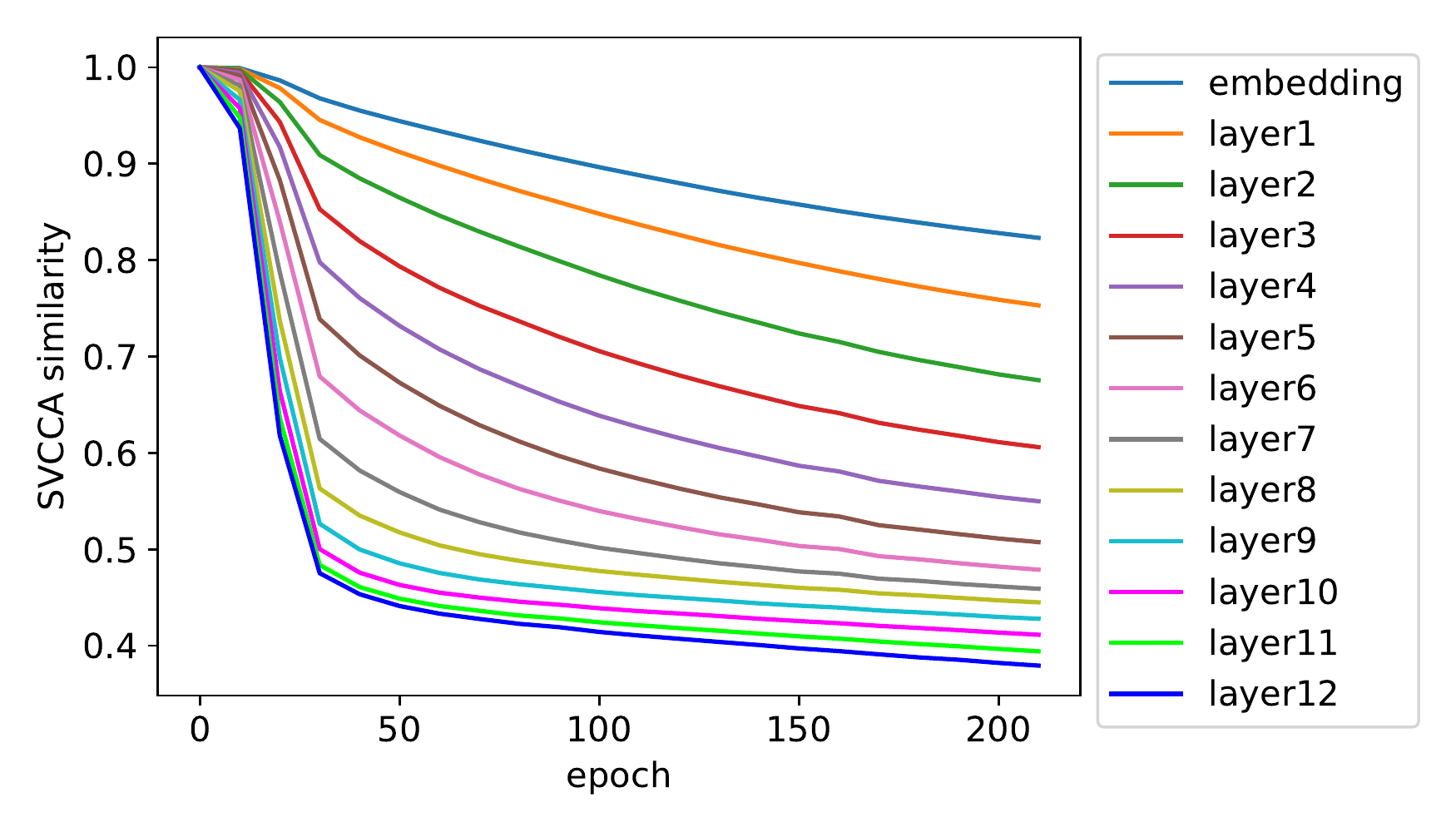}
    \caption{Training dynamics for all layers between $\mymodele{}$ and $\mymodelc{}_{Home}$. Here both model and data size are 100\%. }
    \label{fig:training_dynamics_all_layers_home}
\end{figure}

\begin{figure}[ht]
    \centering
    \includegraphics[width=\linewidth]{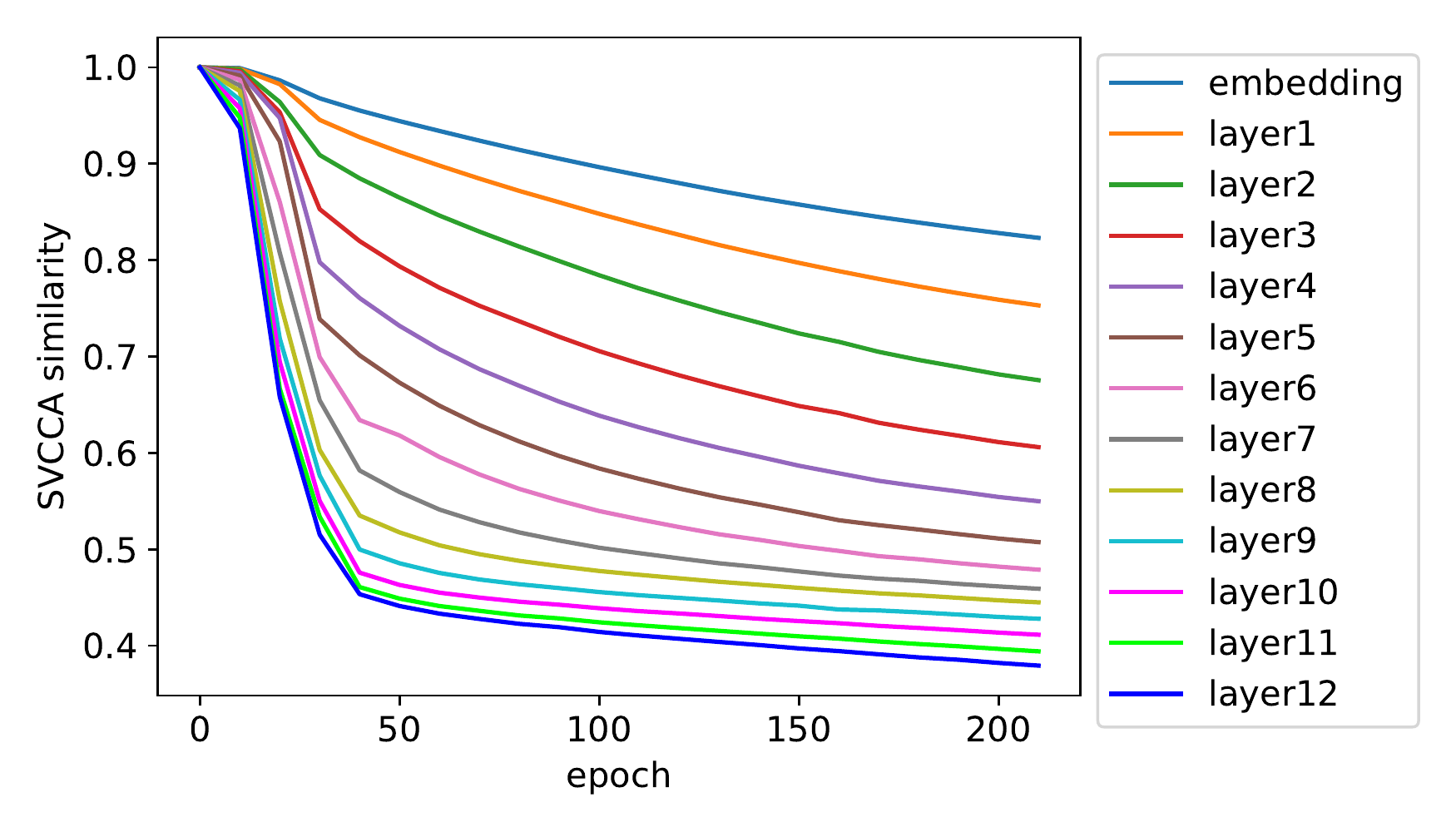}
    \caption{Training dynamics for all layers between $\mymodele{}$ and $\mymodelc{}_{Movies}$. Here both model and data size are 100\%.}
    \label{fig:training_dynamics_all_layers_movie}
\end{figure}

All models are trained on 4 NVIDIA A100 GPUs with a batch size of 32 per GPU. 
We use PyTorch \cite{NEURIPS2019_9015} and the HuggingFace library \cite{wolf2019huggingface} for all model implementation.

\section{Additional Details for Results}
\label{app:details_result}

\subsection{Additional Results for RQ1}
\label{app:rq1}

We provide additional experimental results for training dynamics on \texttt{Clothing Shoes and Jewelry} (Figure~\ref{fig:training_dynamics_all_layers_clothing}), \texttt{Electronics} (Figure~\ref{fig:training_dynamics_all_layers_electronics}), \texttt{Home and Kitchen} (Figure~\ref{fig:training_dynamics_all_layers_home}), and \texttt{Movies and TV} (Figure~\ref{fig:training_dynamics_all_layers_movie}).

\subsection{Additional Results for RQ2}
\label{app:rq2}

In \S~\ref{section:results}, we provided SVCCA results between $\mymodele{}$ and different $\mymodelc{}_i$s for three domains. Here we present the results for the rest of the two domains in Figure~\ref{fig:home-l0-all}, \ref{fig:home-l12-all}, \ref{fig:movie-l0-all}, and \ref{fig:movie-l12-all}.  

\subsection{Example of General and Domain-specific Words}
\label{app:example_specific_general_words}
We provide a sample of general words and domain specific words for each domain in Table~\ref{tab:example_words}. Note that list of general words are domain independent, i.e., the general word list is the same for all domains. 

\begin{table*}[ht]
\centering
\begin{tabular}{m{14cm}}
\Xhline{1pt}
\textbf{General words}: totally, preference, cost, mistake, hello, noticeable, play, factor, common, friend, previously, upon, explain, future, everyone
\\
\hline
\textbf{Books}: gutenberg, appendix, autobiographical, grammatically, bookshelves, democrat, asides, arabic, stagnant, curriculum, minutiae, gripped, publishers, referencing, socialism      \\
\textbf{Clothing}: marten, docker, florsheim, rockports, skechers, buckles, 38d, fleece, nylons, insoles, tees, pantyhose, puckered, slippers, footwear   \\
\textbf{Electronics}: printable, wifi, 105mm, aux, energizer, recordable, directories, reinstall, gigabit, reboots, portability, vga, hitachi, configurations, wirelessly  \\
\textbf{Home}:  cupcakes, kitchenaid, undercooked, ikea, chopper, mugs, steamers, juices, fiesta, kettles, aroma, toasted, rinsed ovens, airtight      \\
\textbf{Movie}: scenic, 16x9, nightclub, cheesiest, filmakers, supernova, serials, weepy, purists, incarnations, lionsgate, reportedly, suggestive, 1931, choreography  \\ 
\Xhline{1pt}
\end{tabular}
\caption{A representative sample of general words (top row) and domain specific words (bottom rows) taken from different categories (domains) of the dataset.}
\label{tab:example_words}
\end{table*}

\subsection{Additional Results for RQ3}
\label{app:rq3}
\begin{figure*}[ht]
     \centering
     \begin{subfigure}[b]{0.32\textwidth}
         \centering
         \includegraphics[width=\textwidth]{figures/svcca/general_vs_control_domain_Clothing_Shoes_and_Jewelry_layer0.pdf}
         \caption{$\ell_0$: all words}
         \label{fig:clothing-l0-all}
     \end{subfigure}
     \hfill
     \begin{subfigure}[b]{0.32\textwidth}
         \centering
         \includegraphics[width=\textwidth]{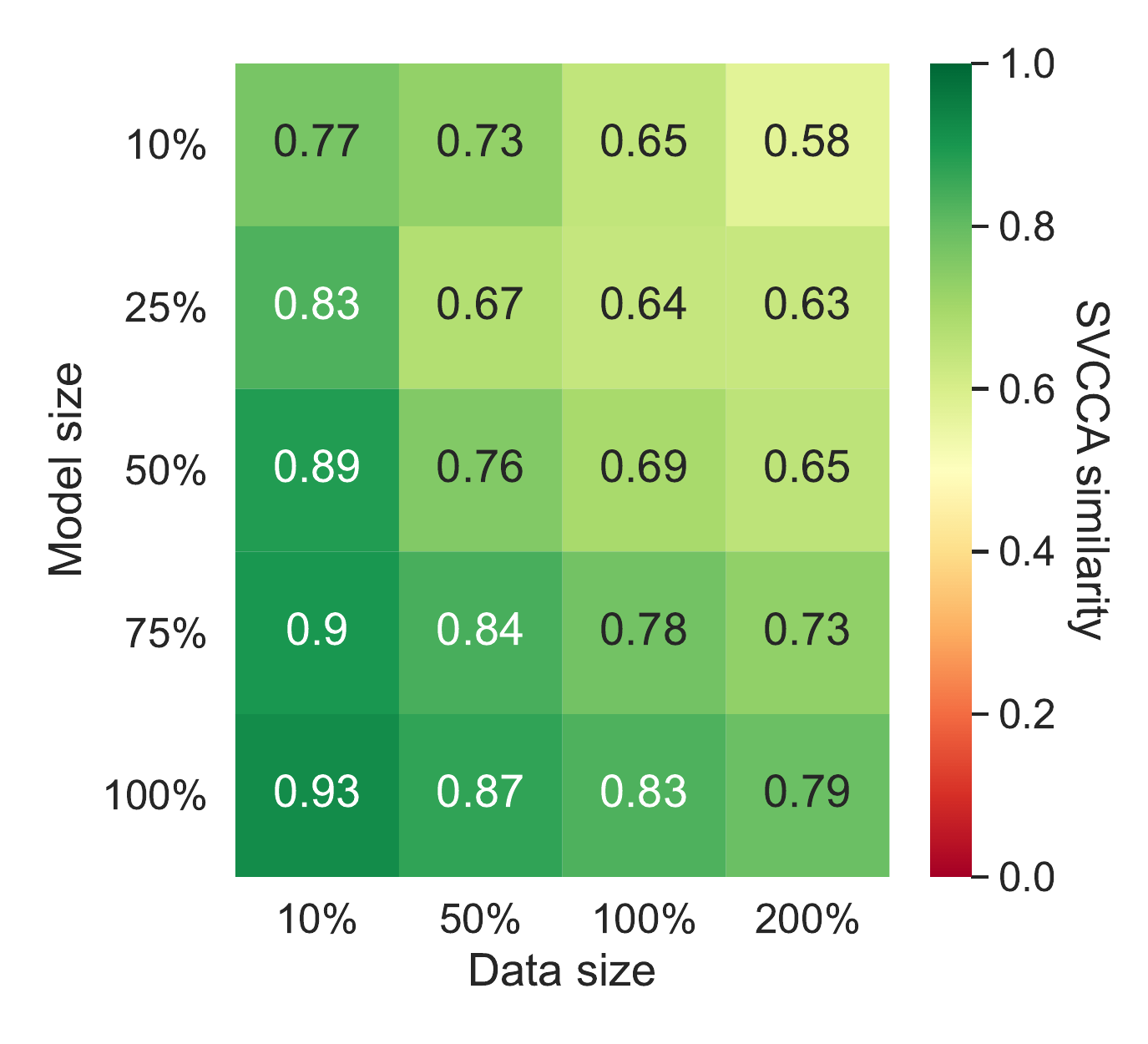}
         \caption{$\ell_0$: general words}
         \label{fig:clothing-l0-general}
     \end{subfigure}
     \hfill
     \begin{subfigure}[b]{0.32\textwidth}
         \centering
         \includegraphics[width=\textwidth]{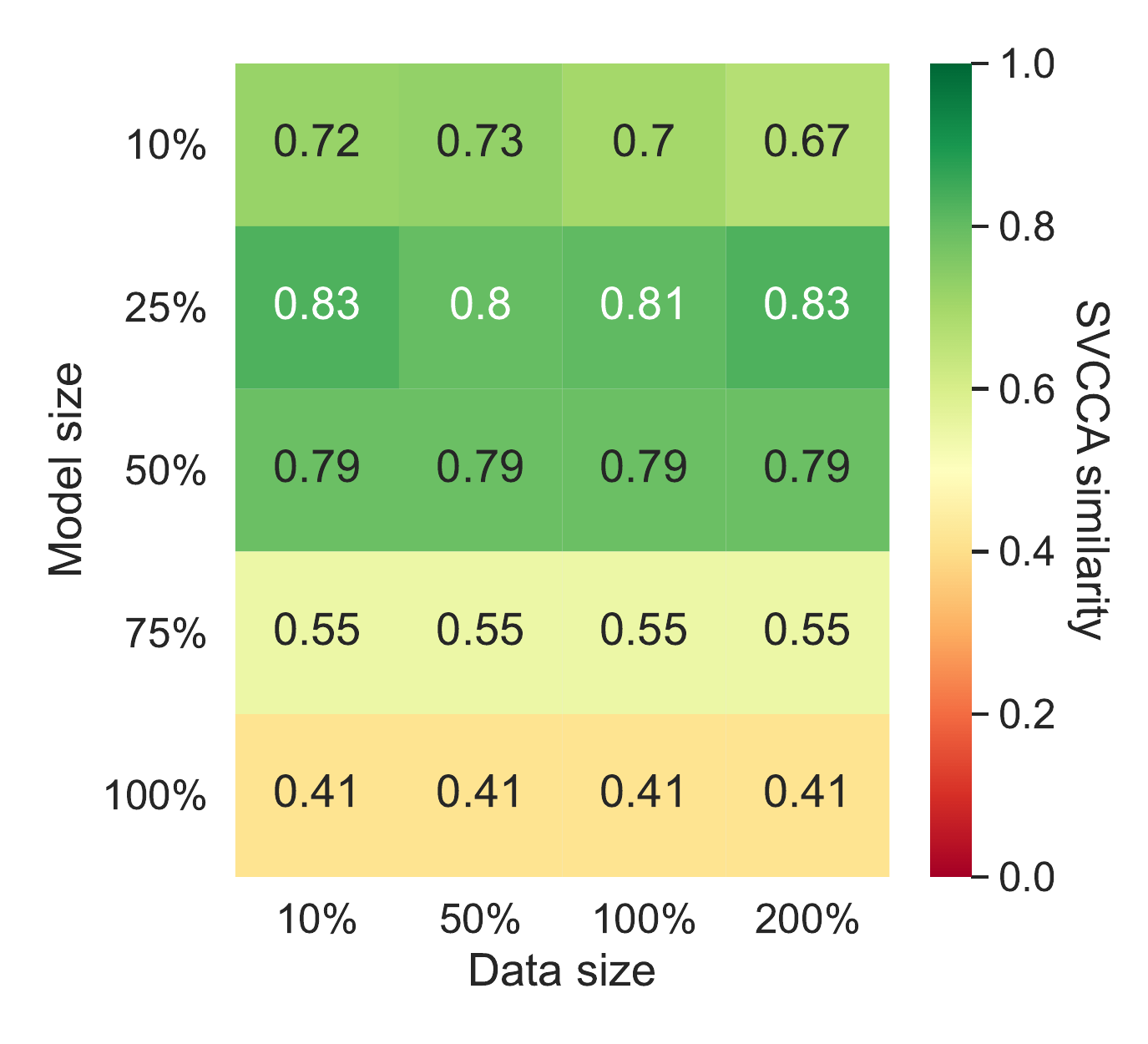}
         \caption{$\ell_0$: domain-specific words}
         \label{fig:clothing-l0-specific}
     \end{subfigure}
     \hfill
     \begin{subfigure}[b]{0.32\textwidth}
         \centering
         \includegraphics[width=\textwidth]{figures/svcca/general_vs_control_domain_Clothing_Shoes_and_Jewelry_layer12.pdf}
         \caption{$\ell_{12}$: all words}
         \label{fig:clothing-l12-all}
     \end{subfigure}
     \hfill
     \begin{subfigure}[b]{0.32\textwidth}
         \centering
         \includegraphics[width=\textwidth]{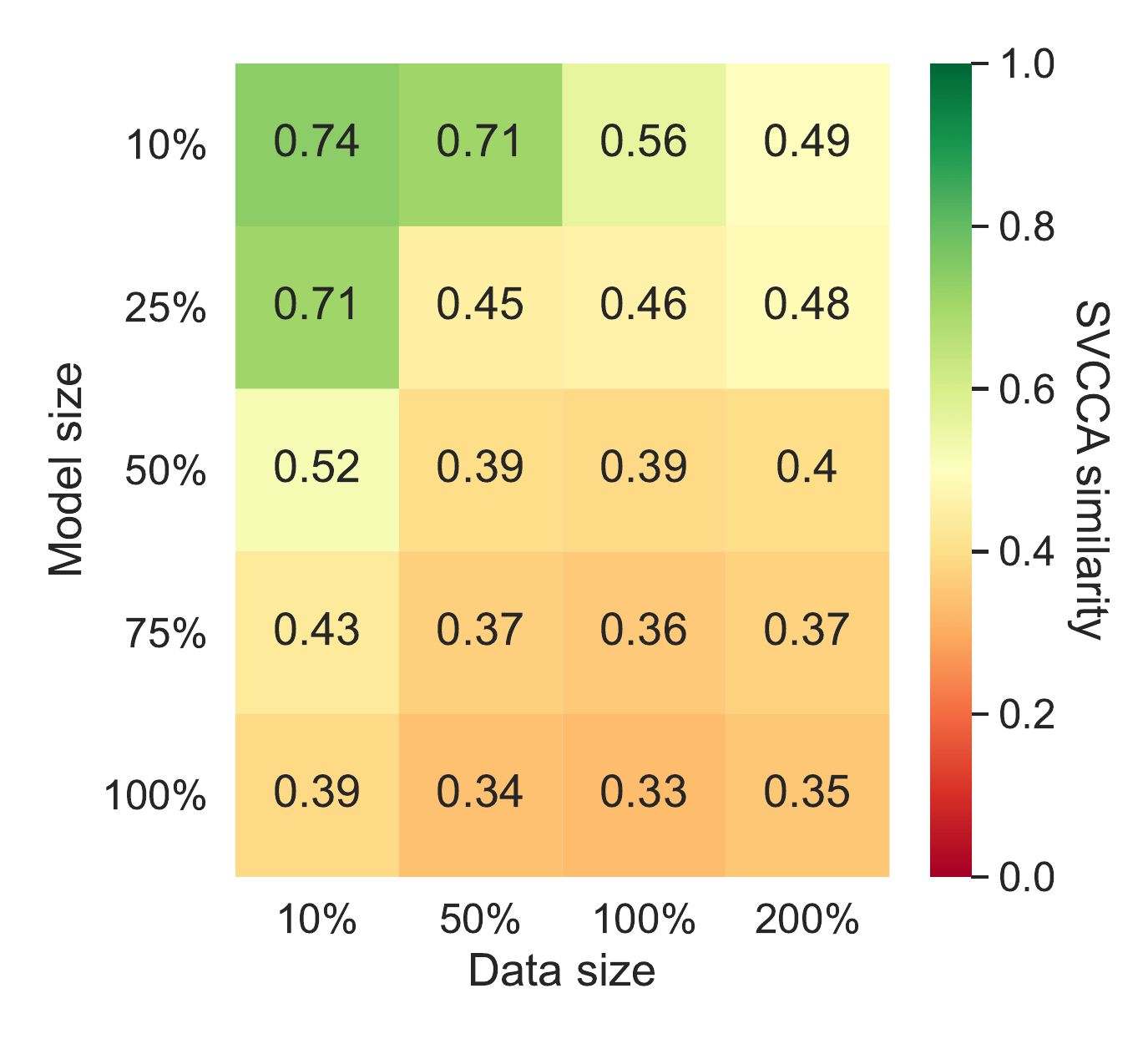}
         \caption{$\ell_{12}$: general words}
         \label{fig:clothing-l12-general}
     \end{subfigure}
     \hfill
     \begin{subfigure}[b]{0.32\textwidth}
         \centering
         \includegraphics[width=\textwidth]{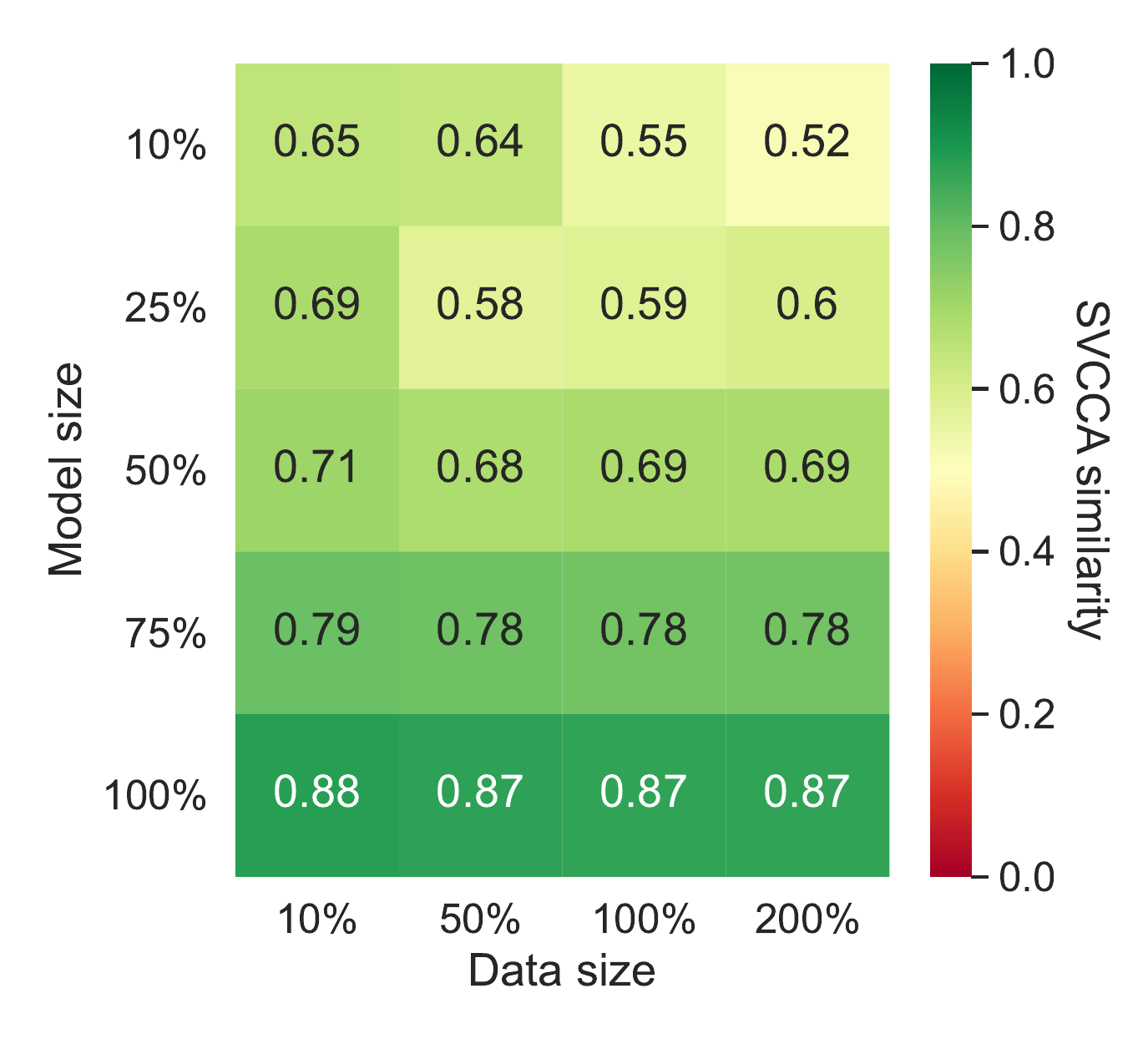}
         \caption{$\ell_{12}$: domain-specific words}
         \label{fig:clothing-l12-specific}
     \end{subfigure}
\caption{The SVCCA score between $\mymodele{}$ and $\mymodelc{}_{Clothing}$ for different subsets of tokens. The top row presents the results for the embedding layer $\ell_{0}$, and the bottom row presents them for the last layer $\ell_{12}$. }
\label{fig:clothing-all}
\end{figure*}

\begin{figure*}[htb]
     \centering
     \begin{subfigure}[b]{0.32\textwidth}
         \centering
         \includegraphics[width=\textwidth]{figures/svcca/general_vs_control_domain_Electronics_layer0.pdf}
         \caption{$\ell_0$: all words}
         \label{fig:electronics-l0-all}
     \end{subfigure}
     \hfill
     \begin{subfigure}[b]{0.32\textwidth}
         \centering
         \includegraphics[width=\textwidth]{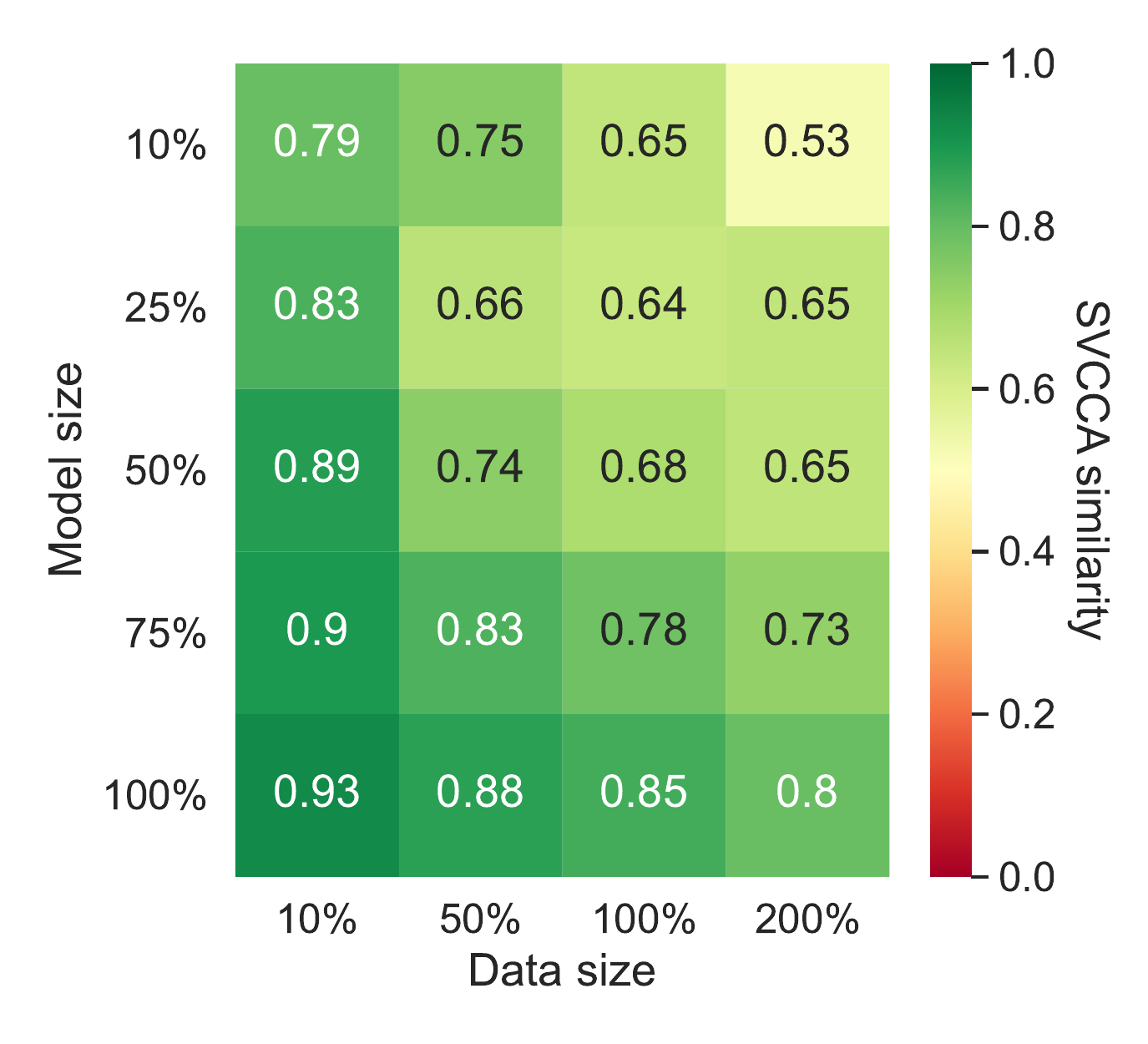}
         \caption{$\ell_0$: general words}
         \label{fig:electronics-l0-general}
     \end{subfigure}
     \hfill
     \begin{subfigure}[b]{0.32\textwidth}
         \centering
         \includegraphics[width=\textwidth]{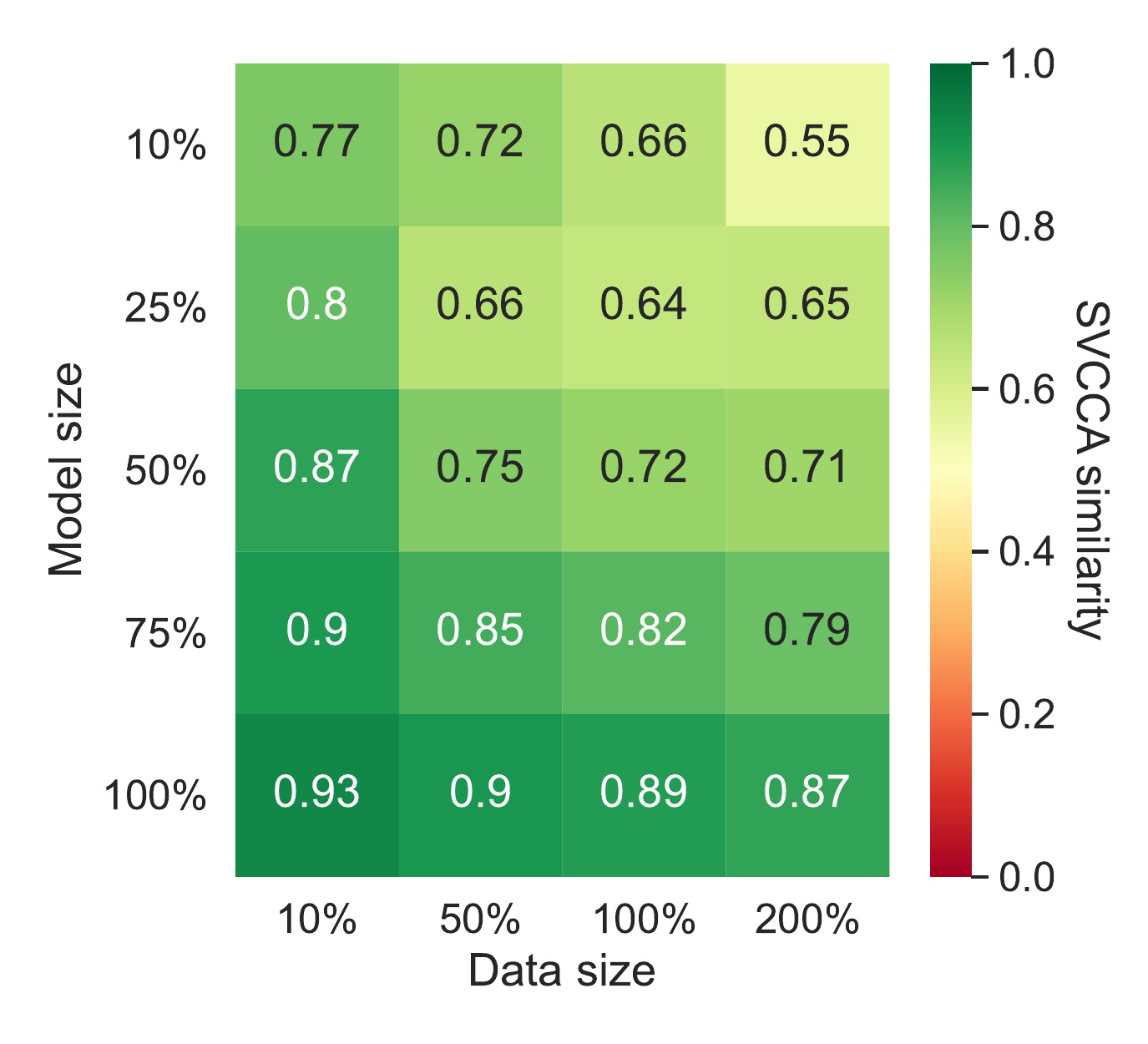}
         \caption{$\ell_0$: domain-specific words}
         \label{fig:electronics-l0-specific}
     \end{subfigure}
     \hfill
     \begin{subfigure}[b]{0.32\textwidth}
         \centering
         \includegraphics[width=\textwidth]{figures/svcca/general_vs_control_domain_Electronics_layer12.pdf}
         \caption{$\ell_{12}$: all words}
         \label{fig:electronics-l12-all}
     \end{subfigure}
     \hfill
     \begin{subfigure}[b]{0.32\textwidth}
         \centering
         \includegraphics[width=\textwidth]{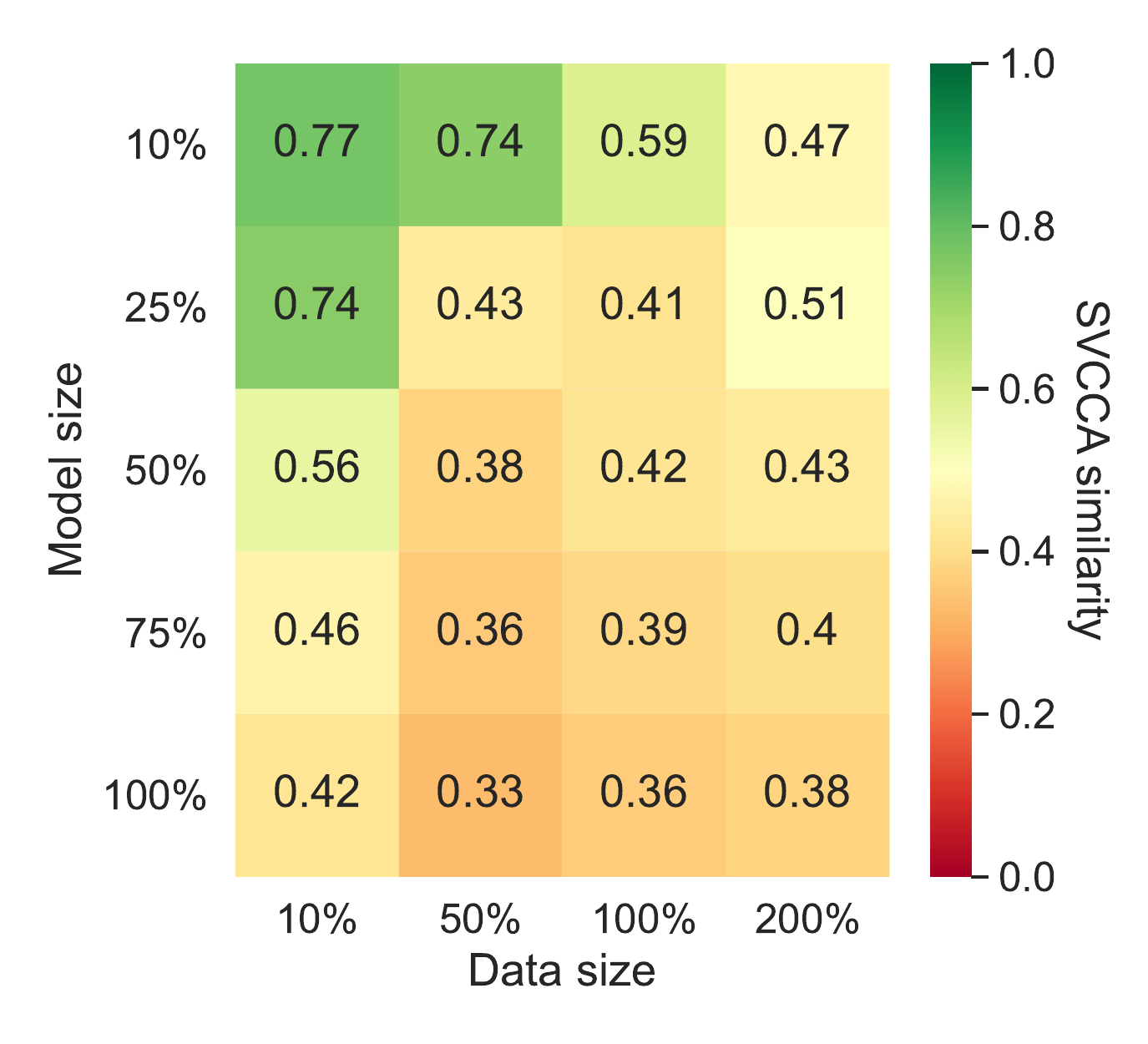}
         \caption{$\ell_{12}$: general words}
         \label{fig:electronics-l12-general}
     \end{subfigure}
     \hfill
     \begin{subfigure}[b]{0.32\textwidth}
         \centering
         \includegraphics[width=\textwidth]{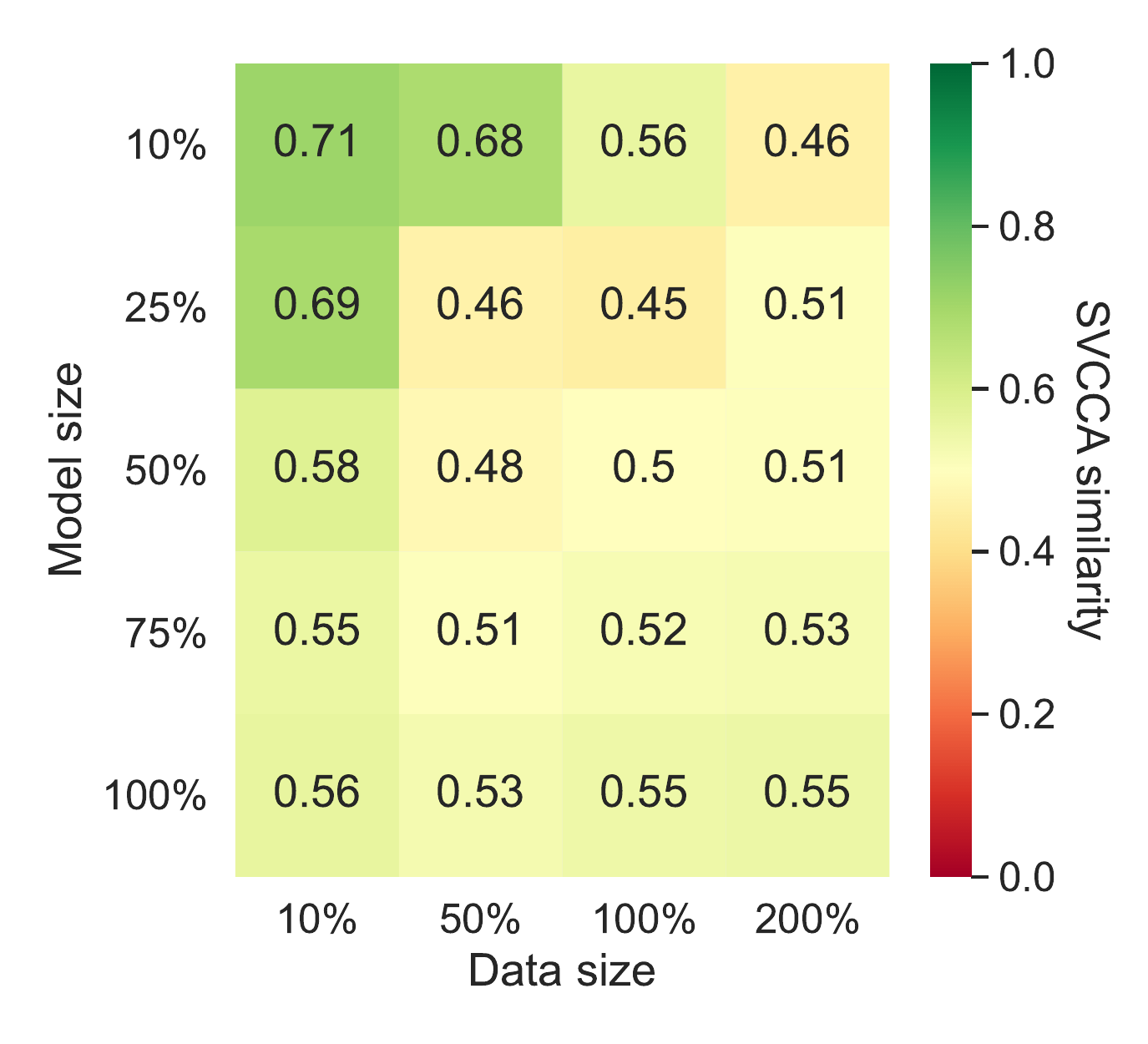}
         \caption{$\ell_{12}$: domain-specific words}
         \label{fig:electronics-l12-specific}
     \end{subfigure}
\caption{The SVCCA score between $\mymodele{}$ and $\mymodelc{}_{Electronics}$ for different subsets of tokens. The top row presents the results for the embedding layer $\ell_{0}$, and the bottom row presents them for the last layer $\ell_{12}$. }
\label{fig:electronics-all}
\end{figure*}

\begin{figure*}[htb]
     \centering
     \begin{subfigure}[b]{0.32\textwidth}
         \centering
         \includegraphics[width=\textwidth]{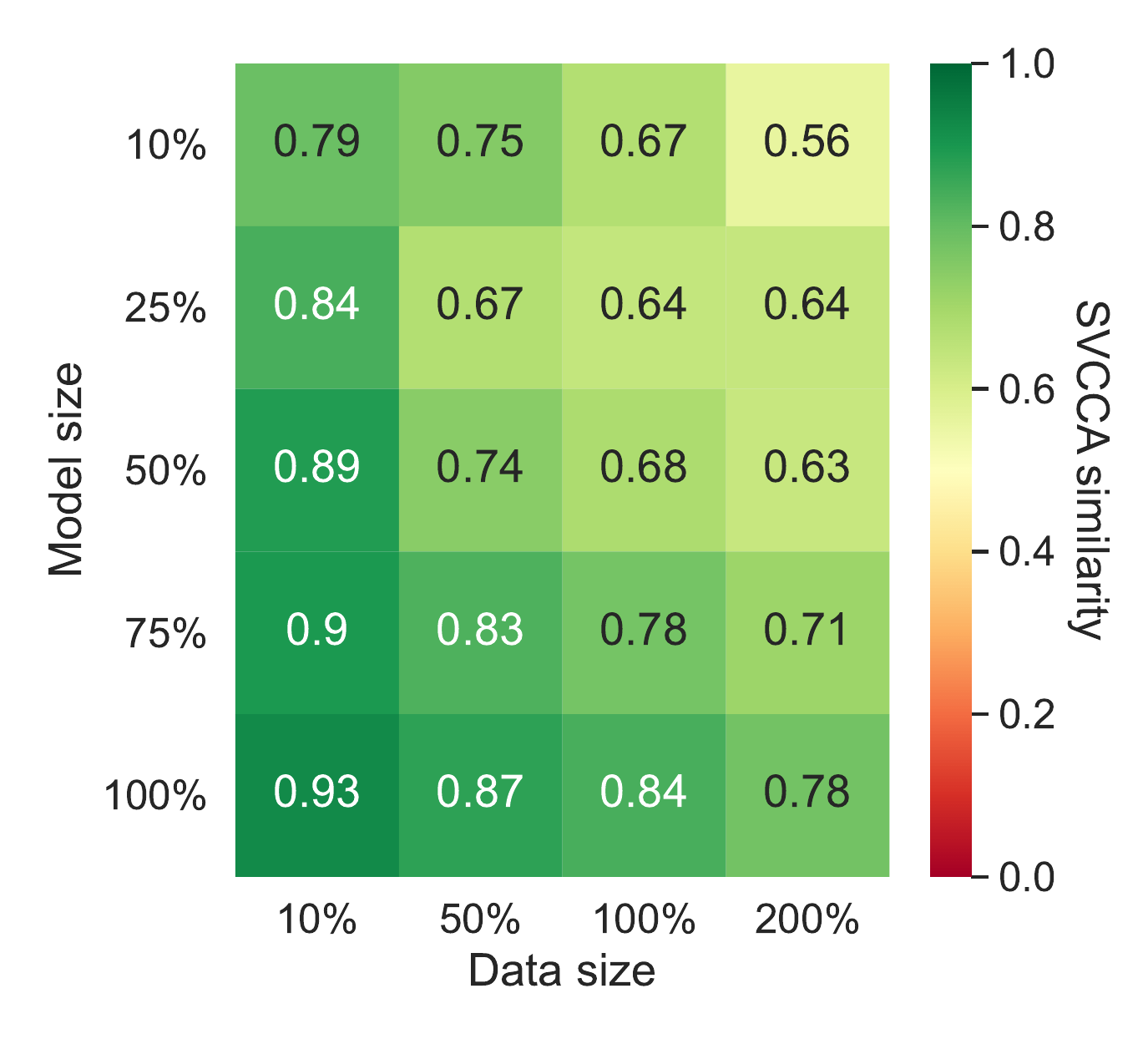}
         \caption{$\ell_0$: all words}
         \label{fig:home-l0-all}
     \end{subfigure}
     \hfill
     \begin{subfigure}[b]{0.32\textwidth}
         \centering
         \includegraphics[width=\textwidth]{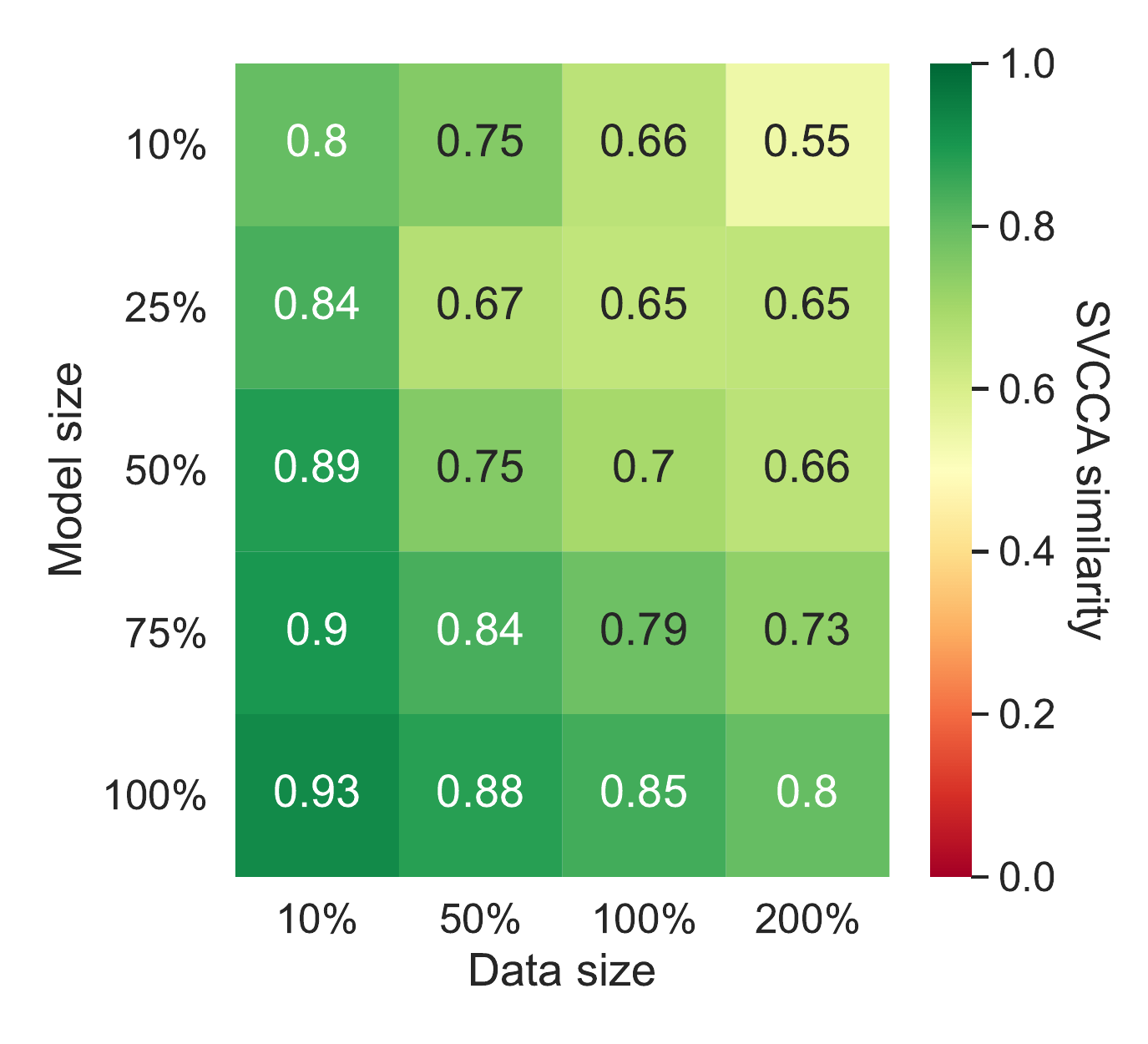}
         \caption{$\ell_0$: general words}
         \label{fig:home-l0-general}
     \end{subfigure}
     \hfill
     \begin{subfigure}[b]{0.32\textwidth}
         \centering
         \includegraphics[width=\textwidth]{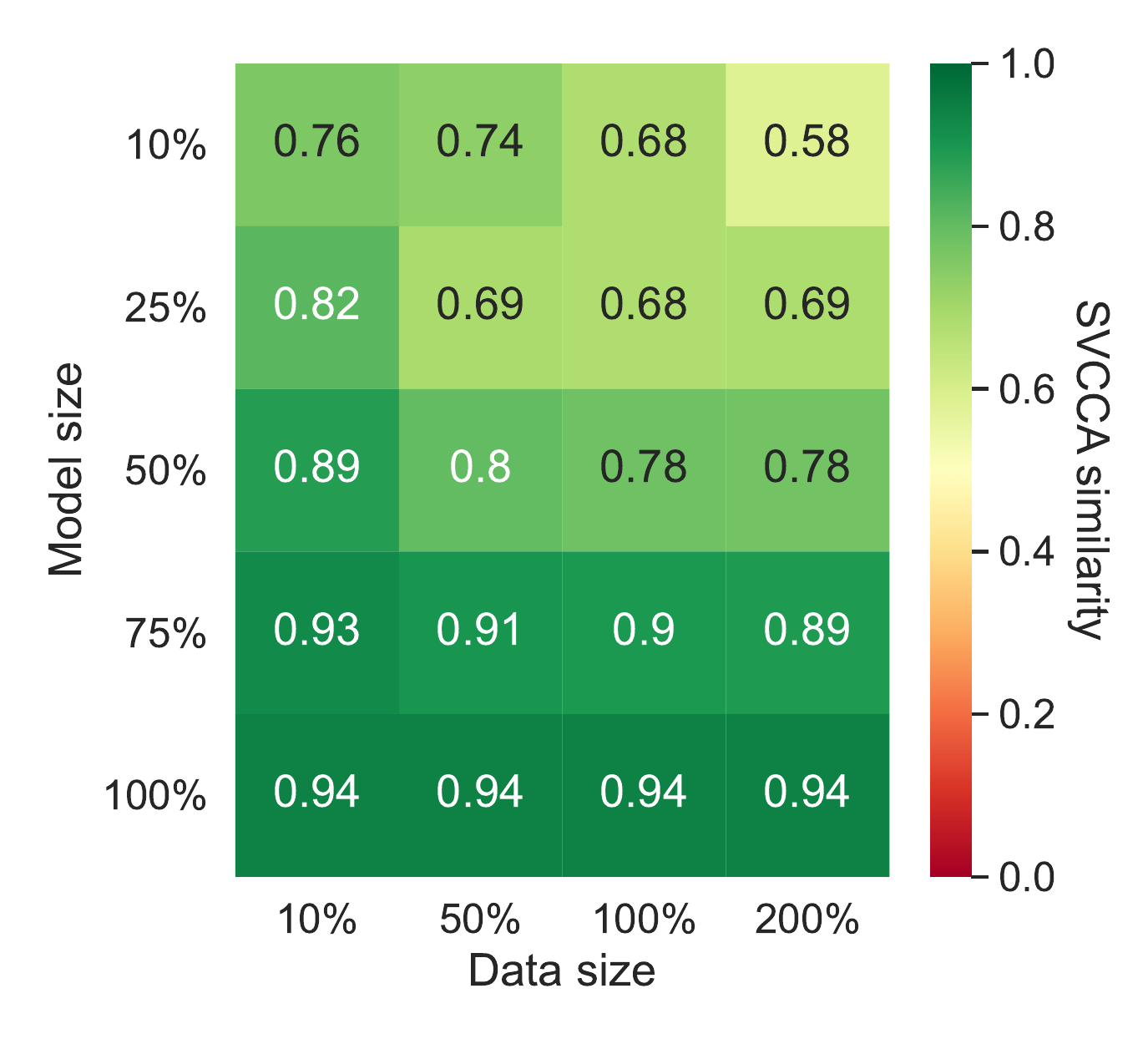}
         \caption{$\ell_0$: domain-specific words}
         \label{fig:home-l0-specific}
     \end{subfigure}
     \hfill
     \begin{subfigure}[b]{0.32\textwidth}
         \centering
         \includegraphics[width=\textwidth]{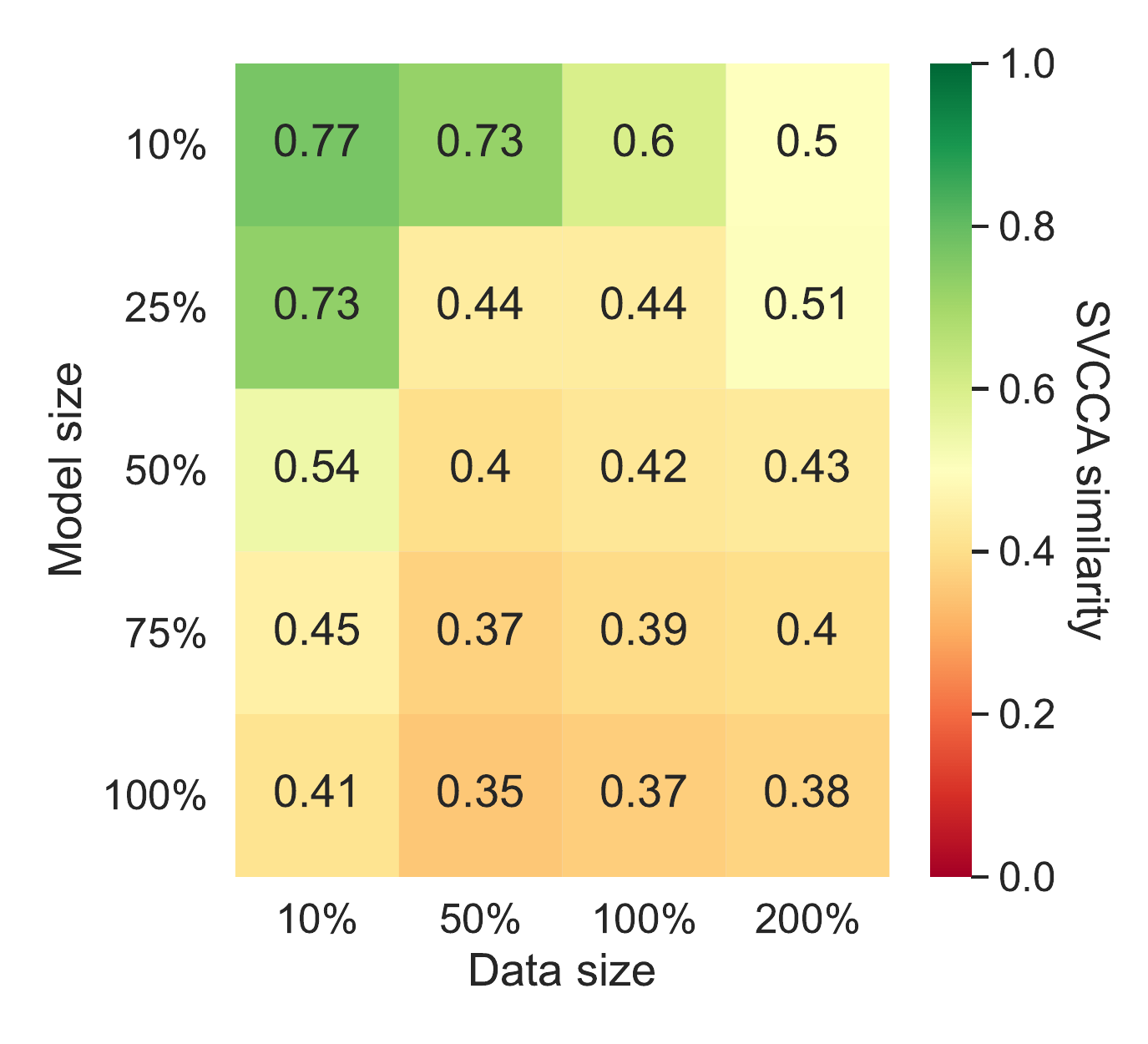}
         \caption{$\ell_{12}$: all words}
         \label{fig:home-l12-all}
     \end{subfigure}
     \hfill
     \begin{subfigure}[b]{0.32\textwidth}
         \centering
         \includegraphics[width=\textwidth]{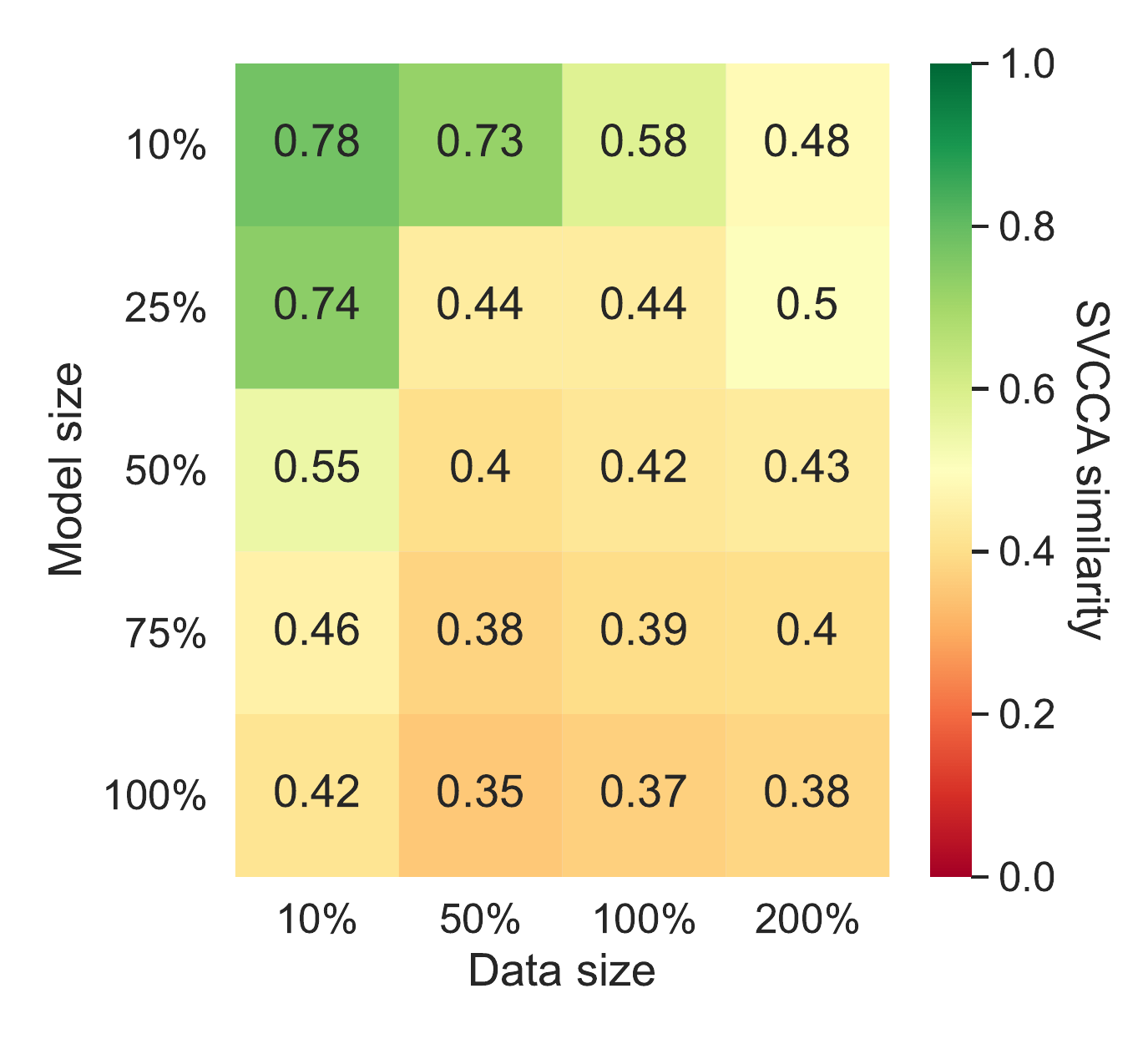}
         \caption{$\ell_{12}$: general words}
         \label{fig:home-l12-general}
     \end{subfigure}
     \hfill
     \begin{subfigure}[b]{0.32\textwidth}
         \centering
         \includegraphics[width=\textwidth]{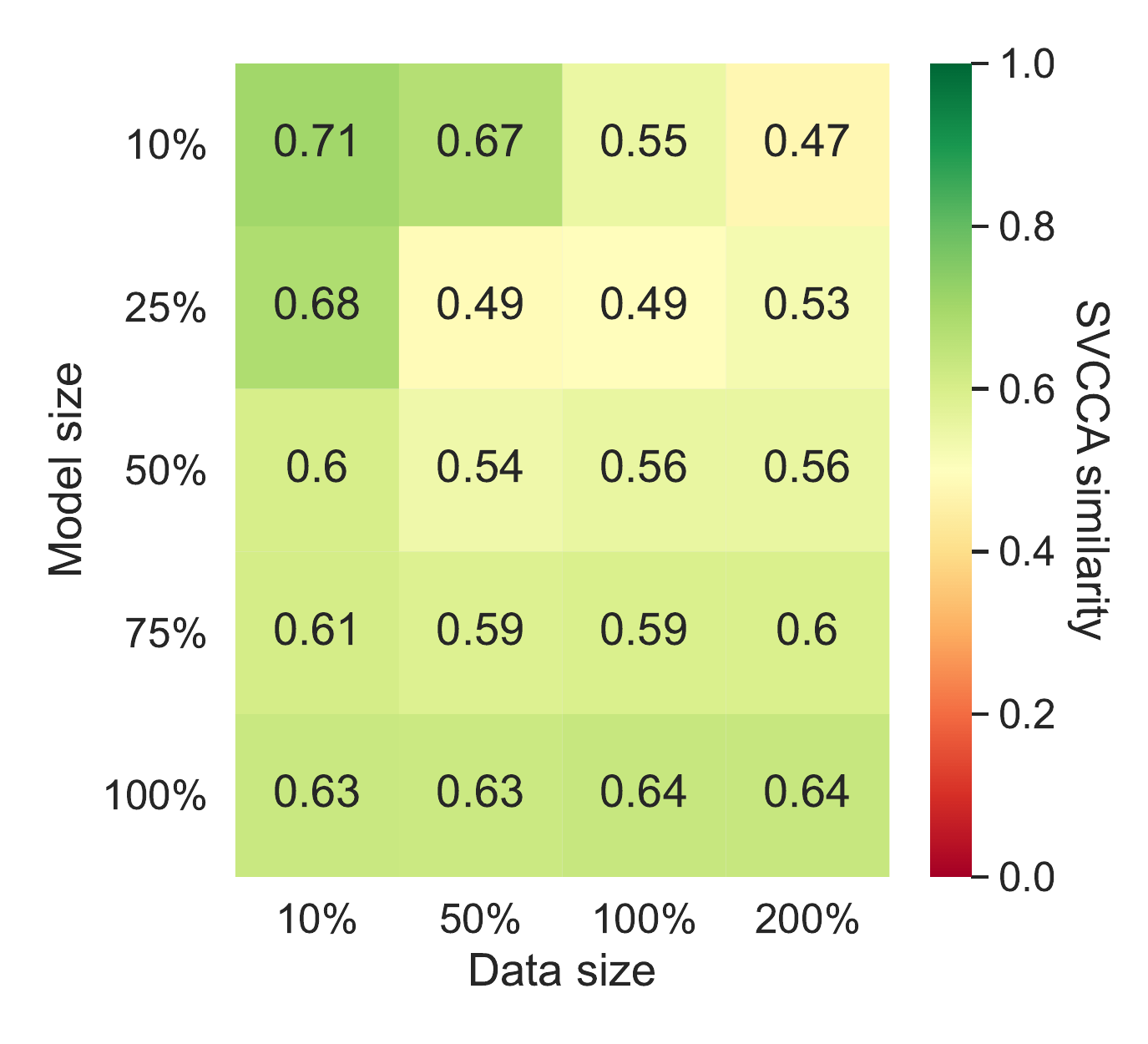}
         \caption{$\ell_{12}$: domain-specific words}
         \label{fig:home-l12-specific}
     \end{subfigure}
\caption{The SVCCA score between $\mymodele{}$ and $\mymodelc{}_{Home}$ for different subsets of tokens. The top row presents the results for the embedding layer $\ell_{0}$, and the bottom row presents them for the last layer $\ell_{12}$. }
\label{fig:home-all}
\end{figure*}

\begin{figure*}[htb]
     \centering
     \begin{subfigure}[b]{0.32\textwidth}
         \centering
         \includegraphics[width=\textwidth]{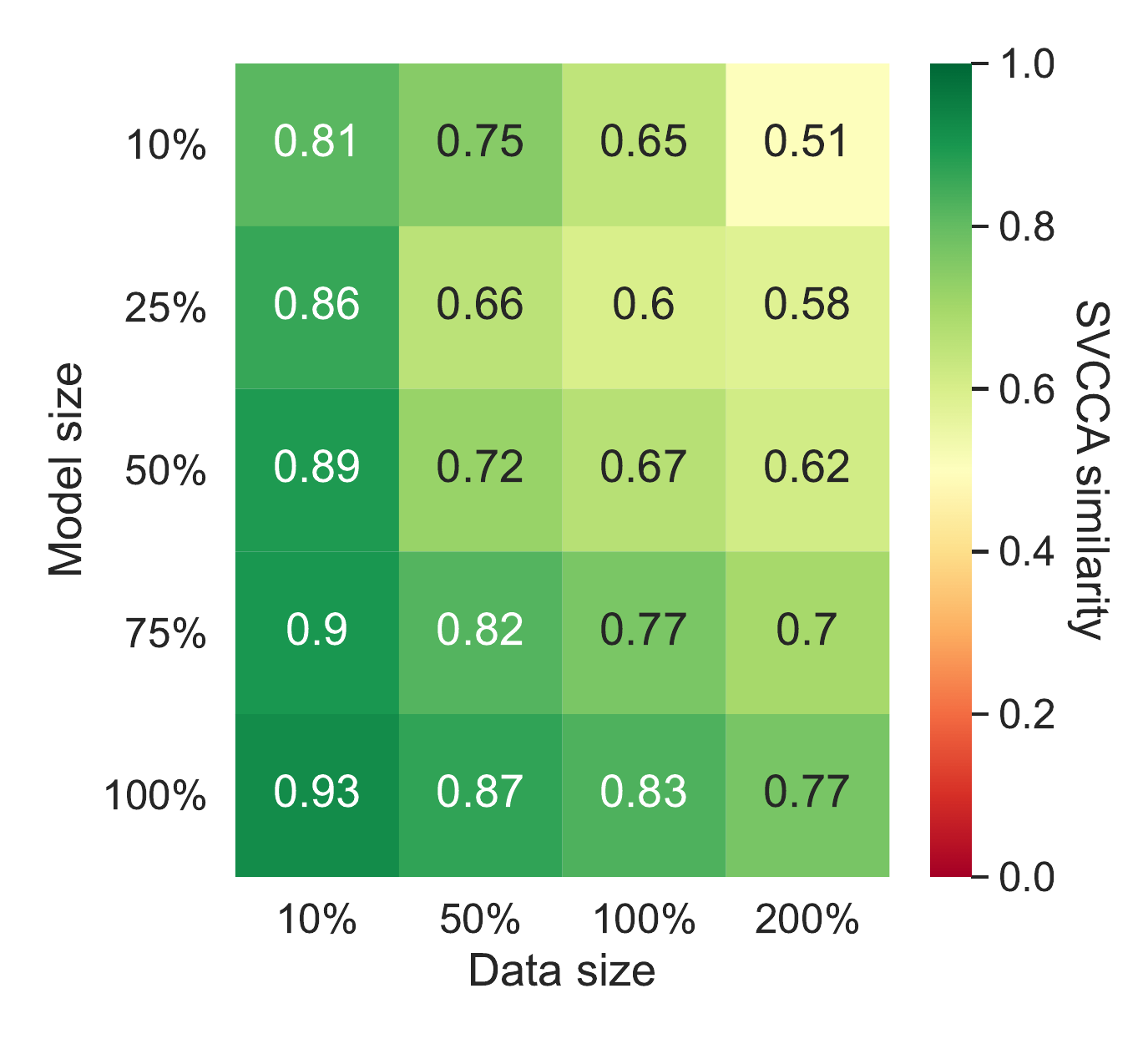}
         \caption{$\ell_0$: all words}
         \label{fig:movie-l0-all}
     \end{subfigure}
     \hfill
     \begin{subfigure}[b]{0.32\textwidth}
         \centering
         \includegraphics[width=\textwidth]{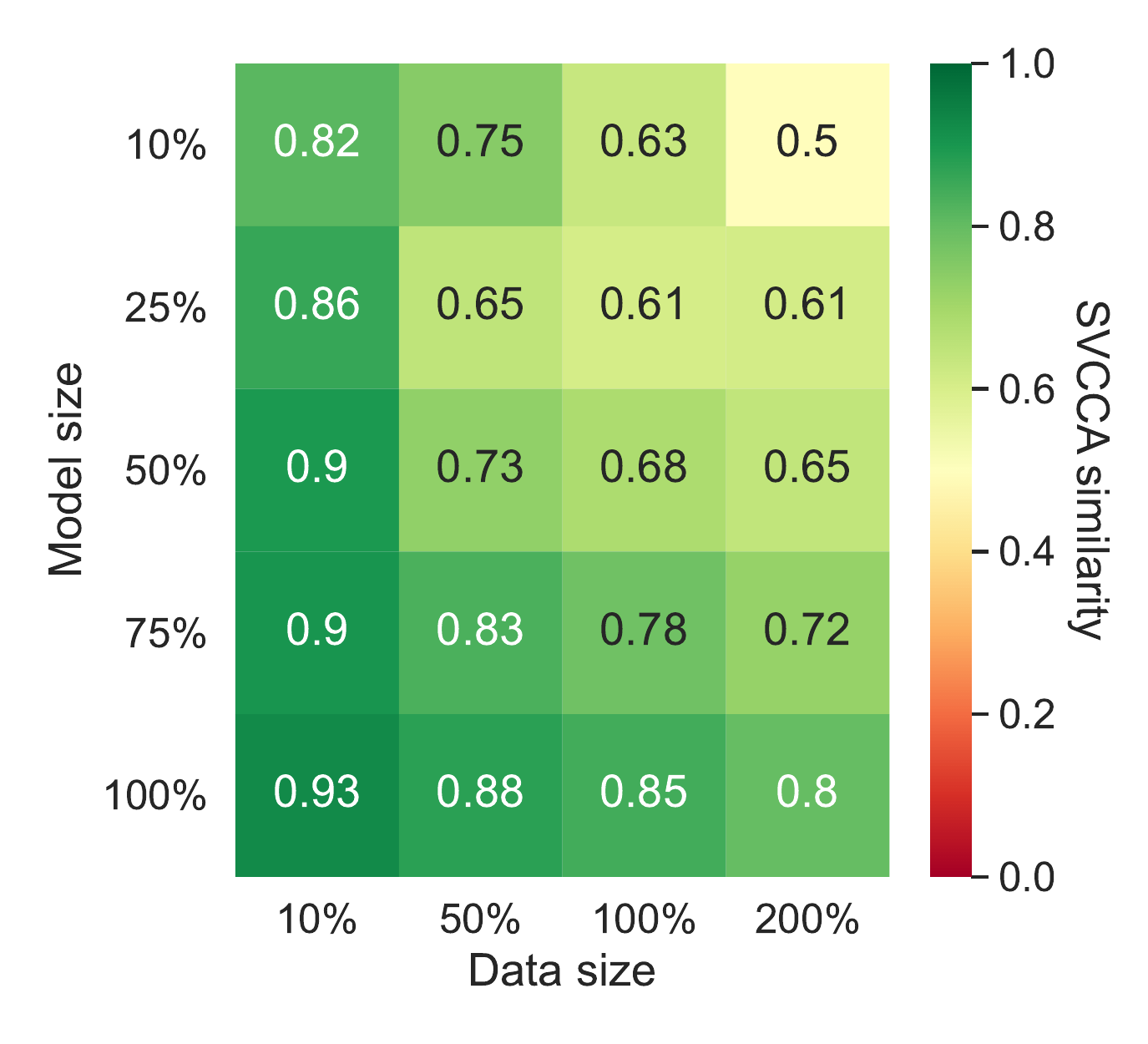}
         \caption{$\ell_0$: general words}
         \label{fig:movie-l0-general}
     \end{subfigure}
     \hfill
     \begin{subfigure}[b]{0.32\textwidth}
         \centering
         \includegraphics[width=\textwidth]{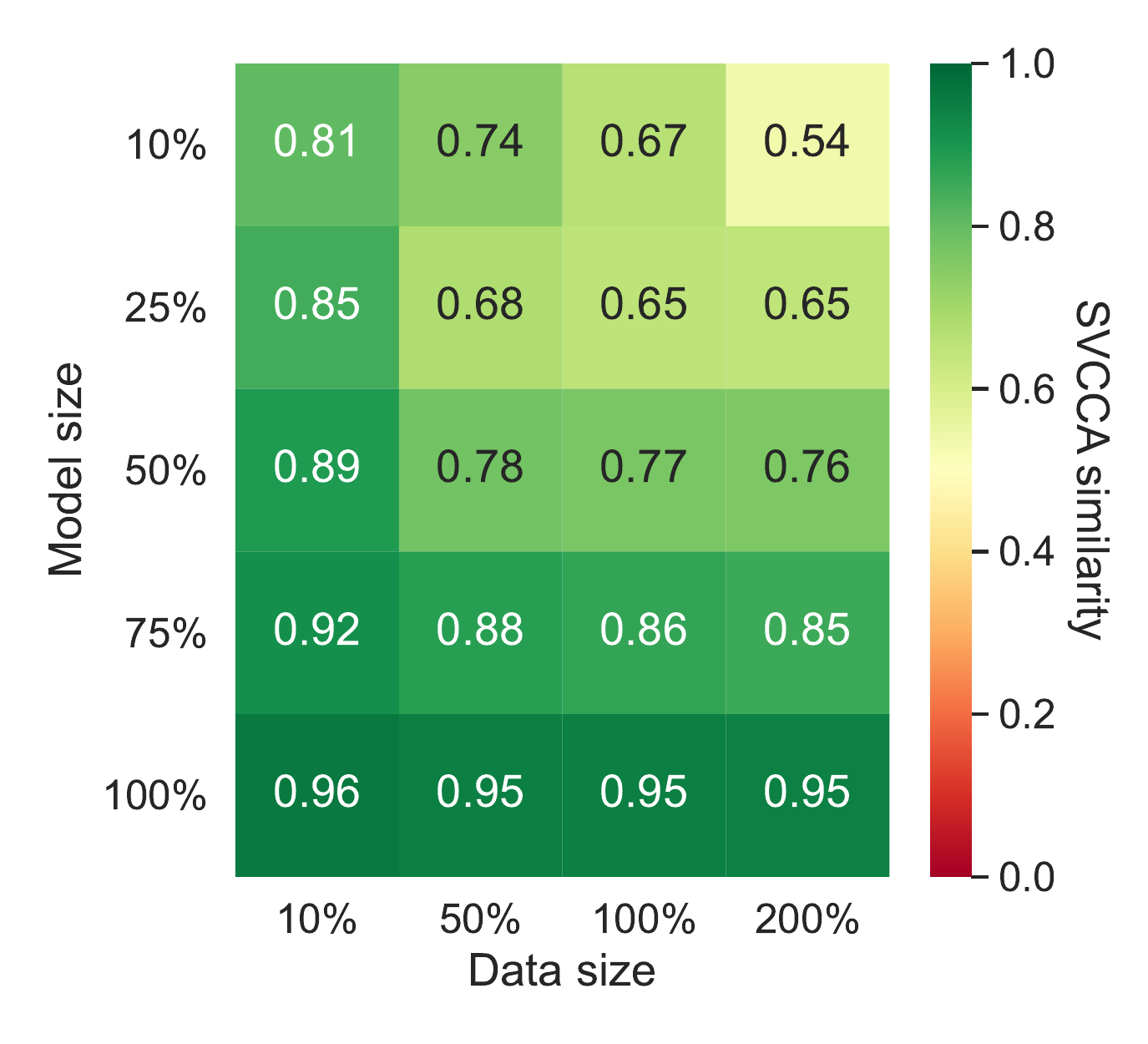}
         \caption{$\ell_0$: domain-specific words}
         \label{fig:movie-l0-specific}
     \end{subfigure}
     \hfill
     \begin{subfigure}[b]{0.32\textwidth}
         \centering
         \includegraphics[width=\textwidth]{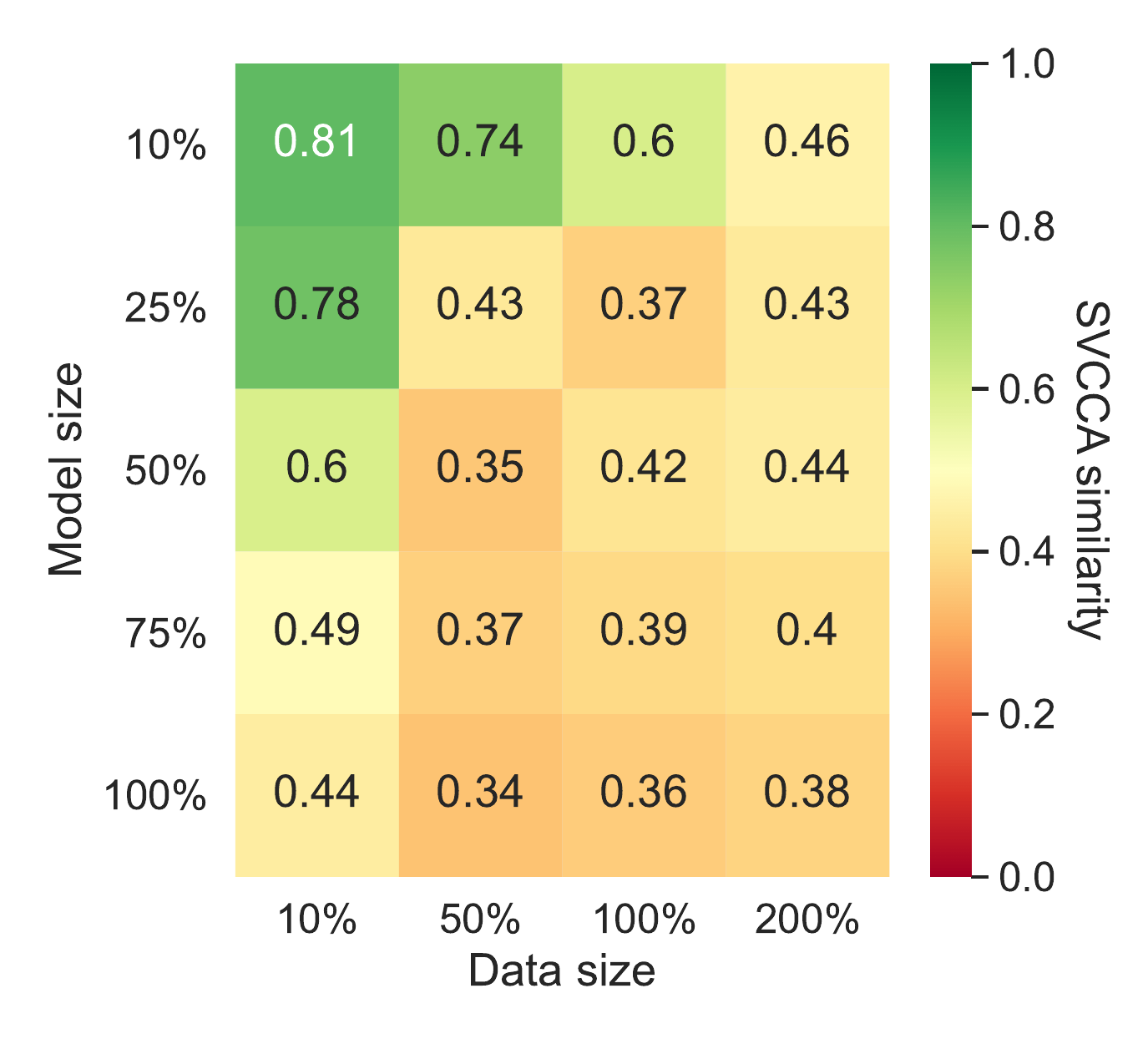}
         \caption{$\ell_{12}$: all words}
         \label{fig:movie-l12-all}
     \end{subfigure}
     \hfill
     \begin{subfigure}[b]{0.32\textwidth}
         \centering
         \includegraphics[width=\textwidth]{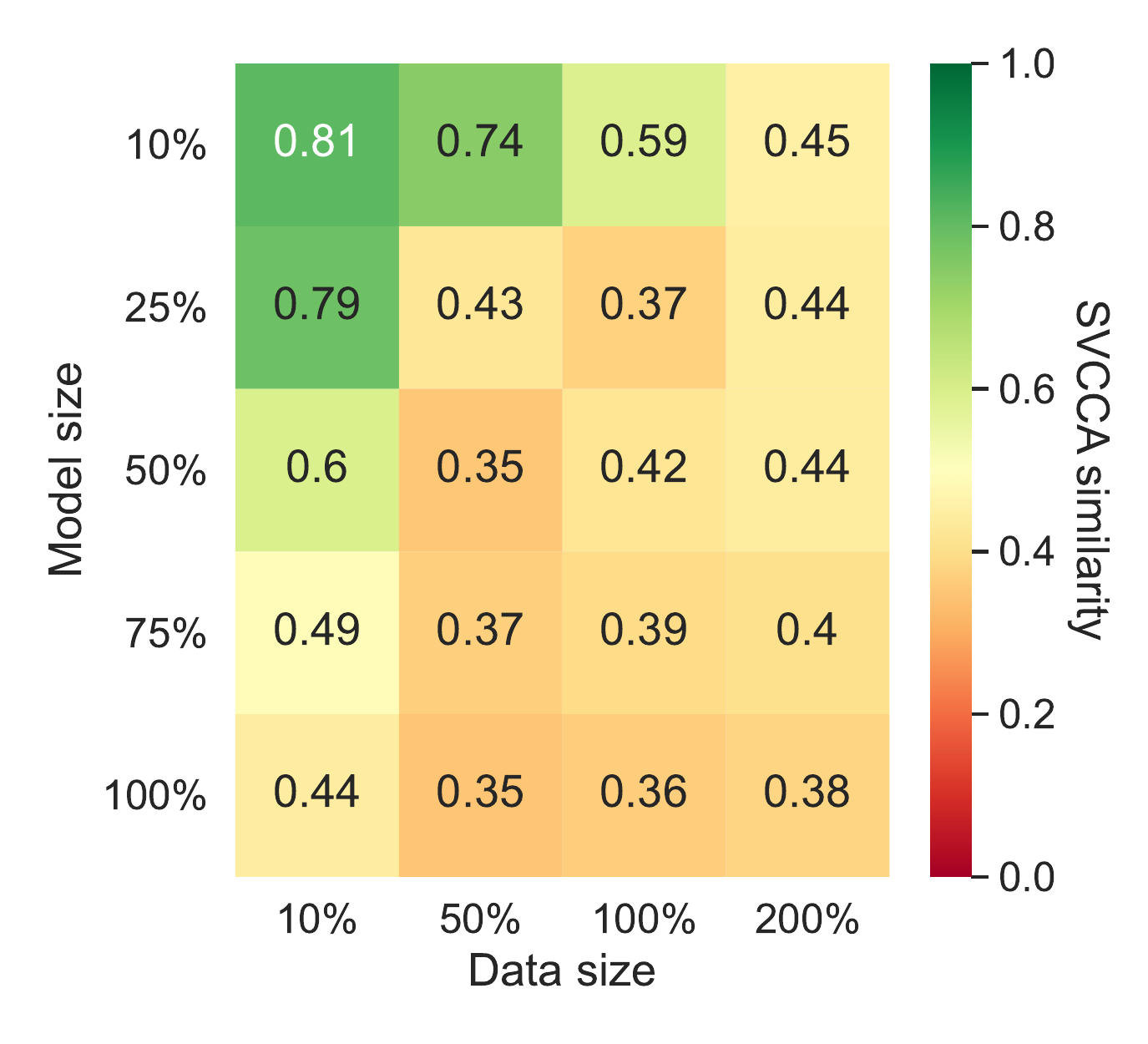}
         \caption{$\ell_{12}$: general words}
         \label{fig:movie-l12-general}
     \end{subfigure}
     \hfill
     \begin{subfigure}[b]{0.32\textwidth}
         \centering
         \includegraphics[width=\textwidth]{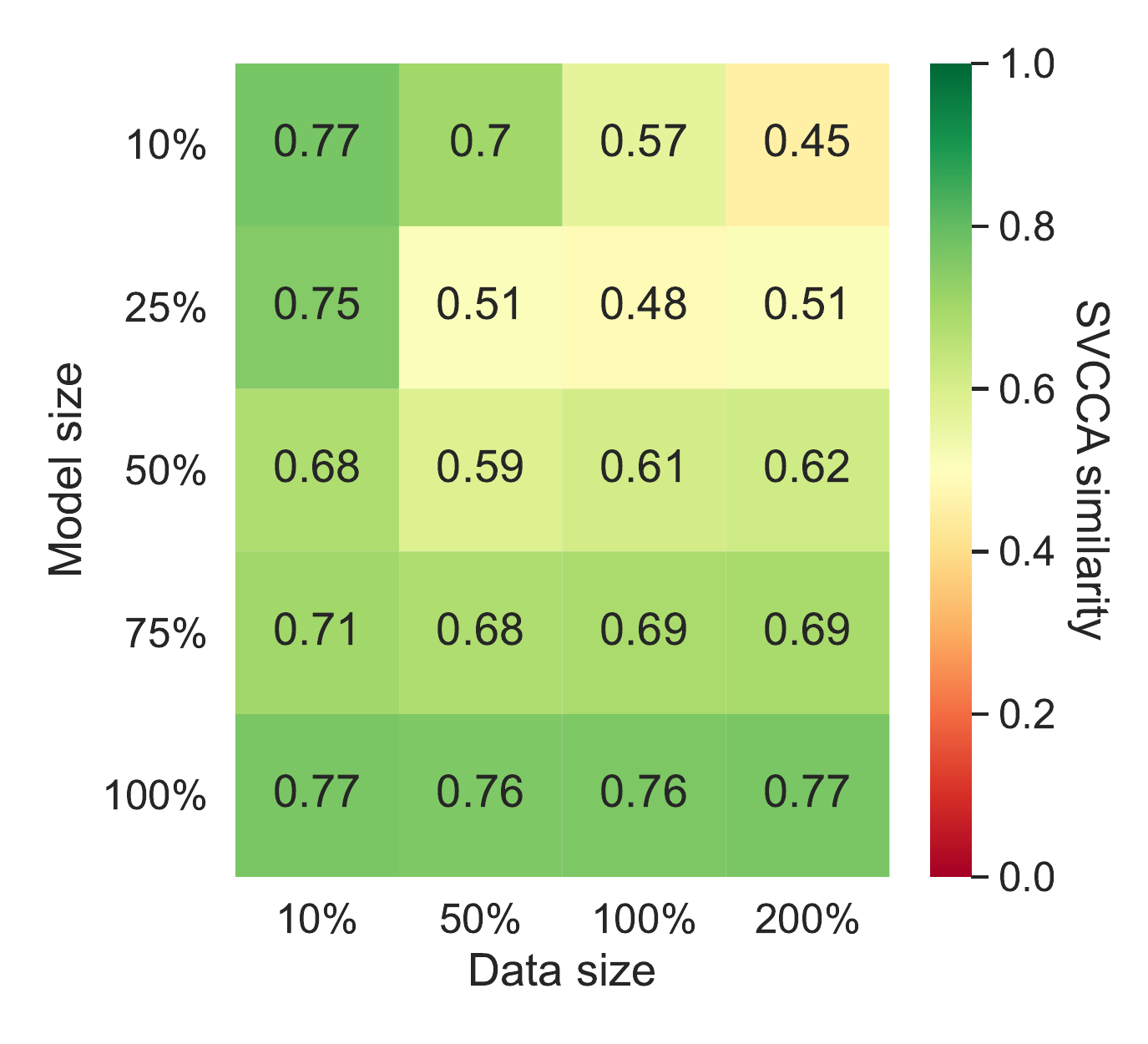}
         \caption{$\ell_{12}$: domain-specific words}
         \label{fig:movie-l12-specific}
     \end{subfigure}
\caption{The SVCCA score between $\mymodele{}$ and $\mymodelc{}_{Movies}$ for different subsets of tokens. The top row presents the results for the embedding layer $\ell_{0}$, and the bottom row presents them for the last layer $\ell_{12}$. }
\label{fig:movie-all}
\end{figure*}

Here we present additional results for SVCCA score between $\mymodele{}$ and $\mymodelc{}_{i}$ for different subsets of tokens. Figure~\ref{fig:clothing-all} illustrates for $\mymodelc{}_{Clothing}$, Figure~\ref{fig:electronics-all} illustrates for $\mymodelc{}_{Electronics}$, Figure~\ref{fig:home-all} illustrates for $\mymodelc{}_{Home}$, and Figure~\ref{fig:movie-all} illustrates for $\mymodelc{}_{Movies}$.

\subsection{Additional Results for RQ4}
\label{app:rq4}
Here we provide more example MLM predictions of $\mymodele{}$ and $\mymodelc{}_i$. Table~\ref{tab:predictions_using_embedding_more} presents predictions using k-nearest neighbors of the word embeddings. Table~\ref{tab:predictions_using_last_layer_more} presents predictions using the final layer representation. 

\begin{table*}[ht]
\small
    \begin{subtable}[h]{0.45\textwidth}
        \centering
        \begin{tabular}{|cc|cc|}
            \hline
            \multicolumn{2}{|c|}{m=50\%} & \multicolumn{2}{c|}{m=100\%} \\ \hline
            \multicolumn{1}{|c|}{$\mymodele{}$} & $\mymodelc{}_i$ & \multicolumn{1}{c|}{$\mymodele{}$} & $\mymodelc{}_i$     \\ \hline
            \multicolumn{1}{|c|}{editors}        & volumns       & \multicolumn{1}{c|}{editors} & editors        \\ 
            \multicolumn{1}{|c|}{publisher}        & buyer       & \multicolumn{1}{c|}{publisher}        & publisher        \\ 
            \multicolumn{1}{|c|}{heirs}        & listing      & \multicolumn{1}{c|}{editor} & editor        \\ 
            \multicolumn{1}{|c|}{libraries}        & edit       & \multicolumn{1}{c|}{writers}    & authors        \\ 
            \multicolumn{1}{|c|}{universities}        & hardcover       & \multicolumn{1}{c|}{authors}        & reviewers        \\ \hline
            \end{tabular}
       \caption{5-nearest neighbors for the domain-specific word \textbf{publishers} with $i$=\texttt{Books}. }

    \end{subtable}
    \hfill
    \begin{subtable}[h]{0.45\textwidth}
        \centering
        \begin{tabular}{|cc|cc|}
            \hline
            \multicolumn{2}{|c|}{m=50\%}    & \multicolumn{2}{c|}{m=100\%}           \\ \hline
            \multicolumn{1}{|c|}{$\mymodele{}$} & $\mymodelc{}_i$ & \multicolumn{1}{c|}{$\mymodele{}$} & $\mymodelc{}_i$ \\ \hline
            \multicolumn{1}{|c|}{towards}   & towards & \multicolumn{1}{c|}{towards}   & towards             \\ 
            \multicolumn{1}{|c|}{beside}      &   settled       & \multicolumn{1}{c|}{against}    & at             \\ 
            \multicolumn{1}{|c|}{surrounding}       & at         & \multicolumn{1}{c|}{onto}  & onto             \\ 
            \multicolumn{1}{|c|}{beneath}       & concerning     & \multicolumn{1}{c|}{at}    & against             \\ 
            \multicolumn{1}{|c|}{against}    & behind     & \multicolumn{1}{c|}{beside}      & near             \\ \hline
        \end{tabular}
       \caption{5-nearest neighbors for the general word \textbf{toward} with $i$=\texttt{Books}.}

    \end{subtable}
    \hfill
    \begin{subtable}[h]{0.45\textwidth}
        \centering
        \begin{tabular}{|cc|cc|}
            \hline
            \multicolumn{2}{|c|}{m=50\%}            & \multicolumn{2}{c|}{m=100\%}           \\ \hline
            \multicolumn{1}{|c|}{$\mymodele{}$} & $\mymodelc{}_i$ & \multicolumn{1}{c|}{$\mymodele{}$} & $\mymodelc{}_i$ \\ \hline
            \multicolumn{1}{|c|}{comics}   & jokes    & \multicolumn{1}{c|}{comics}  & comics             \\ 
            \multicolumn{1}{|c|}{jokes}   & joke    & \multicolumn{1}{c|}{comedian}  & joke             \\ 
            \multicolumn{1}{|c|}{comedian}   & accolades    & \multicolumn{1}{c|}{laughs}  & comedian             \\ 
            \multicolumn{1}{|c|}{directors}   & critics    & \multicolumn{1}{c|}{comedies}  & critics             \\ 
            \multicolumn{1}{|c|}{commentators}   & reviewers    & \multicolumn{1}{c|}{jokes}  & laughs             \\ \hline
        \end{tabular}
       \caption{5-nearest neighbors for the domain-specific word \textbf{comedians} with $i$=\texttt{Movies and TV}.}

    \end{subtable}
    \hfill
    \begin{subtable}[h]{0.45\textwidth}
        \centering
        \begin{tabular}{|cc|cc|}
            \hline
            \multicolumn{2}{|c|}{m=50\%}            & \multicolumn{2}{c|}{m=100\%}           \\ \hline
            \multicolumn{1}{|c|}{$\mymodele{}$} & $\mymodelc{}_i$ & \multicolumn{1}{c|}{$\mymodele{}$} & $\mymodelc{}_i$ \\ \hline
            \multicolumn{1}{|c|}{print}  & vinyl & \multicolumn{1}{c|}{plastic}   & plastic             \\ 
            \multicolumn{1}{|c|}{plastic} & bonded  & \multicolumn{1}{c|}{print}   & vinyl             \\ 
            \multicolumn{1}{|c|}{cloth} & plastic  & \multicolumn{1}{c|}{materials}    & cardboard             \\ 
            \multicolumn{1}{|c|}{cardboard} & junk   & \multicolumn{1}{c|}{paperback}   & print             \\ 
            \multicolumn{1}{|c|}{printed}  & cardboard   & \multicolumn{1}{c|}{cardboard}   & tissue            \\ \hline
        \end{tabular}
       \caption{5-nearest neighbors for the general word \textbf{paper} with $i$=\texttt{Clothing Shoes and Jewelry}.}

    \end{subtable}
    \caption{Example predictions of $\mymodele{}$ and $\mymodelc{}_i$ using 5-nearest neighbors from embedding layer weights. m denotes model capacity. All models here use data size of 100\%. }
    \label{tab:predictions_using_embedding_more}
\end{table*}

\begin{table*}[ht]
\small
    \begin{subtable}[h]{0.45\textwidth}
        \centering
        \begin{tabular}{|cc|cc|}
            \hline
            \multicolumn{2}{|c|}{m=50\%} & \multicolumn{2}{c|}{m=100\%} \\ \hline
            \multicolumn{1}{|c|}{$\mymodele{}$} & $\mymodelc{}_i$ & \multicolumn{1}{c|}{$\mymodele{}$} & $\mymodelc{}_i$     \\ \hline
            \multicolumn{1}{|c|}{food}        & counter       & \multicolumn{1}{c|}{bottle} & counter        \\ 
            \multicolumn{1}{|c|}{counter}        & hands       & \multicolumn{1}{c|}{refrigerator}       & bottle        \\ 
            \multicolumn{1}{|c|}{wine}        & oil      & \multicolumn{1}{c|}{wine} & hands        \\ 
            \multicolumn{1}{|c|}{oil}        & food       & \multicolumn{1}{c|}{food}    & sink        \\ 
            \multicolumn{1}{|c|}{salad}        & salad       & \multicolumn{1}{c|}{fridge}        & stove        \\ \hline
            \end{tabular}
       \caption{\textit{I realize the point of my purchase was to reduce the amount of olive oil I sprayed on my [MASK] but I do end up having to pump it up and mist twice.} The masked word is a domain-specific word \textbf{salad} with $i$=\texttt{Home and Kitchen}. }

    \end{subtable}
    \hfill
    \begin{subtable}[h]{0.45\textwidth}
        \centering
        \begin{tabular}{|cc|cc|}
            \hline
            \multicolumn{2}{|c|}{m=50\%}    & \multicolumn{2}{c|}{m=100\%}           \\ \hline
            \multicolumn{1}{|c|}{$\mymodele{}$} & $\mymodelc{}_i$ & \multicolumn{1}{c|}{$\mymodele{}$} & $\mymodelc{}_i$ \\ \hline
            \multicolumn{1}{|c|}{guy}   & guy & \multicolumn{1}{c|}{girl}   & guy             \\ 
            \multicolumn{1}{|c|}{musician}      & woman         & \multicolumn{1}{c|}{guy}    & woman             \\ 
            \multicolumn{1}{|c|}{dude}       & man         & \multicolumn{1}{c|}{killer}  & hero             \\ 
            \multicolumn{1}{|c|}{kid}       & kid     & \multicolumn{1}{c|}{gal}    & cop             \\ 
            \multicolumn{1}{|c|}{vampire}    & person     & \multicolumn{1}{c|}{dude}      & man             \\ \hline
        \end{tabular}
       \caption{\textit{There had to be the four friends-a hypochondriac, a smoothing-talking [MASK] who gets everyone in trouble, the joker's friend who's a bit of a ham but has slightly more brains, and a girl.} The masked word is a domain-specific word \textbf{joker} with $i$=\texttt{Movies and TV}.}

    \end{subtable}
    \hfill
    \begin{subtable}[h]{0.45\textwidth}
        \centering
        \begin{tabular}{|cc|cc|}
            \hline
            \multicolumn{2}{|c|}{m=50\%}            & \multicolumn{2}{c|}{m=100\%}           \\ \hline
            \multicolumn{1}{|c|}{$\mymodele{}$} & $\mymodelc{}_i$ & \multicolumn{1}{c|}{$\mymodele{}$} & $\mymodelc{}_i$ \\ \hline
            \multicolumn{1}{|c|}{say}   & have    & \multicolumn{1}{c|}{worry}  & worry             \\ 
            \multicolumn{1}{|c|}{think}   & say    & \multicolumn{1}{c|}{complain}  & say             \\ 
            \multicolumn{1}{|c|}{complain}   & know    & \multicolumn{1}{c|}{wonder}  & know             \\ 
            \multicolumn{1}{|c|}{know}   & care    & \multicolumn{1}{c|}{know}  & think             \\ 
            \multicolumn{1}{|c|}{worry}   & understand    & \multicolumn{1}{c|}{say}  & complain             \\ \hline
        \end{tabular}
       \caption{\textit{Amazon replaced it with no hassle, but I always have to [MASK] about these drives.} The masked word is a general word \textbf{worry} with $i$=\texttt{Electronics}.}

    \end{subtable}
    \hfill
    \begin{subtable}[h]{0.45\textwidth}
        \centering
        \begin{tabular}{|cc|cc|}
            \hline
            \multicolumn{2}{|c|}{m=50\%}            & \multicolumn{2}{c|}{m=100\%}           \\ \hline
            \multicolumn{1}{|c|}{$\mymodele{}$} & $\mymodelc{}_i$ & \multicolumn{1}{c|}{$\mymodele{}$} & $\mymodelc{}_i$ \\ \hline
            \multicolumn{1}{|c|}{instructed}  & expected & \multicolumn{1}{c|}{suggested}   & suggested             \\ 
            \multicolumn{1}{|c|}{suggested} & instructed  & \multicolumn{1}{c|}{stated}   & instructed             \\ 
            \multicolumn{1}{|c|}{well} & stated  & \multicolumn{1}{c|}{instructed}    & expected             \\ 
            \multicolumn{1}{|c|}{usual} & advertised   & \multicolumn{1}{c|}{advertised}   & well             \\ 
            \multicolumn{1}{|c|}{indicated}  & normal   & \multicolumn{1}{c|}{well}   & stated            \\ \hline
        \end{tabular}
       \caption{\textit{I ordered a half size down as [MASK] and the size 11 eclipses my foot.} The masked word is a general word \textbf{suggested} with $i$=\texttt{Clothing Shoes and Jewelry}.}

    \end{subtable}
    \caption{Example MLM predictions of $\mymodele{}$ and $\mymodelc{}_i$ using last layer representation. m denotes model capacity. All models here use a data size of 100\%. }
    \label{tab:predictions_using_last_layer_more}
\end{table*}

\subsection{Additional Results on WikiSum}
\label{app:wikisum}
\begin{figure*}[ht]
     \centering
     \begin{subfigure}[b]{0.32\textwidth}
         \centering
         \includegraphics[width=\textwidth]{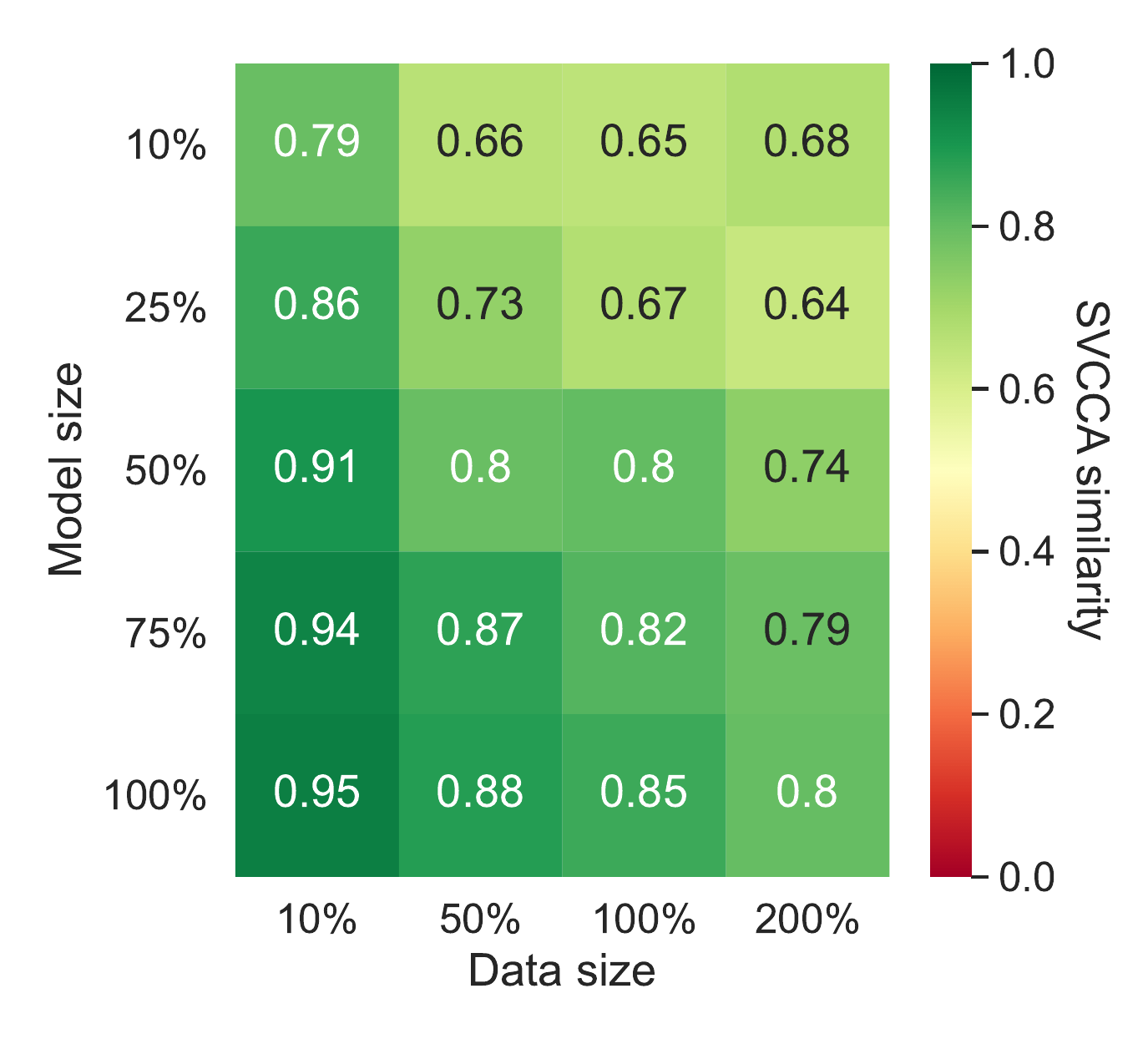}
         \caption{$\ell_0$: all words}
         \label{fig:health-l0-all}
     \end{subfigure}
     \hfill
     \begin{subfigure}[b]{0.32\textwidth}
         \centering
         \includegraphics[width=\textwidth]{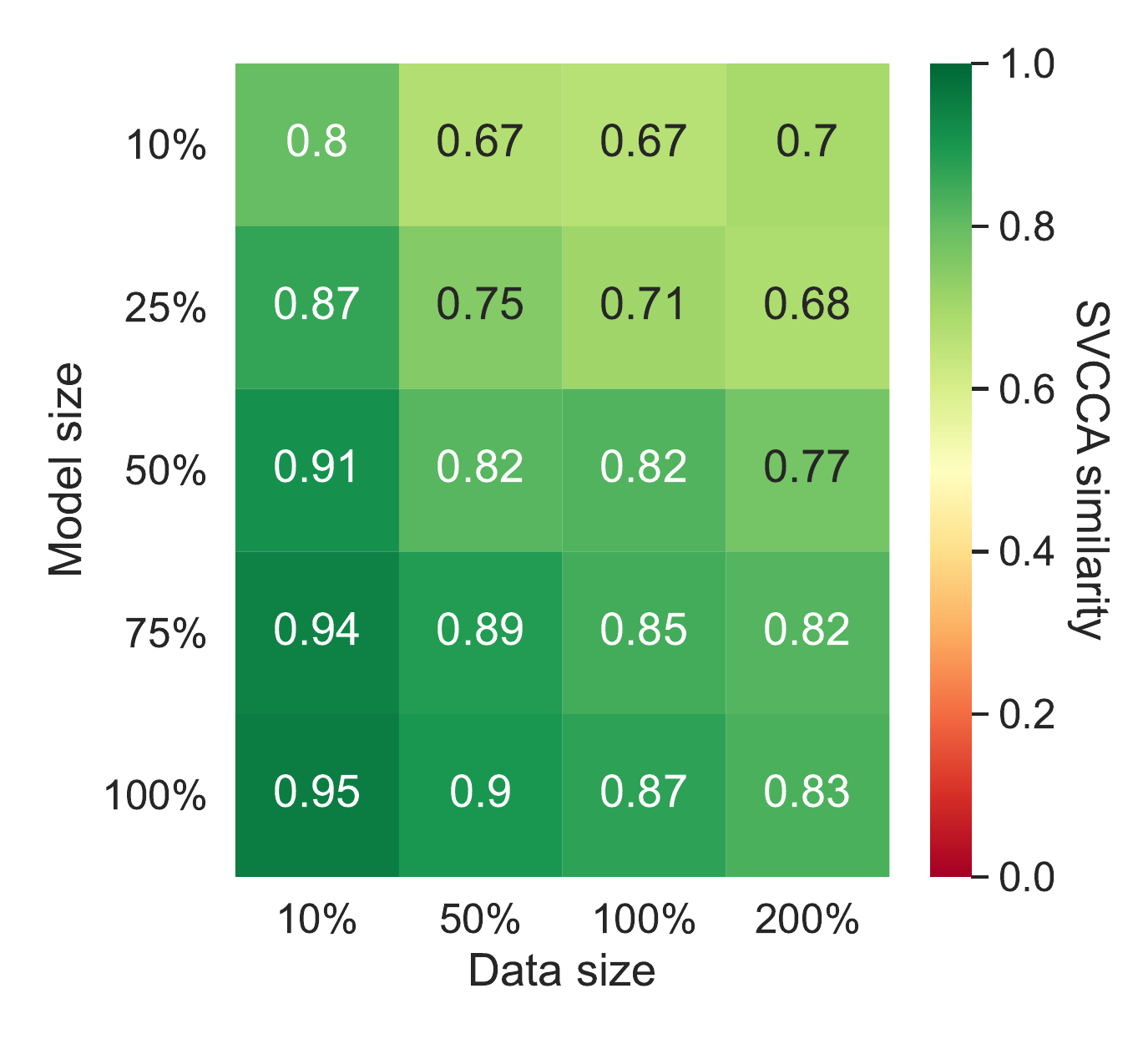}
         \caption{$\ell_0$: general words}
         \label{fig:health-l0-general}
     \end{subfigure}
     \hfill
     \begin{subfigure}[b]{0.32\textwidth}
         \centering
         \includegraphics[width=\textwidth]{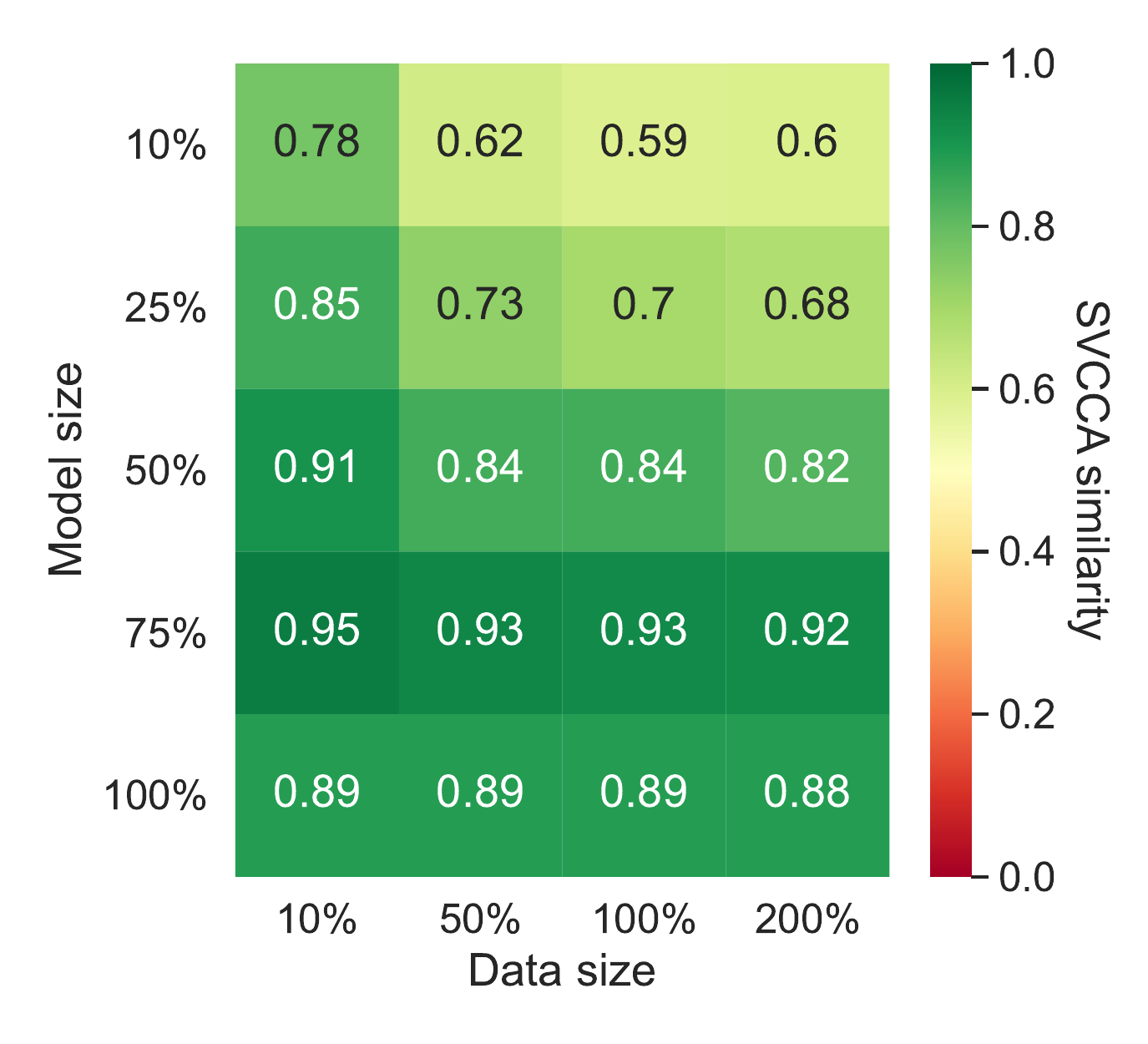}
         \caption{$\ell_0$: domain-specific words}
         \label{fig:health-l0-specific}
     \end{subfigure}
     \hfill
     \begin{subfigure}[b]{0.32\textwidth}
         \centering
         \includegraphics[width=\textwidth]{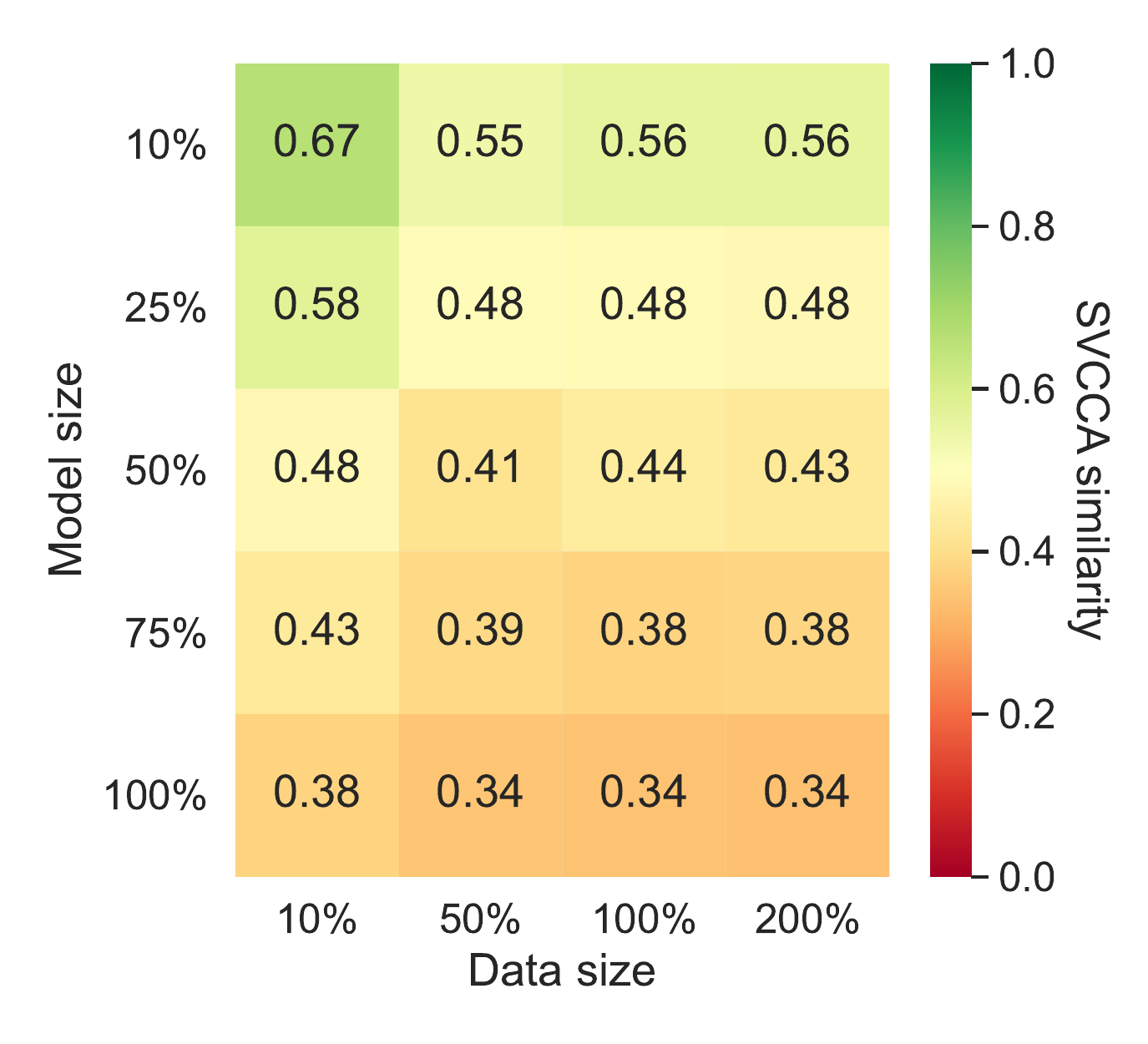}
         \caption{$\ell_{12}$: all words}
         \label{fig:health-l12-all}
     \end{subfigure}
     \hfill
     \begin{subfigure}[b]{0.32\textwidth}
         \centering
         \includegraphics[width=\textwidth]{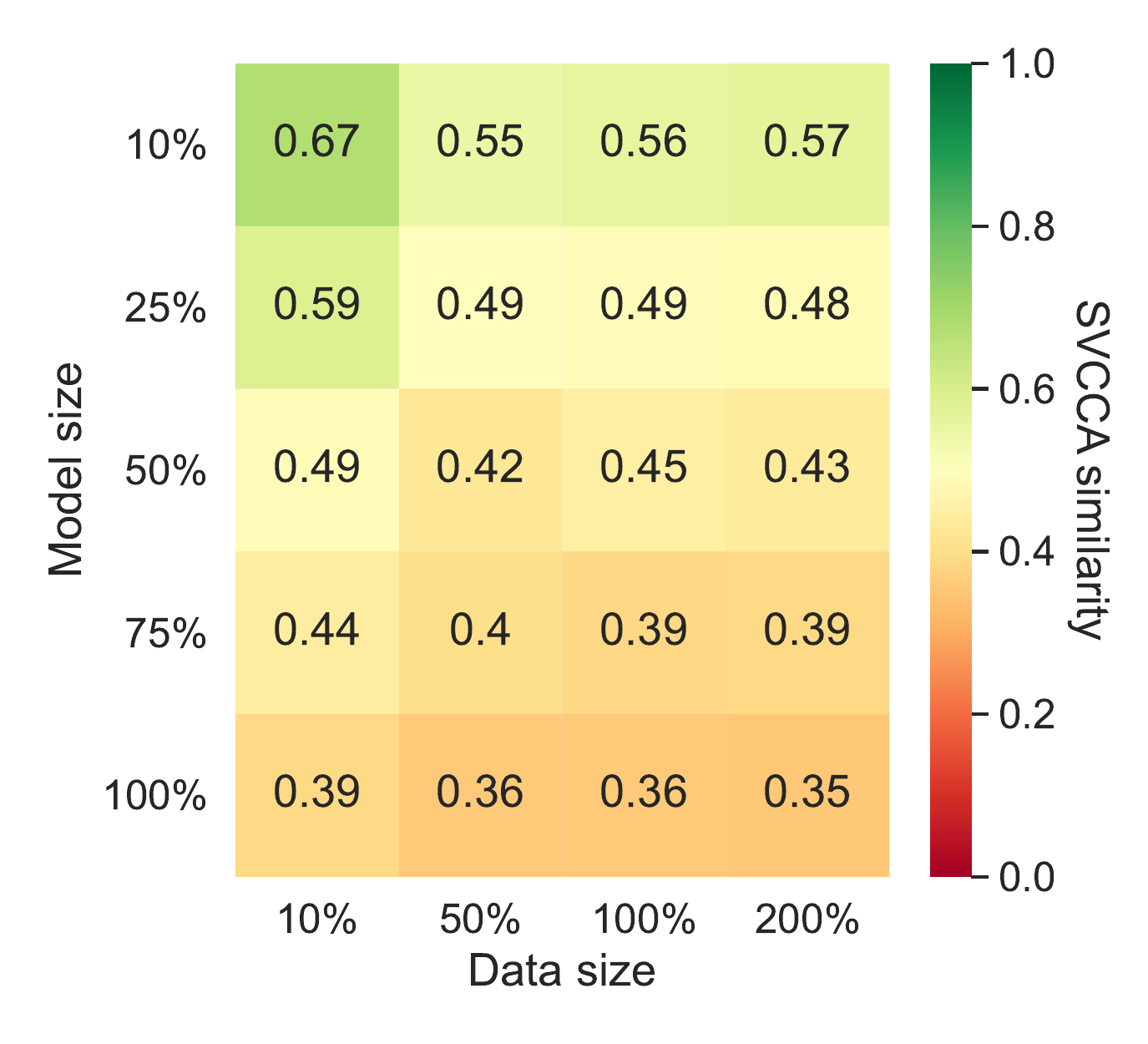}
         \caption{$\ell_{12}$: general words}
         \label{fig:health-l12-general}
     \end{subfigure}
     \hfill
     \begin{subfigure}[b]{0.32\textwidth}
         \centering
         \includegraphics[width=\textwidth]{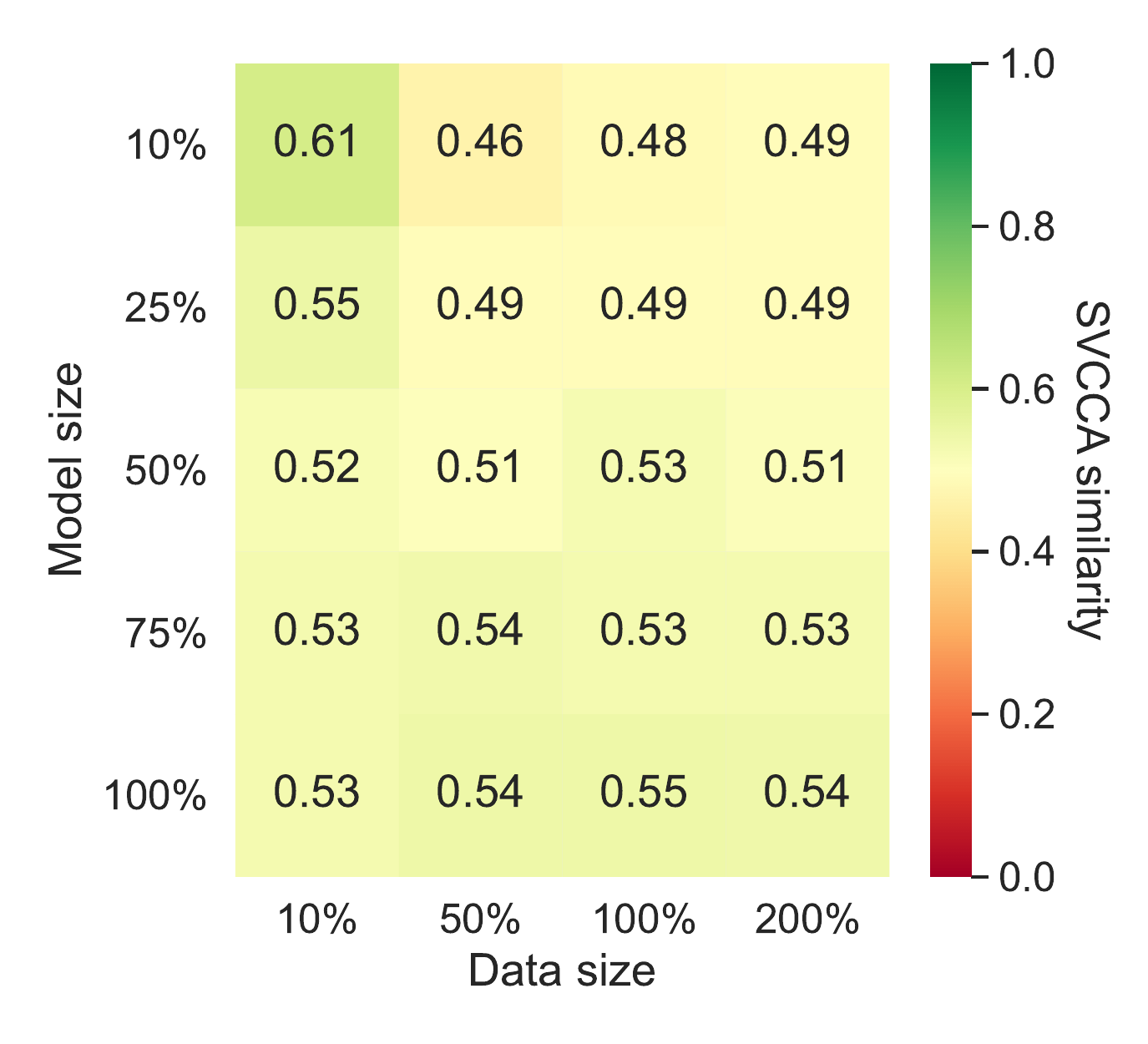}
         \caption{$\ell_{12}$: domain-specific words}
         \label{fig:health-l12-specific}
     \end{subfigure}
\caption{The SVCCA score between $\mymodele{}$ and $\mymodelc{}_{Health}$ for different subsets of tokens. The top row presents the results for the embedding layer $\ell_{0}$, and the bottom row presents them for the last layer $\ell_{12}$. }
\label{fig:health-all}
\end{figure*}

Here we provide additional results on WikiSum \texttt{Health} domain in Figure~\ref{fig:health-all}, including SVCCA results between $\mymodele{}$ and $\mymodelc{}_{Health}$, as well as results for different subsets of tokens. 

%% file: acl.bbl
\begin{thebibliography}{48}
\expandafter\ifx\csname natexlab\endcsname\relax\def\natexlab#1{#1}\fi

\bibitem[{Bau et~al.(2019)Bau, Belinkov, Sajjad, Durrani, Dalvi, and
  Glass}]{bau-etal-2019-identifying}
Anthony Bau, Yonatan Belinkov, Hassan Sajjad, Nadir Durrani, Fahim Dalvi, and
  James~R. Glass. 2019.
\newblock \href {https://openreview.net/forum?id=H1z-PsR5KX} {Identifying and
  controlling important neurons in neural machine translation}.
\newblock In \emph{7th International Conference on Learning Representations,
  {ICLR} 2019, New Orleans, LA, USA, May 6-9, 2019}. OpenReview.net.

\bibitem[{Belinkov et~al.(2017{\natexlab{a}})Belinkov, Durrani, Dalvi, Sajjad,
  and Glass}]{belinkov-etal-2017-neural}
Yonatan Belinkov, Nadir Durrani, Fahim Dalvi, Hassan Sajjad, and James Glass.
  2017{\natexlab{a}}.
\newblock \href {https://doi.org/10.18653/v1/P17-1080} {What do neural machine
  translation models learn about morphology?}
\newblock In \emph{Proceedings of the 55th Annual Meeting of the Association
  for Computational Linguistics (Volume 1: Long Papers)}, pages 861--872,
  Vancouver, Canada. Association for Computational Linguistics.

\bibitem[{Belinkov et~al.(2017{\natexlab{b}})Belinkov, M{\`a}rquez, Sajjad,
  Durrani, Dalvi, and Glass}]{belinkov-etal-2017-evaluating}
Yonatan Belinkov, Llu{\'\i}s M{\`a}rquez, Hassan Sajjad, Nadir Durrani, Fahim
  Dalvi, and James Glass. 2017{\natexlab{b}}.
\newblock \href {https://aclanthology.org/I17-1001} {Evaluating layers of
  representation in neural machine translation on part-of-speech and semantic
  tagging tasks}.
\newblock In \emph{Proceedings of the Eighth International Joint Conference on
  Natural Language Processing (Volume 1: Long Papers)}, pages 1--10, Taipei,
  Taiwan. Asian Federation of Natural Language Processing.

\bibitem[{Blitzer et~al.(2007)Blitzer, Dredze, and
  Pereira}]{blitzer-etal-2007-biographies}
John Blitzer, Mark Dredze, and Fernando Pereira. 2007.
\newblock \href {https://aclanthology.org/P07-1056} {Biographies, {B}ollywood,
  boom-boxes and blenders: Domain adaptation for sentiment classification}.
\newblock In \emph{Proceedings of the 45th Annual Meeting of the Association of
  Computational Linguistics}, pages 440--447, Prague, Czech Republic.
  Association for Computational Linguistics.

\bibitem[{Brown et~al.(2020)Brown, Mann, Ryder, Subbiah, Kaplan, Dhariwal,
  Neelakantan, Shyam, Sastry, Askell, Agarwal, Herbert-Voss, Krueger, Henighan,
  Child, Ramesh, Ziegler, Wu, Winter, Hesse, Chen, Sigler, Litwin, Gray, Chess,
  Clark, Berner, McCandlish, Radford, Sutskever, and
  Amodei}]{brown-etal-2020-language}
Tom~B. Brown, Benjamin Mann, Nick Ryder, Melanie Subbiah, Jared Kaplan,
  Prafulla Dhariwal, Arvind Neelakantan, Pranav Shyam, Girish Sastry, Amanda
  Askell, Sandhini Agarwal, Ariel Herbert-Voss, Gretchen Krueger, Tom Henighan,
  Rewon Child, Aditya Ramesh, Daniel~M. Ziegler, Jeffrey Wu, Clemens Winter,
  Christopher Hesse, Mark Chen, Eric Sigler, Mateusz Litwin, Scott Gray,
  Benjamin Chess, Jack Clark, Christopher Berner, Sam McCandlish, Alec Radford,
  Ilya Sutskever, and Dario Amodei. 2020.
\newblock Language models are few-shot learners.
\newblock In \emph{Proceedings of the 34th International Conference on Neural
  Information Processing Systems}, NIPS'20, Red Hook, NY, USA. Curran
  Associates Inc.

\bibitem[{Clark et~al.(2020)Clark, Luong, Le, and Manning}]{clark2020electra}
Kevin Clark, Minh-Thang Luong, Quoc~V. Le, and Christopher~D. Manning. 2020.
\newblock \href {https://openreview.net/pdf?id=r1xMH1BtvB} {{ELECTRA}:
  Pre-training text encoders as discriminators rather than generators}.
\newblock In \emph{ICLR}.

\bibitem[{Cohan et~al.(2018)Cohan, Dernoncourt, Kim, Bui, Kim, Chang, and
  Goharian}]{cohan-etal-2018-discourse}
Arman Cohan, Franck Dernoncourt, Doo~Soon Kim, Trung Bui, Seokhwan Kim, Walter
  Chang, and Nazli Goharian. 2018.
\newblock \href {https://doi.org/10.18653/v1/N18-2097} {A discourse-aware
  attention model for abstractive summarization of long documents}.
\newblock In \emph{Proceedings of the 2018 Conference of the North {A}merican
  Chapter of the Association for Computational Linguistics: Human Language
  Technologies, Volume 2 (Short Papers)}, pages 615--621, New Orleans,
  Louisiana. Association for Computational Linguistics.

\bibitem[{Cohen et~al.(2021)Cohen, Kalinsky, Ziser, and
  Moschitti}]{cohen-etal-2021-wikisum}
Nachshon Cohen, Oren Kalinsky, Yftah Ziser, and Alessandro Moschitti. 2021.
\newblock \href {https://doi.org/10.18653/v1/2021.acl-short.28} {{W}iki{S}um:
  Coherent summarization dataset for efficient human-evaluation}.
\newblock In \emph{Proceedings of the 59th Annual Meeting of the Association
  for Computational Linguistics and the 11th International Joint Conference on
  Natural Language Processing (Volume 2: Short Papers)}, pages 212--219,
  Online. Association for Computational Linguistics.

\bibitem[{de~Vries et~al.(2020)de~Vries, van Cranenburgh, and
  Nissim}]{de-vries-etal-2020-whats}
Wietse de~Vries, Andreas van Cranenburgh, and Malvina Nissim. 2020.
\newblock \href {https://doi.org/10.18653/v1/2020.findings-emnlp.389} {What{'}s
  so special about {BERT}{'}s layers? a closer look at the {NLP} pipeline in
  monolingual and multilingual models}.
\newblock In \emph{Findings of the Association for Computational Linguistics:
  EMNLP 2020}, pages 4339--4350, Online. Association for Computational
  Linguistics.

\bibitem[{Devlin et~al.(2019)Devlin, Chang, Lee, and
  Toutanova}]{devlin-etal-2019-bert}
Jacob Devlin, Ming-Wei Chang, Kenton Lee, and Kristina Toutanova. 2019.
\newblock \href {https://doi.org/10.18653/v1/N19-1423} {{BERT}: Pre-training of
  deep bidirectional transformers for language understanding}.
\newblock In \emph{Proceedings of the 2019 Conference of the North {A}merican
  Chapter of the Association for Computational Linguistics: Human Language
  Technologies, Volume 1 (Long and Short Papers)}, pages 4171--4186,
  Minneapolis, Minnesota. Association for Computational Linguistics.

\bibitem[{Dimsdale-Zucker and Ranganath(2018)}]{dimsdale2018representational}
Halle~R Dimsdale-Zucker and Charan Ranganath. 2018.
\newblock Representational similarity analyses: a practical guide for
  functional mri applications.
\newblock In \emph{Handbook of behavioral neuroscience}, volume~28, pages
  509--525. Elsevier.

\bibitem[{Du et~al.(2020)Du, Sun, Wang, Qi, and
  Liao}]{du-etal-2020-adversarial}
Chunning Du, Haifeng Sun, Jingyu Wang, Qi~Qi, and Jianxin Liao. 2020.
\newblock \href {https://doi.org/10.18653/v1/2020.acl-main.370} {Adversarial
  and domain-aware {BERT} for cross-domain sentiment analysis}.
\newblock In \emph{Proceedings of the 58th Annual Meeting of the Association
  for Computational Linguistics}, pages 4019--4028, Online. Association for
  Computational Linguistics.

\bibitem[{Fonseca et~al.(2022)Fonseca, Ziser, and Cohen}]{fonseca2022}
Marcio Fonseca, Yftah Ziser, and Shay~B. Cohen. 2022.
\newblock \href {https://arxiv.org/abs/2205.12486} {Factorizing content and
  budget decisions in abstractive summarization of long documents by sampling
  summary views}.
\newblock In \emph{Proceedings of the Conference on Empirical Methods in
  Natural Language Processing (EMNLP)}.

\bibitem[{Frankle and Carbin(2019)}]{frankle-and-carbin-2019-lottery}
Jonathan Frankle and Michael Carbin. 2019.
\newblock \href {https://openreview.net/forum?id=rJl-b3RcF7} {The lottery
  ticket hypothesis: Finding sparse, trainable neural networks}.
\newblock In \emph{7th International Conference on Learning Representations,
  {ICLR} 2019, New Orleans, LA, USA, May 6-9, 2019}. OpenReview.net.

\bibitem[{Giulianelli et~al.(2018)Giulianelli, Harding, Mohnert, Hupkes, and
  Zuidema}]{giulianelli-etal-2018-hood}
Mario Giulianelli, Jack Harding, Florian Mohnert, Dieuwke Hupkes, and Willem
  Zuidema. 2018.
\newblock \href {https://doi.org/10.18653/v1/W18-5426} {Under the hood: Using
  diagnostic classifiers to investigate and improve how language models track
  agreement information}.
\newblock In \emph{Proceedings of the 2018 {EMNLP} Workshop {B}lackbox{NLP}:
  Analyzing and Interpreting Neural Networks for {NLP}}, pages 240--248,
  Brussels, Belgium. Association for Computational Linguistics.

\bibitem[{Grusky et~al.(2018)Grusky, Naaman, and
  Artzi}]{grusky-etal-2018-newsroom}
Max Grusky, Mor Naaman, and Yoav Artzi. 2018.
\newblock \href {https://doi.org/10.18653/v1/N18-1065} {{N}ewsroom: A dataset
  of 1.3 million summaries with diverse extractive strategies}.
\newblock In \emph{Proceedings of the 2018 Conference of the North {A}merican
  Chapter of the Association for Computational Linguistics: Human Language
  Technologies, Volume 1 (Long Papers)}, pages 708--719, New Orleans,
  Louisiana. Association for Computational Linguistics.

\bibitem[{Gu et~al.(2021)Gu, Tinn, Cheng, Lucas, Usuyama, Liu, Naumann, Gao,
  and Poon}]{gu-etal-2021-domain}
Yu~Gu, Robert Tinn, Hao Cheng, Michael Lucas, Naoto Usuyama, Xiaodong Liu,
  Tristan Naumann, Jianfeng Gao, and Hoifung Poon. 2021.
\newblock \href {https://doi.org/10.1145/3458754} {Domain-specific language
  model pretraining for biomedical natural language processing}.
\newblock \emph{ACM Trans. Comput. Healthcare}, 3(1).

\bibitem[{Hardoon et~al.(2004)Hardoon, Szedm{\'{a}}k, and
  Shawe{-}Taylor}]{hardoon-etal-2004-cca}
David~R. Hardoon, S{\'{a}}ndor Szedm{\'{a}}k, and John Shawe{-}Taylor. 2004.
\newblock \href {https://doi.org/10.1162/0899766042321814} {Canonical
  correlation analysis: An overview with application to learning methods}.
\newblock \emph{Neural Comput.}, 16(12):2639--2664.

\bibitem[{Hazen et~al.(2019)Hazen, Dhuliawala, and Boies}]{hazen2019towards}
Timothy~J Hazen, Shehzaad Dhuliawala, and Daniel Boies. 2019.
\newblock Towards domain adaptation from limited data for question answering
  using deep neural networks.
\newblock \emph{arXiv preprint arXiv:1911.02655}.

\bibitem[{Hinkelmann and Kempthorne(2007)}]{hinkelmann2007design}
Klaus Hinkelmann and Oscar Kempthorne. 2007.
\newblock \emph{Design and analysis of experiments, volume 1: Introduction to
  experimental design}, volume~1.
\newblock John Wiley \& Sons.

\bibitem[{Kudugunta et~al.(2019)Kudugunta, Bapna, Caswell, and
  Firat}]{kudugunta-etal-2019-investigating}
Sneha Kudugunta, Ankur Bapna, Isaac Caswell, and Orhan Firat. 2019.
\newblock \href {https://doi.org/10.18653/v1/D19-1167} {Investigating
  multilingual {NMT} representations at scale}.
\newblock In \emph{Proceedings of the 2019 Conference on Empirical Methods in
  Natural Language Processing and the 9th International Joint Conference on
  Natural Language Processing (EMNLP-IJCNLP)}, pages 1565--1575, Hong Kong,
  China. Association for Computational Linguistics.

\bibitem[{Lekhtman et~al.(2021)Lekhtman, Ziser, and
  Reichart}]{lekhtman-etal-2021-dilbert}
Entony Lekhtman, Yftah Ziser, and Roi Reichart. 2021.
\newblock \href {https://doi.org/10.18653/v1/2021.emnlp-main.20} {{DILBERT}:
  Customized pre-training for domain adaptation with category shift, with an
  application to aspect extraction}.
\newblock In \emph{Proceedings of the 2021 Conference on Empirical Methods in
  Natural Language Processing}, pages 219--230, Online and Punta Cana,
  Dominican Republic. Association for Computational Linguistics.

\bibitem[{Liu et~al.(2019)Liu, Ott, Goyal, Du, Joshi, Chen, Levy, Lewis,
  Zettlemoyer, and Stoyanov}]{liu2019roberta}
Yinhan Liu, Myle Ott, Naman Goyal, Jingfei Du, Mandar Joshi, Danqi Chen, Omer
  Levy, Mike Lewis, Luke Zettlemoyer, and Veselin Stoyanov. 2019.
\newblock Roberta: A robustly optimized bert pretraining approach.
\newblock \emph{arXiv preprint arXiv:1907.11692}.

\bibitem[{Long et~al.(2022)Long, Luo, Wang, and Pan}]{long-etal-2022-domain}
Quanyu Long, Tianze Luo, Wenya Wang, and Sinno Pan. 2022.
\newblock \href {https://doi.org/10.18653/v1/2022.naacl-main.217} {Domain
  confused contrastive learning for unsupervised domain adaptation}.
\newblock In \emph{Proceedings of the 2022 Conference of the North American
  Chapter of the Association for Computational Linguistics: Human Language
  Technologies}, pages 2982--2995, Seattle, United States. Association for
  Computational Linguistics.

\bibitem[{Magar and Schwartz(2022)}]{magar-schwartz-2022-data}
Inbal Magar and Roy Schwartz. 2022.
\newblock \href {https://doi.org/10.18653/v1/2022.acl-short.18} {Data
  contamination: From memorization to exploitation}.
\newblock In \emph{Proceedings of the 60th Annual Meeting of the Association
  for Computational Linguistics (Volume 2: Short Papers)}, pages 157--165,
  Dublin, Ireland. Association for Computational Linguistics.

\bibitem[{Morcos et~al.(2018)Morcos, Raghu, and
  Bengio}]{morcos-etal-2018-insights}
Ari~S. Morcos, Maithra Raghu, and Samy Bengio. 2018.
\newblock \href
  {https://proceedings.neurips.cc/paper/2018/hash/a7a3d70c6d17a73140918996d03c014f-Abstract.html}
  {Insights on representational similarity in neural networks with canonical
  correlation}.
\newblock In \emph{Advances in Neural Information Processing Systems 31: Annual
  Conference on Neural Information Processing Systems 2018, NeurIPS 2018,
  December 3-8, 2018, Montr{\'{e}}al, Canada}, pages 5732--5741.

\bibitem[{Ni et~al.(2019)Ni, Li, and McAuley}]{ni-etal-2019-justifying}
Jianmo Ni, Jiacheng Li, and Julian McAuley. 2019.
\newblock \href {https://doi.org/10.18653/v1/D19-1018} {Justifying
  recommendations using distantly-labeled reviews and fine-grained aspects}.
\newblock In \emph{Proceedings of the 2019 Conference on Empirical Methods in
  Natural Language Processing and the 9th International Joint Conference on
  Natural Language Processing (EMNLP-IJCNLP)}, pages 188--197, Hong Kong,
  China. Association for Computational Linguistics.

\bibitem[{Paszke et~al.(2019)Paszke, Gross, Massa, Lerer, Bradbury, Chanan,
  Killeen, Lin, Gimelshein, Antiga, Desmaison, Kopf, Yang, DeVito, Raison,
  Tejani, Chilamkurthy, Steiner, Fang, Bai, and Chintala}]{NEURIPS2019_9015}
Adam Paszke, Sam Gross, Francisco Massa, Adam Lerer, James Bradbury, Gregory
  Chanan, Trevor Killeen, Zeming Lin, Natalia Gimelshein, Luca Antiga, Alban
  Desmaison, Andreas Kopf, Edward Yang, Zachary DeVito, Martin Raison, Alykhan
  Tejani, Sasank Chilamkurthy, Benoit Steiner, Lu~Fang, Junjie Bai, and Soumith
  Chintala. 2019.
\newblock \href
  {http://papers.neurips.cc/paper/9015-pytorch-an-imperative-style-high-performance-deep-learning-library.pdf}
  {Pytorch: An imperative style, high-performance deep learning library}.
\newblock In \emph{NeurIPS}.

\bibitem[{Plank(2016)}]{plank-2016-what}
Barbara Plank. 2016.
\newblock \href
  {https://www.linguistics.rub.de/konvens16/pub/2\_konvensproc.pdf} {What to do
  about non-standard (or non-canonical) language in {NLP}}.
\newblock In \emph{Proceedings of the 13th Conference on Natural Language
  Processing, {KONVENS} 2016, Bochum, Germany, September 19-21, 2016},
  volume~16 of \emph{Bochumer Linguistische Arbeitsberichte}.

\bibitem[{Raffel et~al.(2020)Raffel, Shazeer, Roberts, Lee, Narang, Matena,
  Zhou, Li, and Liu}]{raffel-etal-2020-exploring}
Colin Raffel, Noam Shazeer, Adam Roberts, Katherine Lee, Sharan Narang, Michael
  Matena, Yanqi Zhou, Wei Li, and Peter~J. Liu. 2020.
\newblock \href {http://jmlr.org/papers/v21/20-074.html} {Exploring the limits
  of transfer learning with a unified text-to-text transformer}.
\newblock \emph{JMLR}.

\bibitem[{Raghu et~al.(2017)Raghu, Gilmer, Yosinski, and
  Sohl{-}Dickstein}]{raghu-etal-2017-svcca}
Maithra Raghu, Justin Gilmer, Jason Yosinski, and Jascha Sohl{-}Dickstein.
  2017.
\newblock \href
  {https://proceedings.neurips.cc/paper/2017/hash/dc6a7e655d7e5840e66733e9ee67cc69-Abstract.html}
  {{SVCCA:} singular vector canonical correlation analysis for deep learning
  dynamics and interpretability}.
\newblock In \emph{Advances in Neural Information Processing Systems 30: Annual
  Conference on Neural Information Processing Systems 2017, December 4-9, 2017,
  Long Beach, CA, {USA}}, pages 6076--6085.

\bibitem[{Rajpurkar et~al.(2016)Rajpurkar, Zhang, Lopyrev, and
  Liang}]{rajpurkar-etal-2016-squad}
Pranav Rajpurkar, Jian Zhang, Konstantin Lopyrev, and Percy Liang. 2016.
\newblock \href {https://doi.org/10.18653/v1/D16-1264} {{SQ}u{AD}: 100,000+
  questions for machine comprehension of text}.
\newblock In \emph{Proceedings of the 2016 Conference on Empirical Methods in
  Natural Language Processing}, pages 2383--2392, Austin, Texas. Association
  for Computational Linguistics.

\bibitem[{Rockt{\"a}schel et~al.(2013)Rockt{\"a}schel, Huber, Weidlich, and
  Leser}]{rocktaschel-etal-2013-wbi}
Tim Rockt{\"a}schel, Torsten Huber, Michael Weidlich, and Ulf Leser. 2013.
\newblock \href {https://aclanthology.org/S13-2058} {{WBI}-{NER}: The impact of
  domain-specific features on the performance of identifying and classifying
  mentions of drugs}.
\newblock In \emph{Second Joint Conference on Lexical and Computational
  Semantics (*{SEM}), Volume 2: Proceedings of the Seventh International
  Workshop on Semantic Evaluation ({S}em{E}val 2013)}, pages 356--363, Atlanta,
  Georgia, USA. Association for Computational Linguistics.

\bibitem[{Saphra and Lopez(2019)}]{saphra-lopez-2019-understanding}
Naomi Saphra and Adam Lopez. 2019.
\newblock \href {https://doi.org/10.18653/v1/N19-1329} {Understanding learning
  dynamics of language models with {SVCCA}}.
\newblock In \emph{Proceedings of the 2019 Conference of the North {A}merican
  Chapter of the Association for Computational Linguistics: Human Language
  Technologies, Volume 1 (Long and Short Papers)}, pages 3257--3267,
  Minneapolis, Minnesota. Association for Computational Linguistics.

\bibitem[{Shakeri et~al.(2020)Shakeri, Nogueira~dos Santos, Zhu, Ng, Nan, Wang,
  Nallapati, and Xiang}]{shakeri-etal-2020-end}
Siamak Shakeri, Cicero Nogueira~dos Santos, Henghui Zhu, Patrick Ng, Feng Nan,
  Zhiguo Wang, Ramesh Nallapati, and Bing Xiang. 2020.
\newblock \href {https://doi.org/10.18653/v1/2020.emnlp-main.439} {End-to-end
  synthetic data generation for domain adaptation of question answering
  systems}.
\newblock In \emph{Proceedings of the 2020 Conference on Empirical Methods in
  Natural Language Processing (EMNLP)}, pages 5445--5460, Online. Association
  for Computational Linguistics.

\bibitem[{Shang et~al.(2018)Shang, Liu, Gu, Ren, Ren, and
  Han}]{shang-etal-2018-learning}
Jingbo Shang, Liyuan Liu, Xiaotao Gu, Xiang Ren, Teng Ren, and Jiawei Han.
  2018.
\newblock \href {https://doi.org/10.18653/v1/D18-1230} {Learning named entity
  tagger using domain-specific dictionary}.
\newblock In \emph{Proceedings of the 2018 Conference on Empirical Methods in
  Natural Language Processing}, pages 2054--2064, Brussels, Belgium.
  Association for Computational Linguistics.

\bibitem[{Singh et~al.(2019)Singh, McCann, Socher, and
  Xiong}]{singh-etal-2019-bert}
Jasdeep Singh, Bryan McCann, Richard Socher, and Caiming Xiong. 2019.
\newblock \href {https://doi.org/10.18653/v1/D19-6106} {{BERT} is not an
  interlingua and the bias of tokenization}.
\newblock In \emph{Proceedings of the 2nd Workshop on Deep Learning Approaches
  for Low-Resource NLP (DeepLo 2019)}, pages 47--55, Hong Kong, China.
  Association for Computational Linguistics.

\bibitem[{Srivastava et~al.(2022)Srivastava, Rastogi, Rao, Shoeb, Abid, Fisch,
  Brown, Santoro, Gupta, Garriga-Alonso et~al.}]{srivastava2022beyond}
Aarohi Srivastava, Abhinav Rastogi, Abhishek Rao, Abu Awal~Md Shoeb, Abubakar
  Abid, Adam Fisch, Adam~R Brown, Adam Santoro, Aditya Gupta, Adri{\`a}
  Garriga-Alonso, et~al. 2022.
\newblock Beyond the imitation game: Quantifying and extrapolating the
  capabilities of language models.
\newblock \emph{arXiv preprint arXiv:2206.04615}.

\bibitem[{Vuli{\'c} et~al.(2020)Vuli{\'c}, Ponti, Litschko, Glava{\v{s}}, and
  Korhonen}]{vulic-etal-2020-probing}
Ivan Vuli{\'c}, Edoardo~Maria Ponti, Robert Litschko, Goran Glava{\v{s}}, and
  Anna Korhonen. 2020.
\newblock \href {https://doi.org/10.18653/v1/2020.emnlp-main.586} {Probing
  pretrained language models for lexical semantics}.
\newblock In \emph{Proceedings of the 2020 Conference on Empirical Methods in
  Natural Language Processing (EMNLP)}, pages 7222--7240, Online. Association
  for Computational Linguistics.

\bibitem[{Wang et~al.(2019)Wang, Pruksachatkun, Nangia, Singh, Michael, Hill,
  Levy, and Bowman}]{wang-etal-2019-superglue}
Alex Wang, Yada Pruksachatkun, Nikita Nangia, Amanpreet Singh, Julian Michael,
  Felix Hill, Omer Levy, and Samuel~R. Bowman. 2019.
\newblock \href
  {https://proceedings.neurips.cc/paper/2019/hash/4496bf24afe7fab6f046bf4923da8de6-Abstract.html}
  {Superglue: {A} stickier benchmark for general-purpose language understanding
  systems}.
\newblock In \emph{Advances in Neural Information Processing Systems 32: Annual
  Conference on Neural Information Processing Systems 2019, NeurIPS 2019,
  December 8-14, 2019, Vancouver, BC, Canada}, pages 3261--3275.

\bibitem[{Wang et~al.(2018)Wang, Singh, Michael, Hill, Levy, and
  Bowman}]{wang-etal-2018-glue}
Alex Wang, Amanpreet Singh, Julian Michael, Felix Hill, Omer Levy, and Samuel
  Bowman. 2018.
\newblock \href {https://doi.org/10.18653/v1/W18-5446} {{GLUE}: A multi-task
  benchmark and analysis platform for natural language understanding}.
\newblock In \emph{Proceedings of the 2018 {EMNLP} Workshop {B}lackbox{NLP}:
  Analyzing and Interpreting Neural Networks for {NLP}}, pages 353--355,
  Brussels, Belgium. Association for Computational Linguistics.

\bibitem[{Wang et~al.(2020)Wang, Ding, Hong, Liu, and Caverlee}]{wang2020next}
Jianling Wang, Kaize Ding, Liangjie Hong, Huan Liu, and James Caverlee. 2020.
\newblock Next-item recommendation with sequential hypergraphs.
\newblock In \emph{Proceedings of the 43rd international ACM SIGIR conference
  on research and development in information retrieval}, pages 1101--1110.

\bibitem[{Wolf et~al.(2020)Wolf, Debut, Sanh, Chaumond, Delangue, Moi, Cistac,
  Rault, Louf, Funtowicz, Davison, Shleifer, von Platen, Ma, Jernite, Plu, Xu,
  Le~Scao, Gugger, Drame, Lhoest, and Rush}]{wolf2019huggingface}
Thomas Wolf, Lysandre Debut, Victor Sanh, Julien Chaumond, Clement Delangue,
  Anthony Moi, Pierric Cistac, Tim Rault, Remi Louf, Morgan Funtowicz, Joe
  Davison, Sam Shleifer, Patrick von Platen, Clara Ma, Yacine Jernite, Julien
  Plu, Canwen Xu, Teven Le~Scao, Sylvain Gugger, Mariama Drame, Quentin Lhoest,
  and Alexander Rush. 2020.
\newblock \href {https://doi.org/10.18653/v1/2020.emnlp-demos.6} {Transformers:
  State-of-the-art natural language processing}.
\newblock In \emph{Proc. of EMNLP}.

\bibitem[{Yue et~al.(2021)Yue, Kratzwald, and
  Feuerriegel}]{yue-etal-2021-contrastive}
Zhenrui Yue, Bernhard Kratzwald, and Stefan Feuerriegel. 2021.
\newblock \href {https://doi.org/10.18653/v1/2021.emnlp-main.754} {Contrastive
  domain adaptation for question answering using limited text corpora}.
\newblock In \emph{Proceedings of the 2021 Conference on Empirical Methods in
  Natural Language Processing}, pages 9575--9593, Online and Punta Cana,
  Dominican Republic. Association for Computational Linguistics.

\bibitem[{Zhang and Bowman(2018)}]{zhang-bowman-2018-language}
Kelly Zhang and Samuel Bowman. 2018.
\newblock \href {https://doi.org/10.18653/v1/W18-5448} {Language modeling
  teaches you more than translation does: Lessons learned through auxiliary
  syntactic task analysis}.
\newblock In \emph{Proceedings of the 2018 {EMNLP} Workshop {B}lackbox{NLP}:
  Analyzing and Interpreting Neural Networks for {NLP}}, pages 359--361,
  Brussels, Belgium. Association for Computational Linguistics.

\bibitem[{Zhang et~al.(2020)Zhang, Zhang, Zhong, and
  Wang}]{zhang2020multiclassification}
Shaozhong Zhang, Dingkai Zhang, Haidong Zhong, and Guorong Wang. 2020.
\newblock A multiclassification model of sentiment for e-commerce reviews.
\newblock \emph{IEEE Access}, 8:189513--189526.

\bibitem[{Ziser and Reichart(2017)}]{ziser-reichart-2017-neural}
Yftah Ziser and Roi Reichart. 2017.
\newblock \href {https://doi.org/10.18653/v1/K17-1040} {Neural structural
  correspondence learning for domain adaptation}.
\newblock In \emph{Proceedings of the 21st Conference on Computational Natural
  Language Learning ({C}o{NLL} 2017)}, pages 400--410, Vancouver, Canada.
  Association for Computational Linguistics.

\bibitem[{Ziser and Reichart(2018)}]{ziser-reichart-2018-pivot}
Yftah Ziser and Roi Reichart. 2018.
\newblock \href {https://doi.org/10.18653/v1/N18-1112} {Pivot based language
  modeling for improved neural domain adaptation}.
\newblock In \emph{Proceedings of the 2018 Conference of the North {A}merican
  Chapter of the Association for Computational Linguistics: Human Language
  Technologies, Volume 1 (Long Papers)}, pages 1241--1251, New Orleans,
  Louisiana. Association for Computational Linguistics.

\end{thebibliography}
